\documentclass{article}
\usepackage[square,numbers]{natbib}
\usepackage{amsmath,amssymb,mathtools,bm,empheq,amsthm}

\theoremstyle{plain}

\theoremstyle{definition}

\theoremstyle{remark}

\def\bal#1\eal{\begin{align}#1\end{align}} % align environment

\DeclareMathOperator*{\argmin}{arg\,min} % argmin
\DeclareMathOperator*{\argmax}{arg\,max} % argmax

\def\m{\mathbf}

\def\R{\mathbb{R}}

\newcommand{\norm}[2]{\ensuremath{\left\|#1\right\|_{#2}}}

\newcommand {\bbmtx}{\begin{bmatrix}} % begin matrix environment
\newcommand {\ebmtx}{\end{bmatrix}} % end matrix environment

 \usepackage[main, final]{neurips_2026}
\usepackage[utf8]{inputenc} % allow utf-8 input
\usepackage[T1]{fontenc}    % use 8-bit T1 fonts
\usepackage{hyperref}       % hyperlinks
\usepackage{url}            % simple URL typesetting
\usepackage{booktabs}       % professional-quality tables
\usepackage{amsfonts}       % blackboard math symbols
\usepackage{nicefrac}       % compact symbols for 1/2, etc.
\usepackage{microtype}      % microtypography
\usepackage{xcolor}         % colors

\usepackage{algorithm}
\usepackage{algorithmic}
\usepackage{adjustbox}
\usepackage{multirow}
\usepackage{amsmath}
\usepackage{pifont}% http://ctan.org/pkg/pifont
\usepackage{xcolor}
\usepackage{pgf}
\usepackage{caption}
\usepackage{arydshln} % dashed line
\definecolor{amaranth}{rgb}{0.9, 0.17, 0.31}
\definecolor{americanrose}{rgb}{1.0, 0.01, 0.24}
\definecolor{applegreen}{rgb}{0.55, 0.71, 0.0}
\definecolor{asparagus}{rgb}{0.53, 0.66, 0.42}
\newcommand{\RETURN}{\STATE \textbf{return} }
\newcommand{\cmark}{\textcolor{applegreen}{\scalebox{0.8}{\ding{51}}}}
\newcommand{\xmark}{\textcolor{amaranth}{\scalebox{0.8}{\ding{55}}}}
\newcommand{\txtdag}{\textsuperscript{\textdagger}}

\usepackage{listings}
\definecolor{codegreen}{rgb}{0,0.6,0}
\definecolor{codegray}{rgb}{0.5,0.5,0.5}
\definecolor{codepurple}{rgb}{0.58,0,0.82}
\definecolor{backcolour}{rgb}{0.95,0.95,0.92}

\lstdefinestyle{Python}{
    backgroundcolor=\color{backcolour},
    commentstyle=\color{codegreen},
    keywordstyle=\color{magenta},
    numberstyle=\tiny\color{codegray},
    stringstyle=\color{codepurple},
    basicstyle=\ttfamily\footnotesize,
    breakatwhitespace=false,
    breaklines=true,
    captionpos=b,
    keepspaces=true,
    numbers=left,
    numbersep=5pt,
    showspaces=false,
    showstringspaces=false,
    showtabs=false,
    tabsize=4,
    frame=single,
    framesep=3pt,
    rulecolor=\color{black},
    rulesepcolor=\color{gray},
    xleftmargin=10pt,
    xrightmargin=10pt
}
\title{\raisebox{-1.ex}{\includegraphics[width=1cm]{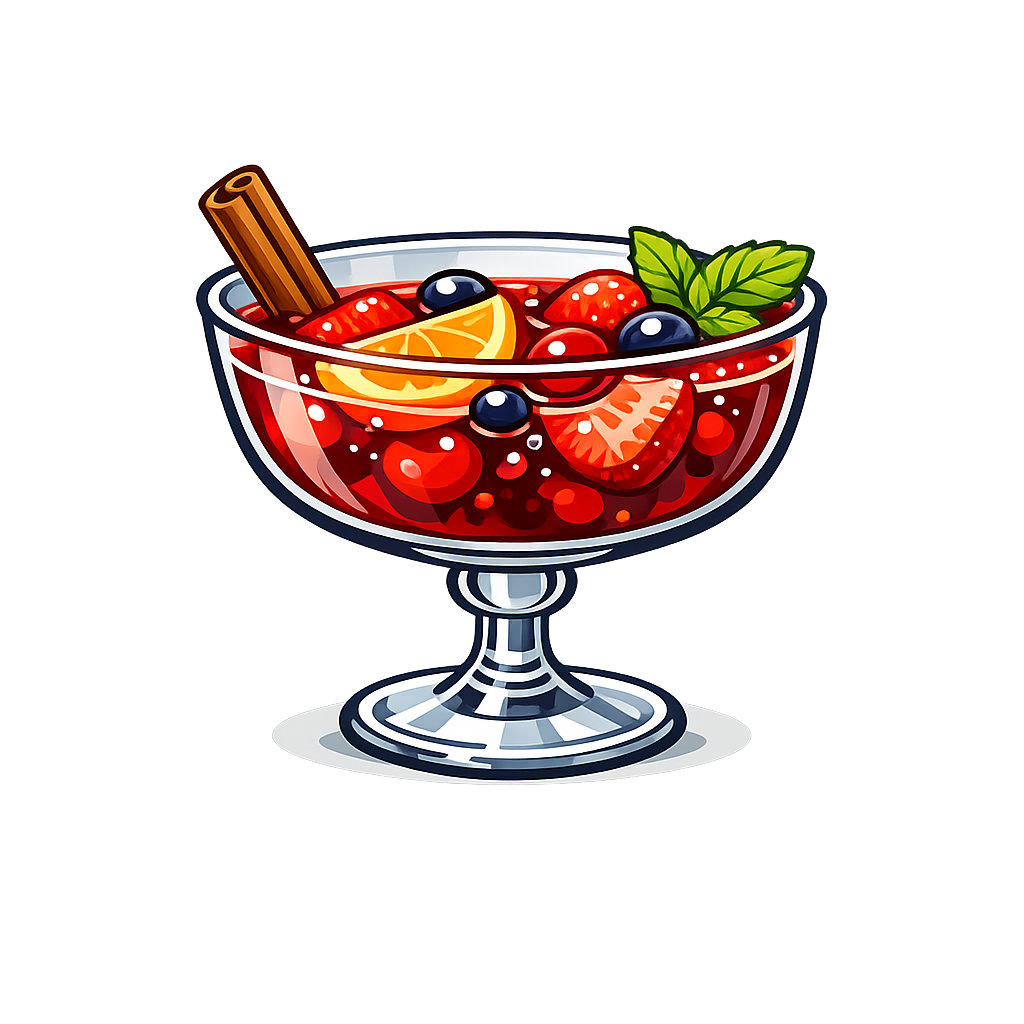}}COMPOT: Calibration-Optimized Matrix Procrustes Orthogonalization for Transformers Compression}

\author{%
  Denis Makhov$^{1,2}$, 
  Dmitriy Shopkhoev$^{1}$, 
  Magauiya Zhussip$^{1,2}$, \\
  \textbf{Ammar Ali$^{1,2}$,}
  \textbf{Baher Mohammad$^{1,2}$,}
  \textbf{Stamatios Lefkimmiatis$^{1,2}$} \\
  $^{1}$MWS AI $\quad$
  $^{2}$ITMO \\
  \texttt{makhovds@gmail.com}\\
  \texttt{\url{https://github.com/mts-ai/COMPOT}}
}

\begin{document}

\maketitle

\begin{abstract}
Post-training compression of transformer models commonly relies on truncated singular value decomposition (SVD) strategies leading to low-rank approximations of the original model weights. However, the underlying single shared subspace modeling can degrade accuracy even at moderate compression. Sparse dictionary learning provides a more flexible union-of-subspaces representation, but existing approaches often suffer from costly iterative dictionary and coefficient updates. We propose \textbf{COMPOT} (\textbf{C}alibration-\textbf{O}ptimized \textbf{M}atrix \textbf{P}rocrustes \textbf{O}rthogonalization for \textbf{T}ransformers), a training-free compression framework that uses a small calibration dataset to estimate a sparse weight factorization.
COMPOT employs orthogonal dictionaries that enable closed-form Procrustes updates for the dictionary and analytical single-step sparse coding for the coefficients, reducing significantly the overall optimization cost.
To handle heterogeneous layer sensitivity under a global compression budget, COMPOT further introduces a one-shot dynamic allocation strategy that adaptively redistributes layer-wise compression rates.
Extensive experiments across diverse architectures and tasks show that COMPOT consistently delivers a superior quality–compression trade-off over strong low-rank and sparse baselines, while remaining fully compatible with post-training quantization for extreme compression.
\end{abstract} \vspace{-0.3cm}
\vspace{-0.3cm}\section{Introduction}\vspace{-0.3cm}
\label{sec:intro}

Transformer-based foundation models underpin state-of-the-art systems across different modalities, including language, vision, vision-language, and audio \citep{vaswani2017attention,dosovitskiy2020image,radford2021learning,gong21b_interspeech}. As they scale to billions of parameters, deployment is increasingly limited by memory footprint, bandwidth, and compute \citep{brown2020language,chowdhery2023palm}. Yet transformers are highly redundant: attention heads and other components can often be removed or simplified with limited degradation, indicating that parameters contribute unevenly to downstream behavior \citep{michel2019sixteen,voita2019analyzing}. This has motivated extensive post-training compression work: quantization, pruning,
and factorization -- to reduce inference cost and memory footprint without expensive retraining \citep{tang2024survey,liu2025survey}.

Low-rank matrix factorization is a widely used post-training paradigm, replacing dense projection matrices by products of smaller matrices. Activation-aware and task-aware low-rank methods show that calibration/activation information is crucial for preserving accuracy \citep{hsu2022fwswd,yuan2023asvd}. Recent SVD-based methods further tailor truncation using calibration and principled losses: SVD-LLM uses whitening-based calibration to relate singular values to truncation loss and applies closed-form updates to mitigate degradation \citep{svdllm}. Despite these advances, SVD methods still enforce a \emph{single shared subspace} per weight matrix, which can be restrictive when different matrix columns lie in different local subspaces~\citep{shopkhoev2025cospadi}.

Sparse dictionary learning offers an alternative strategy, modeling a weight matrix as a dense dictionary times a column-sparse coefficient matrix, enabling a \emph{union-of-subspaces} representation. CoSpaDi shows that calibration-guided sparse dictionary learning can outperform data-aware low-rank baselines for training-free LLM compression \citep{shopkhoev2025cospadi}. Relatedly, MASA proposes matrix-based dictionary learning for parameter sharing in transformer attention, representing layer-specific projections as combinations of shared matrix atoms \citep{zhussip2025shareattention}. 
However, existing dictionary learning pipelines typically rely on iterative and expensive dictionary and sparse coding updates (e.g., K-SVD/OMP \citep{aharon2006ksvd}), limiting practicality at billion-parameter scale.

\begin{figure*}[t]
  \begin{center}
    \centerline{\includegraphics[width=\textwidth]{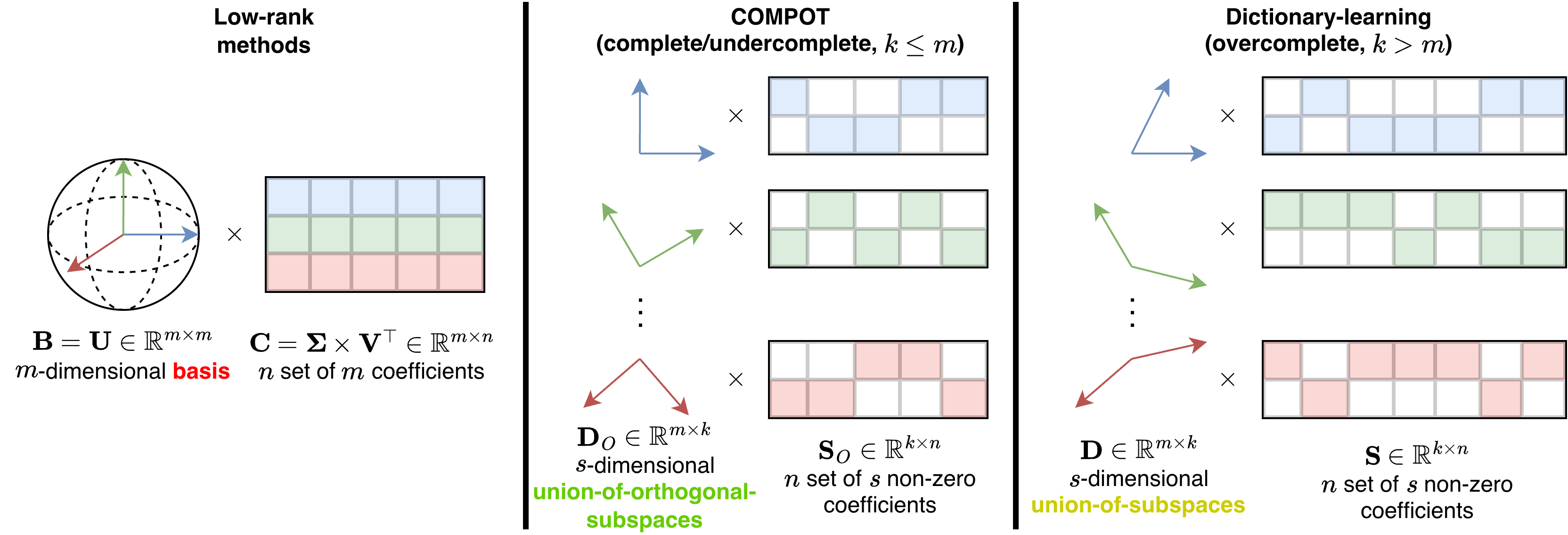}}
    \caption{
      Comparison of low-rank decomposition, sparse dictionary learning, and COMPOT. Low-rank uses a rigid \emph{shared orthogonal basis} $\m B$; dictionary learning enables a flexible \emph{union-of-subspaces}; COMPOT lies in between by using a \emph{union-of-\textbf{orthogonal}-subspaces} (denoted with $_O$), enabling  fast closed-form dictionary and coefficient updates, while improving compression performance.
    }
    \label{fig:low_rank_compot_dl}
  \end{center}
      \vspace{-1.cm}
\end{figure*}

In this work, we introduce \textbf{COMPOT} (\textbf{C}alibration-\textbf{O}ptimized \textbf{M}atrix \textbf{P}rocrustes \textbf{O}rthogonalization for \textbf{T}ransformers), a training-free compression framework that improves calibration-guided dictionary learning in both accuracy and compression time. \textbf{First}, our key design choice is to consider complete/undercomplete dictionaries and enforce \emph{column-orthogonality}, leveraging evidence that orthogonal dictionary learning can be effective and substantially faster than overcomplete alternatives \citep{bao2013fast}. Under orthogonality, the dictionary update transforms to an orthogonal Procrustes problem with a closed-form SVD solution \citep{schoenemann1966procrustes}, replacing atom-wise K-SVD updates. Meanwhile, the sparse coding update also reduces to a closed-form hard-thresholding operation, avoiding iterative pursuits and yielding an efficient alternating procedure \citep{elad2010sparse}.

Transformer layers and projection types exhibit heterogeneous redundancy. Prior work addresses this via non-uniform rank allocation (e.g., SVD-LLM V2) \citep{svdllmv2} or differentiable truncation optimization (Dobi-SVD) \citep{dobisvd}. \textbf{Second}, we instead propose a \emph{one-shot} global allocation strategy: we normalize each weight matrix, pool singular values across matrices into a global spectrum, and truncate globally to satisfy a model-wide budget, while enforcing constraints that prevent negative or low per-matrix compression and cap over-compression of sensitive layers. This yields a simple, deterministic allocation procedure without iterative search over layer-wise ratios.

\textbf{Third}, COMPOT integrates naturally with post-training quantization. Among PTQ methods such as SmoothQuant \citep{xiao2023smoothquant}, GPTQ \citep{frantar2022gptq}, and AWQ \citep{lin2023awq}, we show that 4-bit GPTQ applied on top of COMPOT yields superior results compared to quantization alone under equal memory budgets. \textbf{Fourth}, COMPOT outperforms strong SVD baselines by a wide margin across architectures and downstream tasks (language, audio, vision).

Overall, the proposed method advances the state of the art among structured matrix-factorization compression techniques, with closed-form alternating updates supported by theoretical convergence guarantees, shifting the paradigm beyond established SVD-based approaches.
\vspace{-0.3cm}\section{Related Work}\vspace{-0.3cm}

\subsection{Overview of transformer-based Model Compression}\vspace{-0.2cm}
Compression of transformer-based models spans several complementary families. Early \emph{pruning} work showed that many attention heads can be removed with little degradation, motivating structured pruning of self-attention and encoder-decoder heads to reduce compute \citep{michel2019sixteen,voita2019analyzing}. \emph{Post-training quantization} (PTQ) is now the dominant deployment tool for LLMs: SmoothQuant enables near-lossless W8A8 quantization by migrating activation outliers into weights via offline rescaling \citep{xiao2023smoothquant}; GPTQ/OPTQ use second-order, block-wise weight-only PTQ to reach 3-4 bits \citep{frantar2022gptq}; and AWQ improves low-bit PTQ by identifying and protecting activation-salient weights \citep{lin2023awq}. Orthogonal to quantization, \emph{low-rank and structured factorizations} replace dense projections with products of thinner matrices; recent SVD-based methods (SVD-LLM, SVD-LLM V2) combine truncation-aware whitening, closed-form updates, and optimized rank allocation to better align truncation with functional loss \citep{svdllm,svdllmv2}. Finally, \emph{knowledge distillation} trains a smaller student to match a larger teacher; DistilBERT reduces BERT parameters by 40\% while preserving most downstream accuracy \citep{sanh2019distilbert}. 

\vspace{-0.3cm}\subsection{SVD-based Matrix Factorization for Compression}\vspace{-0.3cm}
Low-rank factorization is a standard post-training approach for compressing transformer projections, motivated by the empirical low-rank structure of weight matrices. While truncated SVD is optimal in Frobenius norm, this objective can be misaligned with preserving task behavior. Early data-aware formulations address this mismatch: DRONE derives a provably optimal low-rank decomposition with closed-form solutions for feed-forward and self-attention matrices in BERT-style models \citep{chen2021drone}, and FWSVD improves robustness by Fisher-weighting reconstruction error \citep{hsu2022fwswd}. Subsequent work incorporates activation information: ASVD leverages activation statistics and layer sensitivity to guide training-free truncation \citep{yuan2023asvd}. More recently, SVD-LLM introduces truncation-aware whitening and closed-form parameter updates to mitigate degradation at higher compression \citep{svdllm}, Dobi-SVD proposes differentiable truncation with a remapping strategy to better connect truncation choices to storage behavior \citep{dobisvd}, and SVD-LLM V2 further refines truncation-loss modeling and allocates non-uniform compression ratios across matrices to reflect heterogeneous redundancy \citep{svdllmv2}. \vspace{-.1cm}

Matrix compression using truncated SVD essentially amounts to applying Principal Component Analysis (PCA) to the matrix columns: it finds the best $L_2$-norm subspace to project the data, creating an optimal low-rank approximation~\cite{bishop2006pattern}. However, this approach forces the entire matrix into one shared subspace. This can be too restrictive if different columns would be better represented by their own unique, local subspaces.

\vspace{-0.3cm}\subsection{Dictionary Learning and Sparse Coding}\vspace{-0.3cm}
Sparse dictionary learning is a complementary paradigm: it represents a weight matrix as the product of a dictionary and a sparse coefficient matrix. This formulation enables different columns to select distinct subsets of atoms, thereby inducing a \emph{union-of-subspaces} structure.  Classic methods such as K-SVD iteratively update dictionary atoms and are widely used in signal and image processing \citep{aharon2006ksvd}. CoSpaDi adapts calibration-guided dictionary learning to LLM compression by minimizing functional output mismatch rather than pure weight error, and shows that structured sparse factorizations can outperform data-aware SVD baselines at practical compression ratios \citep{shopkhoev2025cospadi}. However, K-SVD-style updates are computationally heavy at billion-model scale, motivating alternative constraints and update rules. To address this challenge, our method restricts the weight factorization to employ complete or undercomplete orthogonal dictionaries, depending on the matrix shape and storage budget. This structural constraint substantially simplifies optimization: the dictionary update reduces to a closed-form orthogonal Procrustes solution, while sparse coding admits an analytical solution, entirely eliminating the need for iterative sparse pursuit algorithms. Our approach is well motivated, as the orthogonality constraint on learned dictionaries has been investigated in image processing and shown to improve both numerical stability and computational efficiency \citep{bao2013fast}.\vspace{-.1cm}

\vspace{-0.3cm}\subsection{Dynamic Allocation of Compression Ratios}\vspace{-0.3cm}
A recurring theme in transformer compression is heterogeneous redundancy across layers and projection types, making uniform compression suboptimal. ASVD highlights sensitivity differences and uses activation-aware heuristics to reduce degradation \citep{yuan2023asvd}. SVD-LLM V2 assigns matrix-specific compression ratios via truncation-loss modeling under a global budget \citep{svdllmv2}. Dobi-SVD targets rank selection with differentiable truncation and introduces a remapping that better connects truncation to achievable compression \citep{dobisvd}. Recent work continues this direction: D-Rank allocates ranks dynamically under a fixed budget via optimization \citep{mi2025layerwisedynamicrankcompressing}, and ARA introduces adaptive rank allocation for SVD-based compression \citep{xv2025araadaptiverankallocation}. These results reinforce that \emph{budget allocation} is central to minimizing quality drop for a given compression target. In this work, we contribute a simple one-shot allocation procedure that operates over pooled singular values while enforcing constraints that prevent negative compression and limit over-compression of individual matrices.\vspace{-.2cm}
\vspace{-0.3cm}\section{Method}\vspace{-0.3cm}
\label{sec:method}

\begin{figure*}[ht]
  % \vskip 0.2in
  \begin{center}
    \centerline{\includegraphics[width=\textwidth]{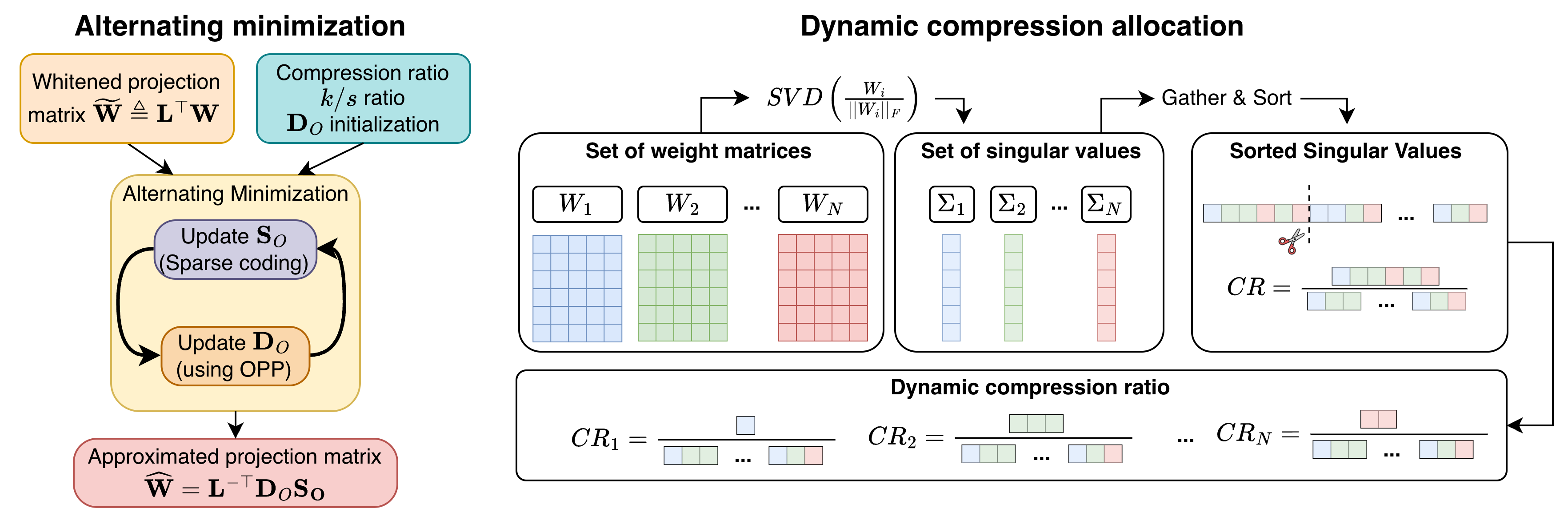}}
    \caption{
      Overview of COMPOT framework. On the \textit{left} part the alternating minimization process is visualized while on the \textit{right} we represent our single-shot strategy for dynamic compression ratio allocation based on singular values of normalized projection matrices.
    }
    \label{fig:compot_overview}
    \vspace{-0.7cm}
  \end{center}
\end{figure*}

\subsection{Problem Setup}\vspace{-0.2cm}
\label{sec:problem_setup}
We consider a pretrained transformer with linear projections parameterized by weight matrices
$\m W \in \R^{m \times n}$ (e.g., attention and MLP projections). Our goal is \emph{post-training}
compression: replace each $\m W$ with a structured factorization that reduces the model parameter count while preserving the layer behavior on a small calibration set, without back-propagation (healing/fine-tuning).

Let $\m X \in \R^{N \times m}$ denote a matrix of calibration inputs (layer activations entering the
projection), where $N$ is the number of calibration tokens/examples aggregated into rows. The layer output
on calibration data is $\m Y=\m X \m W\in\R^{N\times n}$. A natural data-aware objective is therefore to minimize output mismatch
$\|\m X ( \m W - \widehat{\m W})\|_F^2$ over compressed $\widehat{\m W}$.\vspace{-.1cm}

\subsection{Preliminaries: Subspace vs.\ Union-of-subspaces Modeling}\vspace{-.1cm}
\label{sec:preliminaries}

\textbf{Low-rank (single shared subspace).}
For a real-valued matrix $\m W \in \R^{m \times n}$, truncated SVD gives
\bal
\m W \approx \m U_r \m \Sigma_r \m V_r^\top,
\eal
where $\m {U}_r \in \R^{m \times r}$, $\m {\Sigma}_r \in \R^{r \times r}$, and $\m {V}_r \in \R^{n \times r}$.
From the direct link with PCA and the underlying concept of projecting columns of $\m W$ to a lower-dimensional subspace~\citep{bishop2006pattern} we have:
\bal
\m W \approx \m B \m C, \ \text{where} \
\m B \triangleq \m {U}_r,\
\m C \triangleq \m {\Sigma}_r \m V_r^\top,
\label{eq:svd_two_factor}
\eal
so each column $\m w_j$ is represented in the \emph{same} $r$-dimensional subspace spanned by $\m B$ (Fig.~\ref{fig:low_rank_compot_dl} (left)).

\textbf{Sparse dictionary learning (union-of-subspaces).}
Dictionary learning (Fig.~\ref{fig:low_rank_compot_dl} (right)) represents the projection matrix with an \textit{overcomplete} dictionary $\m D$ and corresponding sparse coefficient $\m S$
\begin{equation}
\m W \approx \m D \m S,
\qquad
\m D \in \R^{m \times k},
\quad
\m S \in \R^{k \times n},
\end{equation}
where $\m S$ is column-sparse: each column $\m s_j$ has at most $s$ nonzeros. As a result, each column $\m w_j$ may be represented with 
a different subset of atoms, inducing a \emph{union-of-subspaces} model: columns can live in different $s$-dimensional subspaces of $\R^m$ spanned by subsets of atoms of $\m D$.\vspace{-.1cm}

\subsection{COMPOT: Calibration-optimized Orthogonal Dictionary Factorization}\vspace{-.1cm}
\label{sec:compot}

COMPOT preserves the union-of-subspaces flexibility of sparse coding, but enforces an \emph{orthogonal} (complete/undercomplete) dictionary in a \emph{whitened, data-aware} space (Fig.~\ref{fig:low_rank_compot_dl} (middle)). This yields closed-form updates (Fig.~\ref{fig:compot_overview} (left)): Procrustes (thin SVD) for the dictionary and an analytical sparse coding step under orthogonality, resulting in a \textit{union-of-orthogonal-subspaces} model.

\textbf{Data-aware whitening.}
We adopt a standard calibration objective that measures functional error on calibration activations:
\begin{equation}
\min_{\widehat{\m W}} \ \|\m X(\m W-\widehat{\m W})\|_F^2.
\label{eq:functional_loss}
\end{equation}
First we define the Gram matrix, $
\m G \triangleq \m X^\top \m X \in \R^{m \times m}$, which we assume to be positive definite (this condition is typically satisfied using sufficient calibration data)  and admits the Cholesky decomposition
$\m G = \m L \m L^\top$, with $\m L \in \R^{m \times m}$ the Cholesky factor. Then it holds:
\bal
\|\m X(\m W-\widehat{\m W})\|_F^2 &= \mathrm{Tr}\!\left((\m W-\widehat{\m W})^\top \m G (\m W-\widehat{\m W})\right)
=\|\m L^\top(\m W-\widehat{\m W})\|_F^2.
\label{eq:whiten_equiv}
\eal
Therefore, optimizing the functional error is equivalent to optimizing the reconstruction error in the \emph{whitened space}.
We define the whitened weight matrix
\begin{equation}
\widetilde{\m W} \triangleq \m L^\top \m W \in \R^{m \times n}.
\end{equation}\vspace{-0.5cm}

\textbf{COMPOT objective.}
COMPOT factorizes $\widetilde{\m W}$ using an \emph{orthogonal} (complete or undercomplete) dictionary $\m D_O$ and
column-sparse codes $\m S_O$:

\bal
\m D_O^\star,\m S_O^\star
=
\arg\min_{\m D_O,\m S_O}
\ \|\widetilde{\m W}-\m D_O \m S_O \|_F^2 
\quad\mbox{s.t.}\quad 
\m D_O^\top \m D_O=\m I_k,\ \|\m {s_O}_j\|_0 \le s \ \forall j \in [n],
\label{eq:compot_main}
\eal
where $\m D_O \in \R^{m \times k}$ with $k \le m$ (complete/undercomplete), and $\m S_O \in \R^{k \times n}$.
After solving \eqref{eq:compot_main}, we map the factorization back to the original parameter space:
\begin{equation}
\m W \approx \widehat{\m W} \triangleq \m A \m S_O,
\qquad
\m A \triangleq \m L^{-\top}\m D_O \in \R^{m \times k}.
\label{eq:dewhiten}
\end{equation}
Importantly, $\m A$ is computed \emph{offline} during compression, so inference uses only $(\m A,\m S_O)$.

\textbf{Alternating minimization (Sparse coding + Procrustes).}
Although joint minimization of \eqref{eq:compot_main} is not feasible, we can still opt for a simple alternating minimization with closed-form updates when enforcing $\m D_O^\top \m D_O=\m I_k$.
With $\m D_O$ fixed, sparse coding decouples column-wise and has an analytical solution~\citep{elad2010sparse}:
\begin{equation}
\m S_O \leftarrow \mathcal{H}_s\!\big(\m D_O^\top \widetilde{\m W}\big),
\label{eq:compot_sparse_closed_form_main}
\end{equation}
where $\mathcal{H}_s(\cdot)$ (hard-thresholding operator) keeps the $s$ largest-magnitude entries in each column and zeroes out the rest.
With $\m S_O$ fixed, the dictionary update is an orthogonal Procrustes step \citep{schoenemann1966procrustes}:
\begin{equation}
\m M \triangleq \widetilde{\m W}\m S_O^\top,\quad 
\m M \overset{\text{SVD}}{=} \m P\m\Lambda\m Q^\top\
\ \Rightarrow\ 
\m D_O \leftarrow \m P\m Q^\top.
\label{eq:compot_procrustes_main}
\end{equation}
We provide the full derivation of \eqref{eq:compot_sparse_closed_form_main}, the complete procedure, implementation details and convergence analysis in \ref{apx:compot_algo}.

\textbf{One-shot compression allocation.}
Different transformer projections exhibit heterogeneous redundancy, so uniform compression is often suboptimal.
We propose a simple one-shot global allocation strategy that distributes a model-wide budget by globally truncating
the smallest singular values across matrices, while enforcing per-matrix compression guards and handling matrices
where factorization is not beneficial.

\textit{Original or whitened space?}
Whitening is layer- and projection-specific (depends on $\m X$), so singular values computed in whitened coordinates
are not directly comparable across matrices. We therefore allocate ranks using singular values of the \emph{raw} weight
matrices in the original space, while still using whitening for COMPOT reconstruction.

\textit{Normalize or not?}
Even in the original space, spectra magnitudes vary across layers. To address this, we normalize each matrix by its Frobenius norm: $\m W_i^{(f)} \triangleq \m W_i/\norm{\m W_i}{F}.$

Following normalization, we compute a thin SVD for each $\m W_i^{(f)}$ and pool all singular values into a global multiset. We sort this pool and truncate the smallest values to satisfy the desired model-wide compression ratio; the number truncated per matrix then induces its layer-wise compression ratio. To prevent degenerate behavior, we enforce minimum and maximum compression guards and handle matrices where factorization is not beneficial. Fig.~\ref{fig:compot_overview} (right) illustrates the simplified procedure; ~\ref{apx:dynamic_alloc} provides full details.

\textbf{Compression ratio (CR).}
In the proposed method we need to store both $\m A$ (unwhitened dictionary) and $\m S_O$ with 16-bit values. Instead of storing the sparse matrix directly, we store only its nonzero elements and a binary mask of their position:
\begin{equation}
    \text{CR}_{\mathrm{COMPOT}} = 1-\tfrac{\;\overbrace{16mk}^{\m A}+\overbrace{16sn}^{\m S_O}+\overbrace{(kn)}^{\text{Mask}}\;}{16mn}.
    \label{eq:compression_ratio}
\end{equation}
For the $\text{CR}_{\mathrm{COMPOT}}$ computation we need to specify both the number of atoms \(k\) and the number of column-wise non-zero elements \(s\). We express both parameters with a single hyperparameter \(k/s\)-ratio.
\vspace{-0.3cm}\section{Experiments}\vspace{-0.3cm}
\label{sec:experiments}

We evaluate COMPOT across model families (Llama, OPT, Qwen; 0.6B--30B) and domains (language, vision-language, audio), under compression ratios (CR) $0.2$--$0.6$. We compare against strong SVD-based baselines (SVD-LLM~\citep{svdllm}, SVD-LLM V2~\citep{svdllmv2}, Dobi-SVD~\citep{dobisvd}), sparse dictionary learning (CoSpaDi~\citep{shopkhoev2025cospadi}), and structured pruning (ReplaceMe~\citep{replaceme}, LLM-Pruner~\citep{llmpruner}), and study compatibility with post-training quantization (GPTQ~\citep{frantar2022gptq}). Hardware setup and reproducibility analysis are presented in~\ref{apx:hardware_setup} and~\ref{apx:seed_ablation}, respectively.

\vspace{-0.3cm}\subsection{Experimental Setup}\vspace{-0.3cm}
Unless stated otherwise, we set the dictionary-to-sparsity ratio to $k/s=2$ (see~\ref{apx:ks_ratio}) -- adjusted only when necessary to avoid overcomplete dictionaries that violate the orthogonality constraint -- and compress all dense linear projections in self-attention (Q/K/V/O) and the MLP (gate/up/down), keeping token embeddings and \texttt{lm\_head} uncompressed in alignment with prior works. We report zero-shot accuracy on PIQA \citep{bisk2020piqa}, HellaSwag \citep{zellers2019hellaswag}, OpenAI LAMBADA \citep{paperno2016lambada}, ARC-easy/ARC-challenge \citep{clark2018think}, SciQ \citep{welbl2017crowdsourcing}, RACE \citep{lai2017race}, and MMLU \citep{hendryckstest2021}, and perplexity on WikiText \citep{merity2017pointer} and LAMBADA-OpenAI. For SVD-LLM, CoSpaDi, and pruning baselines we use \texttt{lm-evaluation-harness}~0.4.8 \citep{eval-harness} with normalized accuracies when available; for SVD-LLM V2 and Dobi-SVD comparisons we additionally follow the original SVD-LLM evaluation protocol (unnormalized accuracies and the authors' perplexity script). Additional benchmarks/ablations are reported in ~\ref{apx:more_benchs}.

\vspace{-0.3cm}\subsection{Ablation Study}\vspace{-0.3cm}
COMPOT involves several hyperparameters that must be chosen carefully, in particular the dictionary initialization, the number of alternating minimization steps, the strategy for dynamic allocation and the choice of calibration data. In this section we study their impact on performance. Unless stated otherwise, ablations report average accuracy (Avg. Acc.) over the datasets in Table~\ref{tab:main_static}, together with WikiText and LAMBADA perplexity (Wiki. PPL and LAMB. PPL, correspondingly).

\begin{center}
\centering
% First table (left)
\begin{minipage}{0.5\textwidth}
    \centering
    \captionof{table}{Effect of dictionary initialization for Llama3.2-1B at $0.2$ compression with 20 alternating minimization iterations. Best results are shown in \textbf{bold}.}
    \label{tab:abl_init}
    \resizebox{\textwidth}{!}{%
    \renewcommand{\arraystretch}{1.05}
    \begin{tabular}{ccccc}
        \hline
        \textbf{CR Allocation}            & \textbf{Init.} & \textbf{Avg. Acc.$\uparrow$} & \textbf{Wiki. PPL$\downarrow$} & \textbf{LAMB. PPL$\downarrow$} \\ \hline
        \multirow{2}{*}{Static}  & Rand. & 46.3      & 26.84     & 24.74       \\
                                 & SVD   & \textbf{48.5}      & \textbf{24.67}     & \textbf{20.62}       \\ \hline
        \multirow{2}{*}{Dynamic} & Rand. & 48.5      & 22.05     & 15.06       \\
                                 & SVD   & \textbf{50.1}      & \textbf{20.71}     & \textbf{13.97}       \\ \hline
    \end{tabular}
    }
    \end{minipage}%
    \hfill
    \begin{minipage}{0.47\textwidth}
    \centering
    \captionof{table}{Effect of grouping for dynamic allocation for Llama3.2-1B at $0.2$ compression with 20 alternating minimization iterations. Best results are shown in \textbf{bold}.}
    \label{tab:abl_group}
    \resizebox{\textwidth}{!}{%
    \renewcommand{\arraystretch}{1.05}
    \begin{tabular}{cccc}
    \hline
    \textbf{Grouping}    & \textbf{Avg. Acc.$\uparrow$} & \textbf{Wiki. PPL$\downarrow$} & \textbf{LAMB. PPL$\downarrow$} \\ \hline
    All indiv.  & 48.5      & 24.62     & 18.62       \\
    QKV\&UpGate & 49.2      & 21.41     & 15.52       \\ 
    All grouped & \textbf{50.1}      & \textbf{20.71}     & \textbf{13.97}       \\ \hline
    \end{tabular}
    }
    \end{minipage}
\end{center}

\textbf{Dictionary initialization.}
We compare random-column initialization (randomly orthonormalized permuted subset of $\widetilde{\m W}$ columns) against SVD initialization (top left singular vectors). Table~\ref{tab:abl_init} shows that SVD initialization consistently improves performance at fixed iteration budget.
\begin{center}
  \begin{minipage}{0.5\textwidth}
    \centering
    \includegraphics[width=0.9\linewidth]{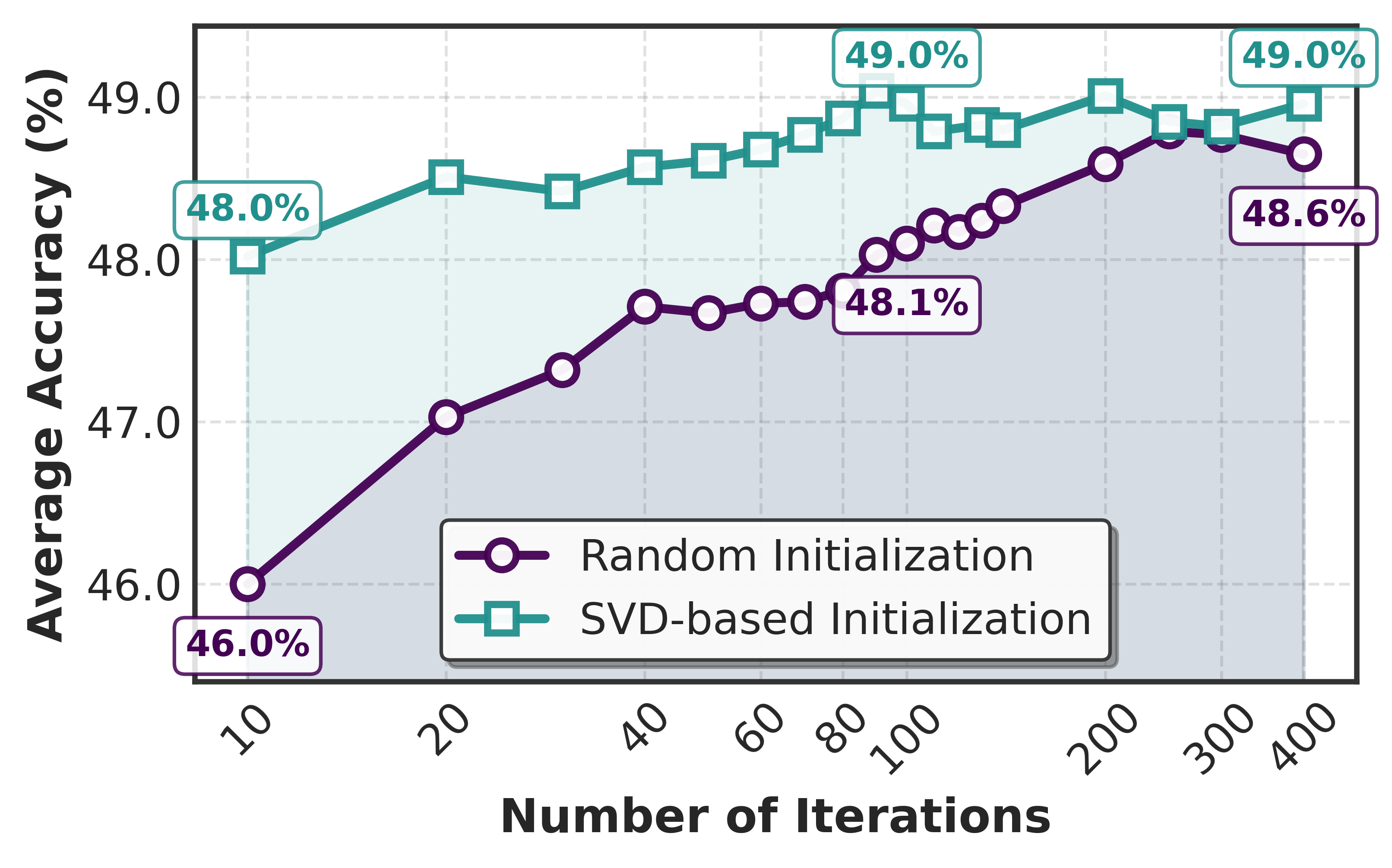}
    \captionof{figure}{Average accuracy as a function of the number of alternating minimization steps on Llama3.2-1B at $0.2$ compression, comparing random and SVD-based dictionary initialization.}
    \label{fig:abl_num_iters}
  \end{minipage}%
  \hfill
  \begin{minipage}{0.45\textwidth}
    \centering
     \captionof{table}{Calibration data effect on COMPOT with dynamic allocation and SVD initialization. Best results are shown in \textbf{bold}.}
    \label{tab:abl_data}
    \resizebox{\textwidth}{!}{%
    \renewcommand{\arraystretch}{1.05}
    \begin{tabular}{ccccc}
    \hline
    \multicolumn{2}{c}{}             & \textbf{Avg. Acc.$\uparrow$} & \textbf{Wiki. PPL$\downarrow$} & \textbf{LAMB. PPL$\downarrow$} \\ \hline \hline %\cline{3-5}
    \multicolumn{2}{c}{\textbf{Llama3.2 1B}}  & 57.6      & 11.57     & 5.73     \\ \hline
    \multirow{3}{*}{CR 0.2} & WikiText  & 50.0      & \textbf{18.52}     & \textbf{13.15}    \\
                            & Alpaca    & \textbf{50.1}      & 23.58     & 18.33    \\
                            & FineWeb& \textbf{50.1}      & 20.71     & 13.97    \\ \hdashline
    \multirow{3}{*}{CR 0.3} & WikiText  & 44.9      & \textbf{31.12}     & 32.97    \\
                            & Alpaca    & 44.8      & 51.28     & 82.64    \\
                            & FineWeb& \textbf{45.0}      & 35.68     & \textbf{32.35}    \\ \hdashline
    \multirow{3}{*}{CR 0.4} & WikiText  & 36.4      & \textbf{109.11}    & 299.72   \\
                            & Alpaca    & 35.3      & 447.37    & 2857.45   \\
                            & FineWeb& \textbf{37.6}      & 129.28    & \textbf{195.41}   \\ \hline
    \end{tabular}
    }
  \end{minipage}
  \vspace{-0.2cm}
\end{center}

\textbf{Number of alternating updates.}
We vary the number of alternating updates on Llama3.2-1B at static CR $0.2$ (Fig.~\ref{fig:abl_num_iters}). Random initialization improves up to $\sim$300 iterations, while SVD initialization saturates much earlier ($\sim$100). We use 20 iterations in all main experiments as a pragmatic accuracy--time compromise.

\textbf{Grouping for dynamic allocation.}
We test pooling singular values by projection type (All indiv. similar to SVD-LLM V2), partial grouping (QKV\&UpGate), and a single global pool (All grouped) on Llama3.2-1B at CR $0.2$. Table~\ref{tab:abl_group} shows global pooling yields the best overall accuracy/perplexity, motivating our default global allocation.

\textbf{Calibration data influence.}
We investigated the influence of calibration data used for Gram matrix estimation. We sampled 256 random samples of length 1024 from WikiText~\citep{wikitext}, FineWeb~\citep{fineweb} and Alpaca~\citep{alpaca} datasets. 
The ablation study in Table~\ref{tab:abl_data} reveals that FineWeb provides a reasonable compromise across all metrics, prompting its use in our further experiments.
We additionally study sensitivity to the number of calibration samples in Appendix~\ref{apx:seed_ablation}.
\vspace{-.3cm}

\subsection{Main Results}
\vspace{-.2cm}
\textbf{Uniform CR (no allocation).}
We first disable dynamic allocation and apply a uniform CR to all projections (COMPOT\txtdag). We compare against SVD-LLM \citep{svdllm} as a strong SVD-based baseline and CoSpaDi \citep{shopkhoev2025cospadi} as a K-SVD-based sparse dictionary method. Calibration uses 256 sequences from FineWeb \citep{fineweb} (length 1024). Table~\ref{tab:main_static} reports results on Llama3-8B and Qwen3-8B; smaller models (Llama3.2-1B, Qwen3-0.6B) are provided in ~\ref{apx:small_models}. Across CRs, COMPOT\txtdag yields higher average accuracy and lower perplexity than both low-rank (SVD-LLM) and K-SVD-style dictionary learning (CoSpaDi), indicating that orthogonal dictionary factorization improves the quality--compression trade-off under a fixed per-layer budget.\vspace{-.2cm}

%%%%%%%%%%%%%%%%%%%%%%%%%%%%%%%%%%%%%%%%%%%%%%%%%%%%%%%
\begin{table*}[t]
\caption{Performance comparison under \textit{static} CR allocation of COMPOT\txtdag vs state-of-the-art SVD-based SVD-LLM~\citep{svdllm} and dictionary-learning CoSpaDi~\citep{shopkhoev2025cospadi} on Llama3-8B and Qwen3-8B at different compression levels on different benchmarks. Best results are shown in \textbf{bold}.}
\label{tab:main_static}
\centering
\resizebox{0.95\textwidth}{!}{%
\renewcommand{\arraystretch}{1.05}
\begin{tabular}{ccccccccccc:cc}
\hline
                            &                         & \multicolumn{9}{c}{\textbf{Accuracy$\uparrow$}}                                                                           & \multicolumn{2}{c}{\textbf{Perplexity$\downarrow$}}       \\ \cline{3-13}
\multirow{-2}{*}{\textbf{Method}}    & \multirow{-2}{*}{\textbf{CR}}    & \textbf{PIQA} & \textbf{Hella Swag} & \textbf{LAMBADA} & \textbf{ARC-e} & \textbf{ARC-c} & \textbf{SciQ} & \textbf{Race}& \textbf{MMLU} & \textbf{Avg.} & \textbf{WikiText} & \textbf{LAMBADA} \\ \hline \hline

\textbf{Llama3 8B}& --                    & 80.7 & 79.1 & 75.6 & 77.7 & 53.5 & 93.9 & 40.3 & 62.2 & 70.4 & 7.3E+00 & 3.1E+00 \\ \hline

SVD-LLM       &                       & 71.1 & 58.4 & 59.3 & 55.5 & 34.0 & 86.4 & 35.5 & 32.6 & 54.1 & 4.1E+01 & 1.1E+01 \\
CoSpaDi       &                       & 75.2 & 66.5 & 73.8 & 66.5 & 41.6 & 89.5 & 38.2 & 42.8 & 61.8 & 2.0E+01 & \textbf{4.3E+00} \\
COMPOT\txtdag & \multirow{-3}{*}{0.2} & 77.8 & 72.7 & 69.4 & 71.0 & 44.5 & 89.8 & 41.0 & 50.1 & \textbf{64.5} & \textbf{1.3E+01} & 5.2E+00 \\ \hline

SVD-LLM       &                       & 65.8 & 46.4 & 38.1 & 41.9 & 27.7 & 70.0 & 31.8 & 27.2 & 43.6 & 1.5E+02 & 6.1E+01 \\
CoSpaDi       &                       & 70.5 & 56.2 & 61.3 & 54.2 & 33.5 & 85.7 & 36.2 & 32.2 & 53.7 & 4.5E+01 & 9.2E+00 \\
COMPOT\txtdag & \multirow{-3}{*}{0.3} & 74.4 & 64.6 & 60.9 & 61.0 & 38.0 & 87.1 & 38.6 & 38.9 & \textbf{58.0} & \textbf{2.1e+01} & \textbf{8.5E+00} \\ \hline

SVD-LLM       &                       & 60.3 & 34.5 & 11.4 & 32.4 & 24.5 & 44.2 & 25.7 & 23.1 & 32.0 & 5.5E+02 & 1.3E+03 \\
CoSpaDi       &                       & 63.7 & 41.4 & 30.3 & 39.1 & 26.6 & 68.5 & 30.5 & 25.4 & 40.7 & 1.8E+02 & 1.2E+02 \\
COMPOT\txtdag & \multirow{-3}{*}{0.4} & 68.2 & 51.4 & 38.8 & 46.9 & 31.0 & 75.4 & 34.4 & 27.9 & \textbf{46.7} & \textbf{6.2e+01} & \textbf{4.2E+01} \\ \hline \hline

\textbf{Qwen3 8B}          & --       & 77.7 & 74.9 & 64.1 & 80.7 & 56.7 & 95.7 & 40.9 & 73.0 & 70.5 & 1.2E+01 & 4.6E+00 \\ \hline

SVD-LLM       &                       & 73.8 & 63.9 & 62.2 & 68.7 & 45.7 & 90.1 & 40.5 & 54.7 & 62.5 & 2.1E+01 & 6.4E+00 \\
CoSpaDi       &                       & 76.5 & 68.0 & 65.6 & 72.2 & 48.9 & 93.2 & 40.7 & 60.8 & 65.7 & 1.8E+01 & \textbf{4.9E+00} \\
COMPOT\txtdag & \multirow{-3}{*}{0.2} & 75.5 & 72.2 & 62.5 & 70.7 & 49.0 & 92.4 & 42.9 & 66.0 & \textbf{66.4} & \textbf{1.5E+01} & 5.7E+00 \\ \hline

SVD-LLM       &                       & 70.4 & 55.2 & 53.8 & 59.3 & 37.1 & 87.2 & 38.4 & 44.8 & 55.8 & 2.7E+01 & 1.1E+01 \\
CoSpaDi       &                       & 72.4 & 60.5 & 62.6 & 63.9 & 41.2 & 88.4 & 39.5 & 51.3 & 60.0 & 2.3E+01 & 6.3E+00 \\
COMPOT\txtdag & \multirow{-3}{*}{0.3} & 75.0 & 65.9 & 62.5 & 66.3 & 44.2 & 90.1 & 42.1 & 58.2 & \textbf{63.0} & \textbf{1.8E+01} & \textbf{6.2E+00} \\ \hline

SVD-LLM       &                       & 66.3 & 44.6 & 37.9 & 45.0 & 28.1 & 77.3 & 35.3 & 29.1 & 45.4 & 4.3E+01 & 3.6E+01 \\
CoSpaDi       &                       & 68.9 & 49.0 & 49.9 & 49.4 & 29.9 & 82.0 & 36.8 & 36.6 & 50.3 & 3.6E+01 & 1.5E+01 \\
COMPOT\txtdag & \multirow{-3}{*}{0.4} & 70.9 & 56.4 & 54.7 & 53.7 & 34.9 & 84.3 & 38.7 & 44.4 & \textbf{54.7} & \textbf{2.5E+01} & \textbf{9.8E+00} \\ \hline

\end{tabular}
}\vspace{-.4cm}
\end{table*}
%%%%%%%%%%%%%%%%%%%%%%%%%%%%%%%%%%%%%%%%%%%%%%%%%%%%%%%

%%%%%%%%%%%%%%%%%%%%%%%%%%%%%%%%%%%%%%%%%%%%%%%%%%%%%%%
\begin{table*}[h]
\caption{Performance comparison under \textit{dynamic} compression ratio allocation of \textit{training-free} COMPOT vs \textit{training-based} state-of-the-art SVD-based Dobi-SVD \citep{dobisvd} on Llama2-7B at different compression levels on different benchmarks. Best results are shown in \textbf{bold}.}
\label{tab:main_dynamic_dobisvd}
\centering
\resizebox{0.95\textwidth}{!}{%
\renewcommand{\arraystretch}{1.05}
\begin{tabular}{ccccc:cccccccc}
\hline
\multirow{2}{*}{\textbf{Method}} & \multirow{2}{*}{\textbf{Training-free}} & \multirow{2}{*}{\textbf{CR}} & \multicolumn{2}{c}{\textbf{Perplexity$\downarrow$}}  & \multicolumn{8}{c}{\textbf{Accuracy$\uparrow$}}       \\ \cline{4-13}
  & & & \textbf{WikiText-2} & \textbf{C4}   & \textbf{Openb.} & \textbf{ARC\_e} & \textbf{WinoG.} & \textbf{HellaS.} & \textbf{ARC\_c} & \textbf{PIQA} & \textbf{MathQA} & \textbf{Avg.} \\ \hline \hline

\textbf{Llama2-7B}        & \multicolumn{2}{c}{---} & 5.49  & 7.27  & 44.0 & 74.6 & 69.0 & 76.3 & 45.5 & 78.6 & 28.4 & 59.5 \\ \hline

Dobi-SVD\txtdag  & \xmark & \multirow{2}{*}{\textbf{0.2}} & 9.39  & 19.70 & 32.8 & 47.6 & 58.5 & 51.5 & 29.3 & 65.2 & 22.7 & 43.9 \\ %\cline{4-13}
COMPOT           & \cmark &                               & \textbf{6.22} & \textbf{9.34} & 30.4 & 72.6 & 67.6 & 51.8 & 39.6 & 76.4 & 26.2 & \textbf{52.1} \\ \hline
Dobi-SVD\txtdag  & \xmark & \multirow{2}{*}{\textbf{0.4}} & 17.50 & 55.58 & 29.2 & 34.4 & 54.4 & 35.4 & 24.5 & 57.3 & 21.3 & 36.6 \\ %\cline{4-13}
COMPOT           & \cmark &                               & \textbf{8.91} & \textbf{20.40} & 25.2 & 63.1 & 63.5 & 39.6 & 28.5 & 67.0 & 23.6 & \textbf{44.4} \\ \hline
Dobi-SVD\txtdag  & \xmark & \multirow{2}{*}{\textbf{0.6}} & 53.40  & 200.11& 26.0 & 28.3 & 50.5 & 28.4 & 24.7 & 51.9 & 22.0 & \textbf{33.1} \\ % \cline{4-13}
COMPOT           & \cmark &                               & \textbf{32.10} & \textbf{179.02}& 13.4 & 33.8 & 53.4 & 28.8 & 18.8 & 54.9 & 22.1 & 32.2 \\ \hline

\end{tabular}
}\vspace{-.1cm}
\end{table*}
%%%%%%%%%%%%%%%%%%%%%%%%%%%%%%%%%%%%%%%%%%%%%%%%%%%%%%%

\textbf{Dynamic CR allocation.}
We evaluate full COMPOT with our one-shot allocation and compare against methods that explicitly optimize layer-wise ranks, namely Dobi-SVD \citep{dobisvd} and SVD-LLM V2 \citep{svdllmv2}. To ensure a like-for-like comparison, we follow the released SVD-LLM/Dobi-SVD evaluation protocols: we calibrate with 256 WikiText train split samples at context length 2048, report \emph{unnormalized} accuracies via \texttt{lm-eval}~0.4.8, and compute perplexity on WikiText (test split) and C4 using the scripts provided in the SVD-LLM repository.
Results are summarized in Tables~\ref{tab:main_dynamic_dobisvd} and~\ref{tab:main_dynamic_svdllmv2}, with additional Llama-13B/30B results in ~\ref{apx:svd_llm_v2_details}. We also present CR allocation plots in ~\ref{apx:allocation}

For SVD-LLM V2, the paper points to the SVD-LLM repository for code, but at the time of our experiments the repository does not provide a dedicated, ready-to-run V2 implementation; therefore, we implemented the V2 dynamic allocation and loss-optimized truncation strategies on top of the released SVD-LLM codebase and report both our reproduction (denoted with `repr.') and the paper-reported numbers in Table ~\ref{tab:main_dynamic_svdllmv2}.

Discrepancies between published and reproduced results, together with reproduction details, are discussed in~\ref{apx:svd_llm_v2_details}.
For Dobi-SVD, we evaluate the provided checkpoints \emph{without} the ``remapping'' step, which alters the effective storage/parameterization and mixes factorization with quantization; we include the full remapping-inclusive comparison in ~\ref{apx:dobi_details}.
Under matched global CR, COMPOT consistently achieves stronger accuracy/perplexity while remaining fully training-free.\vspace{-.3cm}
\begin{center}
\begin{tabular}{@{}p{0.5\textwidth}@{\hspace{0.03\textwidth}}p{0.47\textwidth}@{}}

% ---------- First row: captions ----------
\begin{minipage}[t]{\linewidth}
    \centering
    \captionof{table}{Performance comparison under \textit{dynamic} CR between COMPOT and SVD-LLM V2 \citep{svdllmv2} on Llama-7B, OPT-6.7B, and Llama3-8B at a CR 0.2. Best results are shown in \textbf{bold}}
    \label{tab:main_dynamic_svdllmv2}
\end{minipage}
&
\begin{minipage}[t]{\linewidth}
    \centering
    \captionof{table}{Perplexity ($\downarrow$) of COMPOT vs state-of-the-art SVD-based SVD-LLM V2 \citep{svdllmv2} accompanied by quantization on Llama-7B on WikiText-2. Best results are shown in \textbf{bold}.}
    \label{tab:main_dynamic_quant}
\end{minipage}
\\[0.8em]

% ---------- Second row: tables ----------
\begin{minipage}[t]{\linewidth}
    \centering
    \resizebox{\linewidth}{!}{%
    \renewcommand{\arraystretch}{1.3}
    \begin{tabular}{cccc}
    \hline
    \multicolumn{1}{c}{\multirow{2}{*}{\textbf{Methods}}} & \textbf{Llama-7B} & \textbf{OPT-6.7B} & \textbf{Llama3-8B} \\
    \multicolumn{1}{c}{} & Wiki / C4 $\downarrow$ & Wiki / C4 $\downarrow$ & Wiki / C4 $\downarrow$        \\ \hline \hline
    Original    & 5.67 / 7.33  & 10.92 / 12.57 & 6.14 / 9.47  \\ \hline
    SVD-LLM V2  & 7.12 / 10.47 & 13.46 / 17.72 & \textbf{8.01} / \textbf{11.72} \\
    SVD-LLM V2 (repr.)   & 7.92 / 16.23 & 15.16 / 19.77 & 48.80 / 88.88 \\ 
    COMPOT               & \textbf{6.52} / \textbf{9.63}   & \textbf{10.92} / \textbf{13.17}  & 8.64 / 18.29 \\
    \hline
    \end{tabular}
    }
\end{minipage}
&
\begin{minipage}[t]{\linewidth}
    \centering
    \resizebox{\linewidth}{!}{%
    \renewcommand{\arraystretch}{1.2}
    \begin{tabular}{ccccc}
    \hline
    \textbf{Method} & \textbf{Memory} & \textbf{QCR} & \textbf{FCR} & \textbf{PPL} \\ 
    \hline \hline
    
    \textbf{GPTQ-3bit}              & 2.8 GB & 0.81 & N/A  & 16.28 \\ \hline
    \textbf{SVD-LLM V2+GPTQ-4bit}   & 2.8 GB & 0.75 & 0.25 & 9.97 \\ 
    \textbf{COMPOT\txtdag+GPTQ-4bit}& 2.8 GB & 0.75 & 0.25 & 9.93 \\ 
    \textbf{COMPOT+GPTQ-4bit}       & 2.8 GB & 0.75 & 0.25 & \textbf{9.62} \\ 
    \hline
    
    \end{tabular}
    }
\end{minipage}

\end{tabular}
\end{center}
\vspace{-0.3cm}
\begin{table*}[h]
\caption{Comparison of the COMPOT with other training-free methods including state-of-the-art structured pruning methods ReplaceMe \cite{replaceme} and LLM-Pruner \cite{llmpruner} on Llama3 8B under different compression ratios. We report accuracy on different benchmarks as well as its average and perplexity. Best results are shown in \textbf{bold}.}
\label{tab:main_pruning}
\resizebox{\textwidth}{!}{%
\renewcommand{\arraystretch}{1.05}
\begin{tabular}{ccccccccccc:cc}
\hline
                              &                                           & \multicolumn{9}{c}{\textbf{Accuracy$\uparrow$}}                                                                           & \multicolumn{2}{c}{\textbf{Perplexity$\downarrow$}}       \\ \cline{3-13}
\multirow{-2}{*}{Method}      & \multirow{-2}{*}{CR}                      & \textbf{PIQA} & \textbf{Hella Swag} & \textbf{LAMBADA} & \textbf{ARC-e} & \textbf{ARC-c} & \textbf{SciQ} & \textbf{Race}& \textbf{MMLU} & \textbf{Avg.} & \textbf{WikiText} & \textbf{LAMBADA} \\ \hline \hline
 
\multicolumn{2}{c}{\textbf{Llama3 8B}} & 80.7 & 79.1 & 75.6 & 77.7 & 53.5 & 93.9 & 40.3 & 62.2 & 70.4 & 7.3E+00  & 3.1E+00            \\ \hline

ReplaceMe                   & 0.22 & 73.1 & 65.7 & 42.1 & 65.9 & 43.7 & 86.4 & 35.4 & 51.7 & 58.0 & 3.4E+01  & 2.0E+01            \\
LLM-Pruner                  &      & 75.5 & 67.5 & 51.0 & 62.1 & 36.6 & 87.8 & 35.1 & 25.0 & 55.1 & 1.6E+01  & 1.1E+01            \\
COMPOT & \multirow{-2}{*}{0.20}     & 78.0 & 73.5 & 70.5 & 71.3 & 45.3 & 91.0 & 40.1 & 52.4 & \textbf{65.3} & \textbf{1.2E+01} & \textbf{4.6E+00} \\ \hline

ReplaceMe                   & 0.31 & 66.6 & 53.8 & 24.0 & 50.7 & 37.9 & 77.3 & 34.0 & 30.6 & 46.9 & 6.7E+01  & 1.3E+02            \\
LLM-Pruner     &            & 67.3 & 45.1 & 20.9 & 45.4 & 28.8 & 63.4 & 30.1 & 22.9 & 40.5 & 3.8E+01  & 2.2E+02            \\
COMPOT                      & \multirow{-2}{*}{0.30}   & 76.6 & 67.2 & 64.3 & 69.4 & 41.6 & 90.1 & 38.4 & 39.8 & \textbf{60.9} & \textbf{1.7E+01} & \textbf{6.6E+00} \\ \hline

ReplaceMe                   & 0.41                    & 61.7 & 44.3 & 9.8  & 37.4 & 27.5 & 60.4 & 31.6 & 26.4 & 37.4 & 2.3E+02  & 1.8E+03            \\
LLM-Pruner                  &                         & 50.3 & 25.8 & 1.5  & 26.4 & 25.8 & 28.1 & 21.8 & 23.2 & 25.4 & $\infty$ & 5.7E+05            \\
COMPOT                      & \multirow{-2}{*}{0.40}   & 71.0 & 55.8 & 44.9 & 56.6 & 33.5 & 85.4 & 36.6 & 29.6 & \textbf{51.7} & \textbf{4.3E+01} & \textbf{2.2E+01} \\ \hline

\end{tabular}
}
\end{table*}

\textbf{Comparison to structured pruning.}
We compare COMPOT against ReplaceMe \citep{replaceme} and LLM-Pruner \citep{llmpruner} on Llama3-8B (Table~\ref{tab:main_pruning}). We exclude unstructured pruning methods since high sparsity often does not translate into memory/latency savings on dense backends. At comparable effective compression, COMPOT preserves substantially higher accuracy and lower perplexity than structured pruning, suggesting that factorizing projections is a more robust mechanism for removing redundancy than deleting entire channels/heads.

\textbf{Compatibility with post-training quantization.}
Following SVD-LLM V2 \citep{svdllmv2}, we apply 4-bit GPTQ on top of both SVD-based and COMPOT-based factorizations and compare against GPTQ-only under matched weight memory. Table~\ref{tab:main_dynamic_quant} reports WikiText-2 perplexity on Llama-7B, together with the decomposition of total compression into quantization and factorization components (QCR and FCR, correspondingly). COMPOT+GPTQ yields lower perplexity than GPTQ-only and SVD-LLM V2+GPTQ, showing that COMPOT provides gains complementary to modern PTQ.

\textbf{Vision-language transfer.}
We evaluate COMPOT on Qwen3-VL-8B-Instruct at CR $0.2$ (additional CRs and setup are provided in the ~\ref{apx:omni}). Table~\ref{tab:vlm_benchmarks} reports MMMU~\citep{yue2023mmmu}, OCRBench~\citep{Liu_2024}, RealWorldQA, and MMStar~\citep{mmstar}. SVD-LLM degrades severely, whereas static COMPOT\txtdag and COMPOT retain strong multimodal performance, with COMPOT achieving the best average accuracy.

\textbf{Audio transfer.}
We evaluate COMPOT on Whisper \citep{whisper2023} (Whisper Base: 72M; Whisper Medium: 763M; Whisper Large: 2B) and report WER on LibriSpeech \textit{test-clean} (N=2{,}650) and \textit{test-other} (N=2{,}939)~\citep{librispeech}. As a baseline we apply SVD-LLM to the same decoder projections and calibrate on the same LibriSpeech-derived calibration set. Table~\ref{tab:whisper_results} shows that on Whisper Large, COMPOT is consistently more robust than SVD-LLM across CRs; additional audio results and full setup are provided in ~\ref{apx:omni}

\paragraph{Grouped sharing scenario.}
COMPOT naturally extends to a grouped setting where a single dictionary is shared across multiple projection layers. While related to BasisSharing~\citep{wang2025basis}, we instead learn a shared orthogonal dictionary across heterogeneous projections. For fair comparison, we follow their grouping protocol with static rank allocation and include CoSpaDi as a dictionary-learning baseline. Under equal compression, performance is regime-dependent: at $0.2$, grouped CoSpaDi is on par in accuracy, but COMPOT is substantially more efficient due to orthogonal updates and no iterative sparse coding; at higher compression, COMPOT consistently outperforms both %low-rank and dictionary-learning methods 
by a wide margin in reconstruction and downstream metrics (Table~\ref{tab:main_results_group}). We report normalized accuracies, with unnormalized values in parentheses following BasisSharing.

\vspace{-0.3cm}
\begin{center}
  \begin{minipage}{0.53\textwidth}
    \centering
    \captionof{table}{Multimodal performance of Qwen3-VL-8B-Instruct under $0.2$ compression ratio (CR) on vision-language benchmarks. Best results are shown in \textbf{bold}.}
    \label{tab:vlm_benchmarks}
    \resizebox{\textwidth}{!}{%
    \renewcommand{\arraystretch}{1.1}
    \begin{tabular}{l c c c c c}
    \hline
    \textbf{Method}     & \textbf{MMMU} $\uparrow$ & \textbf{OCRBench} $\uparrow$   & \textbf{RealWorldQA} $\uparrow$   & \textbf{MMStar} $\uparrow$ \\ \hline \hline
    Original            & 52.6          & 82.6                & 69.7          & 63.2 \\ \hline
    SVD-LLM             & 29.2          & 35.1                & 52.9          & 31.6 \\
    COMPOT\txtdag       & 41.9          & 64.1                & 59.2          & 50.8 \\
    COMPOT              & \textbf{44.7} & \textbf{66.9}       & \textbf{62.2} & \textbf{54.2} \\
    \hline
    \end{tabular}
    }
  \end{minipage}%
  \hfill
  \begin{minipage}{0.45\textwidth}
    \captionof{table}{ASR Performance (WER $\downarrow$) on LibriSpeech benchmarks for Whisper Large V3 model. Best results are shown in \textbf{bold}.}
    \label{tab:whisper_results}
    \resizebox{0.9\textwidth}{!}{%
    \renewcommand{\arraystretch}{1.05}
    \begin{tabular}{@{}lccc@{}}
    \hline
    \textbf{Method} & \textbf{CR} & $\textbf{WER}_{\textbf{test-clean}}$ $\downarrow$ & $\textbf{WER}_{\textbf{test-other}}$ $\downarrow$ \\ \hline \hline
    Original  & -    & 2.74& 4.53\\ \hline
    SVD-LLM                    & 0.2  & 4.12& 6.8\\
    COMPOT\txtdag     & 0.2  & \textbf{2.46}& \textbf{4.51}\\ \cline{1-4}
    SVD-LLM                    & 0.3  & 12.78& 15.54\\
    COMPOT\txtdag     & 0.3  & \textbf{2.74}& \textbf{5.21}\\ \hline 
    \end{tabular}
    }
  \end{minipage}
\end{center}
\vspace{-0.4cm}

\begin{table}[h]
\caption{Performance comparison of COMPOT\txtdag (static CR) vs low-rank and dictionary-learning methods in per-layer and grouping scenarios on Llama2-7B under different compression levels on various benchmarks. Best average accuracy and perplexities are highlighted with \textbf{bold}. Unnormalized accuracies are shown in parentheses for direct comparison with~\citep{wang2025basis, shopkhoev2025cospadi}.}
\label{tab:main_results_group}
\resizebox{\textwidth}{!}{%
\renewcommand{\arraystretch}{1.35}
\begin{tabular}{lcccccccccc:cc}
\hline
\multirow{2}{*}{\textbf{Method}} & \multirow{2}{*}{\textbf{CR}} & \multicolumn{9}{c}{\textbf{Accuracy$\uparrow$ (\% (unnorm))}} & \multicolumn{2}{c}{\textbf{Perplexity$\downarrow$}} \\ \cline{3-13}
                                  &                             & \textbf{PIQA} & \textbf{Hella Swag} & \textbf{LAMBADA} & \textbf{ARC-e} & \textbf{ARC-c} & \textbf{SciQ} & \textbf{Race} & \textbf{MMLU} & \textbf{Avg.} & \textbf{WikiText} & \textbf{LAMBADA} \\ \hline

\textbf{Llama2 7B}                & --                          & 78.9 (78.0) & 76.1 (57.3) & 73.8 & 74.2 (76.0) & 45.8 (42.7) & 91.4 (93.9) & 39.7 & 40.8 & 65.1 (62.8) & 8.70 & 3.38 \\ \hline \hline

SVD-LLM                           & \multirow{6}{*}{0.2}        & 73.1 (72.1) & 60.3 (44.4) & 63.9 & 55.0 (62.0) & 32.0 (29.1) & 79.5 (89.2) & 36.6 & 25.5 & 53.2 (52.9) & 20.18 & 6.33 \\
CoSpaDi (per-layer)               &                             & 74.5 (73.6) & 64.5 (47.7) & 70.0 & 61.5 (67.2) & 35.2 (34.0) & 85.4 (91.3) & 39.2 & 28.0 & 57.3 (56.4) & 15.97 & 4.58 \\
COMPOT\txtdag (per-layer)                &                             & 76.9 (75.4) & 68.8 (50.3) & 69.4 & 64.2 (69.8) & 38.7 (36.7) & 85.7 (91.8) & 39.5 & 29.4 & 59.1 (57.8) & 12.46 & 5.00 \\ \cdashline{3-13}
BasisSharing                     &                             & 71.1 (70.2) & 60.0 (43.4) & 62.8 & 60.5 (65.5) & 37.8 (33.5) & 85.0 (90.8) & 34.7 & 25.0 & 54.6 (53.2) & 15.17 & 7.03 \\
CoSpaDi (grouped)                 &                             & 75.3 (74.5) & 66.3 (48.8) & 71.1 & 68.1 (71.3) & 39.3 (37.3) & 88.5 (92.2) & 38.5 & 27.0 & \textbf{59.3 (57.6)} & 11.73 & \textbf{4.42} \\
COMPOT\txtdag (grouped)                  &                             & 74.2 (72.7) & 70.0 (51.0) & 68.6 & 63.6 (69.1) & 39.8 (37.7) & 90.4 (93.2) & 39.6 & 28.4 & \textbf{59.3} (57.5) & \textbf{11.58} & 4.98 \\ \hline

SVD-LLM                           & \multirow{6}{*}{0.3}        & 68.2 (68.0) & 52.2 (39.4) & 51.6 & 47.3 (53.6) & 28.1 (25.2) & 75.9 (84.9) & 34.6 & 23.5 & 47.7 (47.6) & 33.99 & 13.54 \\
CoSpaDi (per-layer)               &                             & 71.4 (70.1) & 57.1 (42.4) & 62.7 & 53.2 (60.2) & 31.1 (27.5) & 81.7 (88.3) & 36.3 & 26.7 & 52.5 (51.8) & 23.90 & 6.75 \\
COMPOT\txtdag (per-layer)                &                             & 74.1 (72.6) & 64.0 (47.0) & 66.6 & 59.7 (65.2) & 36.5 (33.1) & 83.2 (90.8) & 38.3 & 27.4 & 56.2 (55.1) & 16.00 & 5.74 \\ \cdashline{3-13}
BasisSharing                     &                             & 66.5 (65.4) & 50.3 (37.6) & 53.6 & 54.2 (58.7) & 29.3 (27.2) & 81.4 (86.9) & 32.4 & 23.3 & 48.9 (48.1) & 22.21 & 13.18 \\
CoSpaDi (grouped)                 &                             & 71.0 (68.8) & 58.5 (42.3) & 64.1 & 63.5 (67.4) & 35.5 (33.4) & 87.7 (91.3) & 35.8 & 24.0 & 55.0 (53.4) & 15.43 & 6.51 \\
COMPOT\txtdag (grouped)                  &                             & 73.0 (71.4) & 66.6 (48.8) & 67.3 & 61.2 (66.5) & 36.9 (36.0) & 88.7 (92.8) & 38.4 & 27.4 & \textbf{57.4} (56.1) & \textbf{13.12} & \textbf{5.66} \\ \hline

SVD-LLM                           & \multirow{6}{*}{0.4}        & 63.3 (62.9) & 42.9 (33.4) & 33.0 & 37.7 (41.9) & 24.7 (21.3) & 70.1 (77.0) & 32.3 & 23.0 & 40.9 (40.6) & 86.50 & 66.91 \\
CoSpaDi (per-layer)          &                                  & 66.0 (65.1) & 47.9 (36.9) & 46.4 & 44.4 (48.9) & 26.5 (22.7) & 76.7 (84.0) & 33.4 & 23.5 & 45.6 (45.1) & 50.46 & 21.50 \\
COMPOT\txtdag (per-layer)           &                             & 71.3 (69.3) & 56.4 (41.8) & 56.2 & 50.8 (56.9) & 30.1 (27.1) & 77.6 (86.8) & 35.0 & 23.8 & 50.2 (49.6) & 25.01 & 9.95 \\ \cdashline{3-13}
BasisSharing                     &                             & 60.7 (60.3) & 41.5 (33.0) & 41.0 & 44.6 (49.1) & 26.5 (23.5) & 75.4 (82.8) & 30.1 & 23.2 & 42.9 (42.9) & 39.58 & 36.49 \\
CoSpaDi (grouped)                 &                             & 64.8 (63.4) & 48.0 (36.1) & 51.8 & 51.9 (56.7) & 29.1 (25.6) & 80.5 (88.2) & 32.7 & 23.0 & 47.7 (47.2) & 25.29 & 14.99 \\
COMPOT\txtdag (grouped)                  &                             & 68.0 (67.3) & 58.5 (42.3) & 58.2 & 57.0 (62.4) & 33.7 (30.3) & 86.1 (89.5) & 34.8 & 23.2 & \textbf{52.5 (51.0)} & \textbf{17.84} & \textbf{9.74} \\ \hline
\end{tabular}
}
\end{table}

\textbf{Computational efficiency.}
~\ref{apx:wallclock} reports substantial compression-time speedups over iterative dictionary learning.
~\ref{apx:acceleration} further shows comparable theoretical inference complexity to SVD and, using
MACKO sparse kernels, a 21.1\% average throughput improvement over dense inference.
\vspace{-0.3cm}\section{Limitations and Conclusions}\vspace{-0.3cm}
\label{sec:lim_conc}

We presented \textsc{COMPOT}, a training-free weight compression framework using orthogonal dictionary learning with closed-form Procrustes updates, analytical sparse coding, one-shot global allocation and theoretical convergence guarantees for its alternating optimization procedure. Across language, vision-language, and speech models, \textsc{COMPOT} improves the quality--compression trade-off over strong SVD and sparse dictionary baselines.

\textsc{COMPOT} shares limitations of calibration-driven compression: performance depends on representative calibration data. Whitening is straightforward with a well-conditioned Gram/covariance matrix; otherwise mild regularization is needed. Future work could improve allocation, grouping or add a lightweight ``healing'' step, either with fixed or learned sparsity pattern on the matrix coefficients.

\textbf{Broader impacts.}
\textsc{COMPOT} can reduce memory, compute, deployment cost, and energy use, improving access to large language, vision-language, and speech models. Easier deployment may also amplify existing model risks such as bias, privacy leakage, misinformation, or misuse; compressed models should therefore follow the same safeguards and domain-specific evaluations as their uncompressed counterparts.

\section*{Acknowledgments}
This research was supported as part of the ITMO University Research Projects in AI Initiative

\bibliography{bibliography}

\begin{thebibliography}{10}

\bibitem{aharon2006ksvd}
{\sc Aharon, M., Elad, M., and Bruckstein, A.}
\newblock {K-SVD}: An algorithm for designing overcomplete dictionaries for
  sparse representation.
\newblock {\em IEEE Transactions on signal processing 54}, 11 (2006),
  4311--4322.

\bibitem{bao2013fast}
{\sc Bao, C., Cai, J.-F., and Ji, H.}
\newblock Fast sparsity-based orthogonal dictionary learning for image
  restoration.
\newblock In {\em Proceedings of the IEEE International Conference on Computer
  Vision (ICCV)\/} (2013).

\bibitem{bishop2006pattern}
{\sc Bishop, C.~M.}
\newblock {\em Pattern recognition and machine learning}, vol.~4.
\newblock Springer, 2006.

\bibitem{bisk2020piqa}
{\sc Bisk, Y., Zellers, R., Gao, J., Choi, Y., et~al.}
\newblock Piqa: Reasoning about physical commonsense in natural language.
\newblock In {\em Proceedings of the AAAI conference on artificial
  intelligence\/} (2020), vol.~34, pp.~7432--7439.

\bibitem{brown2020language}
{\sc Brown, T., Mann, B., Ryder, N., Subbiah, M., Kaplan, J.~D., Dhariwal, P.,
  Neelakantan, A., Shyam, P., Sastry, G., Askell, A., et~al.}
\newblock Language models are few-shot learners.
\newblock {\em Advances in neural information processing systems 33\/} (2020),
  1877--1901.

\bibitem{mmstar}
{\sc Chen, L., Li, J., Dong, X., Zhang, P., Zang, Y., Chen, Z., Duan, H., Wang,
  J., Qiao, Y., Lin, D., and Zhao, F.}
\newblock Are we on the right way for evaluating large vision-language models?
\newblock In {\em The Thirty-eighth Annual Conference on Neural Information
  Processing Systems\/} (2024).

\bibitem{chen2021drone}
{\sc Chen, P.~H., Yu, H.-F., Dhillon, I.~S., and Hsieh, C.-J.}
\newblock {DRONE}: Data-aware low-rank compression for large {NLP} models.
\newblock In {\em Advances in Neural Information Processing Systems\/} (2021),
  vol.~34, pp.~29321--29334.

\bibitem{chowdhery2023palm}
{\sc Chowdhery, A., Narang, S., Devlin, J., Bosma, M., Mishra, G., Roberts, A.,
  Barham, P., Chung, H.~W., Sutton, C., Gehrmann, S., et~al.}
\newblock Palm: Scaling language modeling with pathways.
\newblock {\em Journal of Machine Learning Research 24}, 240 (2023), 1--113.

\bibitem{clark2018think}
{\sc Clark, P., Cowhey, I., Etzioni, O., Khot, T., Sabharwal, A., Schoenick,
  C., and Tafjord, O.}
\newblock Think you have solved question answering? try arc, the ai2 reasoning
  challenge.
\newblock {\em arXiv preprint arXiv:1803.05457\/} (2018).

\bibitem{dosovitskiy2020image}
{\sc Dosovitskiy, A., Beyer, L., Kolesnikov, A., Weissenborn, D., Zhai, X.,
  Unterthiner, T., Dehghani, M., Minderer, M., Heigold, G., Gelly, S.,
  Uszkoreit, J., and Houlsby, N.}
\newblock An image is worth 16x16 words: Transformers for image recognition at
  scale.
\newblock In {\em International Conference on Learning Representations\/}
  (2021).

\bibitem{elad2010sparse}
{\sc Elad, M.}
\newblock {\em Sparse and redundant representations: from theory to
  applications in signal and image processing}.
\newblock Springer Science \& Business Media, 2010.

\bibitem{open-llm-leaderboard-v2}
{\sc Fourrier, C., Habib, N., Lozovskaya, A., Szafer, K., and Wolf, T.}
\newblock Open llm leaderboard v2.
\newblock
  \url{https://huggingface.co/collections/open-llm-leaderboard/open-llm-leaderboard-2},
  2024.

\bibitem{frantar2022gptq}
{\sc Frantar, E., Ashkboos, S., Hoefler, T., and Alistarh, D.}
\newblock Optq: Accurate quantization for generative pre-trained transformers.
\newblock In {\em The Eleventh International Conference on Learning
  Representations\/} (2023).

\bibitem{eval-harness}
{\sc Gao, L., Tow, J., Abbasi, B., Biderman, S., Black, S., DiPofi, A., Foster,
  C., Golding, L., Hsu, J., Le~Noac'h, A., Li, H., McDonell, K., Muennighoff,
  N., Ociepa, C., Phang, J., Reynolds, L., Schoelkopf, H., Skowron, A.,
  Sutawika, L., Tang, E., Thite, A., Wang, B., Wang, K., and Zou, A.}
\newblock The language model evaluation harness.
\newblock \url{https://zenodo.org/records/12608602}, 07 2024.

\bibitem{gong21b_interspeech}
{\sc Gong, Y., Chung, Y.-A., and Glass, J.}
\newblock {AST: Audio Spectrogram Transformer}.
\newblock In {\em Proc. Interspeech 2021\/} (2021), pp.~571--575.

\bibitem{hendryckstest2021}
{\sc Hendrycks, D., Burns, C., Basart, S., Zou, A., Mazeika, M., Song, D., and
  Steinhardt, J.}
\newblock Measuring massive multitask language understanding.
\newblock {\em Proceedings of the International Conference on Learning
  Representations (ICLR)\/} (2021).

\bibitem{hendrycks2021measuringmathematicalproblemsolving}
{\sc Hendrycks, D., Burns, C., Kadavath, S., Arora, A., Basart, S., Tang, E.,
  Song, D., and Steinhardt, J.}
\newblock Measuring mathematical problem solving with the math dataset.
\newblock In {\em Proceedings of the Neural Information Processing Systems
  Track on Datasets and Benchmarks\/} (2021), J.~Vanschoren and S.~Yeung, Eds.,
  vol.~1.

\bibitem{hsu2022fwswd}
{\sc Hsu, Y.-C., Hua, T., Chang, S., Lou, Q., Shen, Y., and Jin, H.}
\newblock Language model compression with weighted low-rank factorization.
\newblock In {\em International Conference on Learning Representations\/}
  (2022).

\bibitem{lai2017race}
{\sc Lai, G., Xie, Q., Liu, H., Yang, Y., and Hovy, E.}
\newblock {RACE}: Large-scale {R}e{A}ding comprehension dataset from
  examinations.
\newblock In {\em Proceedings of the 2017 Conference on Empirical Methods in
  Natural Language Processing\/} (Copenhagen, Denmark, Sept. 2017), M.~Palmer,
  R.~Hwa, and S.~Riedel, Eds., Association for Computational Linguistics,
  pp.~785--794.

\bibitem{lin2023awq}
{\sc Lin, J., Tang, J., Tang, H., Yang, S., Chen, W.-M., Wang, W.-C., Xiao, G.,
  Dang, X., Gan, C., and Han, S.}
\newblock Awq: Activation-aware weight quantization for on-device llm
  compression and acceleration.
\newblock {\em Proceedings of machine learning and systems 6\/} (2024),
  87--100.

\bibitem{liu2025survey}
{\sc Liu, D., Zhu, Y., Liu, Z., Liu, Y., Han, C., Tian, J., Li, R., and Yi, W.}
\newblock A survey of model compression techniques: Past, present, and future.
\newblock {\em Frontiers in Robotics and AI 12\/} (2025), 1518965.

\bibitem{Liu_2024}
{\sc Liu, Y., Li, Z., Huang, M., Yang, B., Yu, W., Li, C., Yin, X.-C., Liu,
  C.-L., Jin, L., and Bai, X.}
\newblock Ocrbench: on the hidden mystery of ocr in large multimodal models.
\newblock {\em Science China Information Sciences 67}, 12 (Dec. 2024).

\bibitem{llmpruner}
{\sc Ma, X., Fang, G., and Wang, X.}
\newblock Llm-pruner: On the structural pruning of large language models.
\newblock {\em Advances in neural information processing systems 36\/} (2023),
  21702--21720.

\bibitem{macko2026macko}
{\sc Macko, V., and Bo{\v{z}}a, V.}
\newblock {MACKO}: Sparse matrix-vector multiplication for low sparsity.
\newblock In {\em Forty-third International Conference on Machine Learning\/}
  (2026).

\bibitem{shopkhoev2025cospadi}
{\sc Makhov, D., Shopkhoev, D., Zhussip, M., Ali, A., and Lefkimmiatis, S.}
\newblock {CoSpaDi}: Compressing llms via calibration-guided sparse dictionary
  learning.
\newblock {\em EMNLP Findings\/} (2026).

\bibitem{wikitext}
{\sc Merity, S., Xiong, C., Bradbury, J., and Socher, R.}
\newblock Pointer sentinel mixture models, 2016.

\bibitem{merity2017pointer}
{\sc Merity, S., Xiong, C., Bradbury, J., and Socher, R.}
\newblock Pointer sentinel mixture models.
\newblock In {\em International Conference on Learning Representations\/}
  (2017).

\bibitem{mi2025layerwisedynamicrankcompressing}
{\sc Mi, Z., Sun, B., Zhang, G.~L., and Huang, S.}
\newblock Layer-wise dynamic rank for compressing large language models.
\newblock {\em arXiv preprint arXiv:2509.25622\/} (2025).

\bibitem{michel2019sixteen}
{\sc Michel, P., Levy, O., and Neubig, G.}
\newblock Are sixteen heads really better than one?
\newblock {\em Advances in neural information processing systems 32\/} (2019).

\bibitem{librispeech}
{\sc Panayotov, V., Chen, G., Povey, D., and Khudanpur, S.}
\newblock Librispeech: An asr corpus based on public domain audio books.
\newblock In {\em 2015 IEEE International Conference on Acoustics, Speech and
  Signal Processing (ICASSP)\/} (2015), pp.~5206--5210.

\bibitem{paperno2016lambada}
{\sc Paperno, D., Kruszewski, G., Lazaridou, A., Pham, N.-Q., Bernardi, R.,
  Pezzelle, S., Baroni, M., Boleda, G., and Fern{\'a}ndez, R.}
\newblock The lambada dataset: Word prediction requiring a broad discourse
  context.
\newblock In {\em Proceedings of the 54th annual meeting of the association for
  computational linguistics (volume 1: Long papers)\/} (2016), pp.~1525--1534.

\bibitem{fineweb}
{\sc Penedo, G., Kydl{\'\i}{\v{c}}ek, H., allal, L.~B., Lozhkov, A., Mitchell,
  M., Raffel, C., Werra, L.~V., and Wolf, T.}
\newblock The fineweb datasets: Decanting the web for the finest text data at
  scale.
\newblock In {\em The Thirty-eight Conference on Neural Information Processing
  Systems Datasets and Benchmarks Track\/} (2024).

\bibitem{dobisvd}
{\sc Qinsi, W., Ke, J., Tomizuka, M., Keutzer, K., and Xu, C.}
\newblock Dobi-{SVD}: Differentiable {SVD} for {LLM} compression and some new
  perspectives.
\newblock In {\em The Thirteenth International Conference on Learning
  Representations\/} (2025).

\bibitem{radford2021learning}
{\sc Radford, A., Kim, J.~W., Hallacy, C., Ramesh, A., Goh, G., Agarwal, S.,
  Sastry, G., Askell, A., Mishkin, P., Clark, J., et~al.}
\newblock Learning transferable visual models from natural language
  supervision.
\newblock In {\em International conference on machine learning\/} (2021), PmLR,
  pp.~8748--8763.

\bibitem{whisper2023}
{\sc Radford, A., Kim, J.~W., Xu, T., Brockman, G., McLeavey, C., and
  Sutskever, I.}
\newblock Robust speech recognition via large-scale weak supervision.
\newblock In {\em International conference on machine learning\/} (2023), PMLR,
  pp.~28492--28518.

\bibitem{rein2023gpqagraduatelevelgoogleproofqa}
{\sc Rein, D., Hou, B.~L., Stickland, A.~C., Petty, J., Pang, R.~Y., Dirani,
  J., Michael, J., and Bowman, S.~R.}
\newblock {GPQA}: A graduate-level google-proof q\&a benchmark.
\newblock In {\em First Conference on Language Modeling\/} (2024).

\bibitem{sanh2019distilbert}
{\sc Sanh, V., Debut, L., Chaumond, J., and Wolf, T.}
\newblock Distil{BERT}, a distilled version of {BERT}: Smaller, faster, cheaper
  and lighter.
\newblock In {\em NeurIPS 2019 Workshop on Energy Efficient Machine Learning
  and Cognitive Computing\/} (2019).

\bibitem{schoenemann1966procrustes}
{\sc Sch{\"o}nemann, P.~H.}
\newblock A generalized solution of the orthogonal procrustes problem.
\newblock {\em Psychometrika 31}, 1 (1966), 1--10.

\bibitem{replaceme}
{\sc Shopkhoev, D., Ali, A., Zhussip, M., Malykh, V., Lefkimmiatis, S.,
  Komodakis, N., and Zagoruyko, S.}
\newblock Replaceme: Network simplification via depth pruning and transformer
  block linearization.
\newblock In {\em The Thirty-ninth Annual Conference on Neural Information
  Processing Systems\/} (2025).

\bibitem{sprague2024musrtestinglimitschainofthought}
{\sc Sprague, Z.~R., Ye, X., Bostrom, K., Chaudhuri, S., and Durrett, G.}
\newblock Mu{SR}: Testing the limits of chain-of-thought with multistep soft
  reasoning.
\newblock In {\em The Twelfth International Conference on Learning
  Representations\/} (2024).

\bibitem{suzgun2022challengingbigbenchtaskschainofthought}
{\sc Suzgun, M., Scales, N., Sch{\"a}rli, N., Gehrmann, S., Tay, Y., Chung,
  H.~W., Chowdhery, A., Le, Q., Chi, E., Zhou, D., et~al.}
\newblock Challenging big-bench tasks and whether chain-of-thought can solve
  them.
\newblock In {\em Findings of the Association for Computational Linguistics:
  ACL 2023\/} (2023), pp.~13003--13051.

\bibitem{tang2024survey}
{\sc Tang, Y., Wang, Y., Guo, J., Tu, Z., Han, K., Hu, H., and Tao, D.}
\newblock A survey on transformer compression.
\newblock {\em arXiv preprint arXiv:2402.05964\/} (2024).

\bibitem{alpaca}
{\sc Taori, R., Gulrajani, I., Zhang, T., Dubois, Y., Li, X., Guestrin, C.,
  Liang, P., and Hashimoto, T.~B.}
\newblock Stanford alpaca: An instruction-following llama model.
\newblock \url{https://github.com/tatsu-lab/stanford_alpaca}, 2023.

\bibitem{vaswani2017attention}
{\sc Vaswani, A., Shazeer, N., Parmar, N., Uszkoreit, J., Jones, L., Gomez,
  A.~N., Kaiser, {\L}., and Polosukhin, I.}
\newblock Attention is all you need.
\newblock {\em Advances in neural information processing systems 30\/} (2017).

\bibitem{voita2019analyzing}
{\sc Voita, E., Talbot, D., Moiseev, F., Sennrich, R., and Titov, I.}
\newblock Analyzing multi-head self-attention: Specialized heads do the heavy
  lifting, the rest can be pruned.
\newblock In {\em Proceedings of the 57th Annual Meeting of the Association for
  Computational Linguistics\/} (Florence, Italy, July 2019), Association for
  Computational Linguistics, pp.~5797--5808.

\bibitem{wang2025basis}
{\sc Wang, J., Chen, Y.-G., Lin, I.-C., Li, B., and Zhang, G.~L.}
\newblock Basis sharing: Cross-layer parameter sharing for large language model
  compression.
\newblock In {\em The Thirteenth International Conference on Learning
  Representations\/} (2025).

\bibitem{svdllmv2}
{\sc Wang, X., Alam, S., Wan, Z., Shen, H., and Zhang, M.}
\newblock Svd-llm v2: Optimizing singular value truncation for large language
  model compression.
\newblock In {\em Proceedings of the 2025 Conference of the Nations of the
  Americas Chapter of the Association for Computational Linguistics: Human
  Language Technologies (Volume 1: Long Papers)\/} (2025), pp.~4287--4296.

\bibitem{svdllm}
{\sc Wang, X., Zheng, Y., Wan, Z., and Zhang, M.}
\newblock Svd-llm: Truncation-aware singular value decomposition for large
  language model compression.
\newblock In {\em The Thirteenth International Conference on Learning
  Representations\/} (2025).

\bibitem{welbl2017crowdsourcing}
{\sc Welbl, J., Liu, N.~F., and Gardner, M.}
\newblock Crowdsourcing multiple choice science questions.
\newblock In {\em Proceedings of the 3rd Workshop on Noisy User-generated
  Text\/} (Copenhagen, Denmark, Sept. 2017), L.~Derczynski, W.~Xu, A.~Ritter,
  and T.~Baldwin, Eds., Association for Computational Linguistics, pp.~94--106.

\bibitem{xiao2023smoothquant}
{\sc Xiao, G., Lin, J., Seznec, M., Wu, H., Demouth, J., and Han, S.}
\newblock Smoothquant: Accurate and efficient post-training quantization for
  large language models.
\newblock In {\em International Conference on Machine Learning (ICML)\/}
  (2023).

\bibitem{xv2025araadaptiverankallocation}
{\sc Xv, L., Gao, J., Gao, X., Liu, T., and Fu, Y.}
\newblock Ara: Adaptive rank allocation for efficient large language model svd
  compression.
\newblock {\em arXiv preprint arXiv:2510.19389\/} (2025).

\bibitem{yuan2023asvd}
{\sc Yuan, Z., Shang, Y., Song, Y., Yang, D., Wu, Q., Yan, Y., and Sun, G.}
\newblock Asvd: Activation-aware singular value decomposition for compressing
  large language models.
\newblock {\em arXiv preprint arXiv:2312.05821\/} (2023).

\bibitem{yue2023mmmu}
{\sc Yue, X., Ni, Y., Zhang, K., Zheng, T., Liu, R., Zhang, G., Stevens, S.,
  Jiang, D., Ren, W., Sun, Y., Wei, C., Yu, B., Yuan, R., Sun, R., Yin, M.,
  Zheng, B., Yang, Z., Liu, Y., Huang, W., Sun, H., Su, Y., and Chen, W.}
\newblock Mmmu: A massive multi-discipline multimodal understanding and
  reasoning benchmark for expert agi.
\newblock In {\em Proceedings of CVPR\/} (2024).

\bibitem{zellers2019hellaswag}
{\sc Zellers, R., Holtzman, A., Bisk, Y., Farhadi, A., and Choi, Y.}
\newblock Hellaswag: Can a machine really finish your sentence?
\newblock In {\em Proceedings of the 57th Annual Meeting of the Association for
  Computational Linguistics\/} (2019), Association for Computational
  Linguistics.

\bibitem{zhou2023instructionfollowingevaluationlargelanguage}
{\sc Zhou, J., Lu, T., Mishra, S., Brahma, S., Basu, S., Luan, Y., Zhou, D.,
  and Hou, L.}
\newblock Instruction-following evaluation for large language models.
\newblock {\em arXiv preprint arXiv:2311.07911\/} (2023).

\bibitem{zhussip2025shareattention}
{\sc Zhussip, M., Shopkhoev, D., Ali, A., and Lefkimmiatis, S.}
\newblock Share your attention: Transformer weight sharing via matrix-based
  dictionary learning.
\newblock In {\em Proceedings of the AAAI Conference on Artificial
  Intelligence\/} (2026), vol.~40, pp.~29260--29268.

\end{thebibliography}
\bibliographystyle{acm}

\appendix
\onecolumn
\section{Appendix}
Due to the page limitations in the main text, we present here additional results and descriptions to provide more details and evidence on the superiority of our proposed COMPOT compression method.

\subsection{COMPOT Optimization Details}
\label{apx:compot_algo}

In this section we provide the full optimization details for COMPOT. We first describe the general calibration-aware setup and the orthogonal sparse factorization problem, then present the alternating minimization procedure, the dictionary initialization strategies used in practice, the analytical solutions for the sparse coding and dictionary update steps, and finally the convergence analysis.

\subsubsection{General Setup}

Recall the calibration loss \eqref{eq:functional_loss} and the Gram matrix $\m G=\m X^\top \m X$.
Assuming $\m G\succ 0$, let $\m G=\m L\m L^\top$ be its Cholesky factorization. Then
\begin{align}
\|\m X(\m W-\widehat{\m W})\|_F^2
&=
\mathrm{Tr}\!\big((\m W-\widehat{\m W})^\top \m G (\m W-\widehat{\m W})\big)
=
\|\m L^\top(\m W-\widehat{\m W})\|_F^2.
\label{eq:apx_whitened_loss}
\end{align}
Defining $\widetilde{\m W}\triangleq \m L^\top \m W$ and $\widetilde{\widehat{\m W}}\triangleq \m L^\top \widehat{\m W}$, minimizing functional error in the original space is equivalent to minimizing reconstruction error in the whitened space:
\begin{equation}
\min_{\widehat{\m W}}\ \|\m X(\m W-\widehat{\m W})\|_F^2
\ \Longleftrightarrow\
\min_{\widetilde{\widehat{\m W}}}\ \|\widetilde{\m W}-\widetilde{\widehat{\m W}}\|_F^2.
\label{eq:apx_whitened_equiv}
\end{equation}

COMPOT solves the orthogonal sparse factorization in whitened coordinates:
\begin{equation}
\min_{\m D_O,\m S_O}\ \|\widetilde{\m W}-\m D_O\m S_O\|_F^2
\quad
\text{s.t.}\quad
\m D_O^\top \m D_O=\m I_k,\ \ \|\m {s_O}_j\|_0\le s\ \forall j\in[n],
\label{eq:apx_compot_main}
\end{equation}
with $\m D_O\in\R^{m\times k}$, $k\le m$ (complete/undercomplete), and $\m S_O\in\R^{k\times n}$.

After optimizing \eqref{eq:apx_compot_main}, we map the solution back to the original parameter space as in \eqref{eq:dewhiten}:
\begin{equation}
\widehat{\m W} \triangleq \m A \m S_O,\qquad
\m A \triangleq \m L^{-\top}\m D_O.
\label{eq:apx_dewhiten}
\end{equation}
The dewhitening factor $\m A$ is computed offline during compression; inference uses only $(\m A,\m S_O)$.

\paragraph{Numerically stable whitening.}
When the number of calibration samples is limited, the empirical Gram matrix
$\m G=\m X^\top \m X$ may be singular or poorly conditioned, which can make the Cholesky-based
whitening step unstable. In practice, we follow the stabilization used in the SVD-LLM
implementation~\citep{svdllm}. We first attempt the Cholesky decomposition in double precision.
If it succeeds, no ridge term is added. In our calibration setup, plain Cholesky factorization failed for only $3/576$ ($0.52\%$) activation Gram matrices, all for Qwen3-8B ($3/144$, $2.1\%$); no failures were observed for the remaining model and calibration-data. If it fails, we compute the smallest eigenvalue
$\lambda_{\min}(\m G)$ and replace $\m G$ by
\[
\m G_{\lambda}=\m G+\left(-\lambda_{\min}(\m G)+7\cdot 10^{-6}\right)\m I .
\]
This shifts the smallest eigenvalue to $7\cdot 10^{-6}$ before retrying the Cholesky
decomposition. The corresponding Cholesky factor is then used for whitening and
dewhitening exactly as in the non-regularized case.

\subsubsection{Alternating Minimization Procedure}

We optimize \eqref{eq:apx_compot_main} by alternating between sparse coding and dictionary update steps. With $\m D_O$ fixed, sparse coding admits a closed-form hard-thresholding solution under the orthogonality constraint. With $\m S_O$ fixed, the dictionary update becomes an orthogonal Procrustes problem with a closed-form SVD solution~\citep{schoenemann1966procrustes}. Algorithm~\ref{alg:compot_altmin} summarizes the full optimization for a single projection matrix $\m W$ given calibration inputs $\m X$.

\begin{algorithm}[t]
\caption{COMPOT factorization for one projection matrix}
\label{alg:compot_altmin}
\begin{algorithmic}[1]
\REQUIRE Weight matrix $\m W\in\R^{m\times n}$; calibration inputs $\m X\in\R^{N\times m}$; dictionary size $k\le m$; sparsity $s$; number of iterations $T_{\mathrm{it}}$; initialization mode $\mathrm{init}\in\{\mathrm{random},\mathrm{svd}\}$.
\ENSURE Compressed factors $\m A\in\R^{m\times k}$ and $\m S_O\in\R^{k\times n}$ such that $\m W\approx \m A\m S_O$.

\STATE Compute $\m G\leftarrow \m X^\top \m X$ and its Cholesky factor $\m G=\m L\m L^\top$.
\STATE Whiten weights: $\widetilde{\m W}\leftarrow \m L^\top \m W$.
\STATE Initialize $\m D_O^{(0)}\in\R^{m\times k}$ with $(\m D_O^{(0)})^\top \m D_O^{(0)}=\m I_k$ using either random-column or SVD initialization.
\FOR{$t=0,\dots,T_{\mathrm{it}}-1$}
  \STATE \textbf{Sparse coding (closed form):} $\m S_O^{(t+1)} \leftarrow \mathcal{H}_s\!\big((\m D_O^{(t)})^\top \widetilde{\m W}\big)$.
  \STATE \textbf{Dictionary update (Procrustes):} $\m M^{(t+1)}\leftarrow \widetilde{\m W}(\m S_O^{(t+1)})^\top$; compute thin SVD $\m M^{(t+1)}=\m P^{(t+1)}\m\Lambda^{(t+1)}(\m Q^{(t+1)})^\top$; set $\m D_O^{(t+1)} \leftarrow \m P^{(t+1)}(\m Q^{(t+1)})^\top$.
\ENDFOR
\STATE Dewhiten dictionary: $\m A \leftarrow \m L^{-\top}\m D_O^{(T_{\mathrm{it}})}$.
\RETURN $(\m A,\m S_O^{(T_{\mathrm{it}})})$.
\end{algorithmic}
\end{algorithm}

\subsubsection{Analytical Solution of the Sparse Coding Step}

Fix $\m D_O$ with $\m D_O^\top \m D_O=\m I_k$. The sparse coding step for a column $\widetilde{\m w}_j$ is
\begin{equation}
\min_{\m s\in\R^k}\ \|\widetilde{\m w}_j-\m D_O\m s\|_2^2
\quad\text{s.t.}\quad \|\m s\|_0\le s.
\label{eq:apx_sparse_coding}
\end{equation}
Let $\m z_j \triangleq \m D_O^\top \widetilde{\m w}_j$. Using $\m D_O^\top \m D_O=\m I_k$,
\begin{align}
\|\widetilde{\m w}_j-\m D_O\m s\|_2^2
&=
\|\widetilde{\m w}_j\|_2^2 - 2\,\m s^\top \m D_O^\top \widetilde{\m w}_j + \|\m D_O\m s\|_2^2 \nonumber\\
&=
\|\widetilde{\m w}_j\|_2^2 - 2\,\m s^\top \m z_j + \|\m s\|_2^2
=
\|\widetilde{\m w}_j\|_2^2 + \|\m s-\m z_j\|_2^2 - \|\m z_j\|_2^2.
\label{eq:apx_ortho_expand}
\end{align}
Since the first and last terms in \eqref{eq:apx_ortho_expand} do not depend on $\m s$, \eqref{eq:apx_sparse_coding} is equivalent to
\begin{equation}
\min_{\m s\in\R^k}\ \|\m s-\m z_j\|_2^2
\quad\text{s.t.}\quad \|\m s\|_0\le s,
\label{eq:apx_proj}
\end{equation}
i.e., Euclidean projection of $\m z_j$ onto the set of $s$-sparse vectors. This projection has a closed form:
\begin{equation}
\m s_O^\star = \mathcal{H}_s(\m z_j),
\qquad
\m z_j=\m D_O^\top \widetilde{\m w}_j,
\label{eq:apx_hard_threshold}
\end{equation}
where $\mathcal{H}_s(\cdot)$ retains the $s$ largest entries by magnitude and sets the rest to zero; ties can be broken arbitrarily without changing optimality. Applying this column-wise yields the matrix update
\begin{equation}
\m S_O \leftarrow \mathcal{H}_s\!\big(\m D_O^\top \widetilde{\m W}\big).
\label{eq:apx_sparse_matrix_update}
\end{equation}
Thus, under orthogonality, sparse coding is solved exactly without iterative pursuit methods.

\subsubsection{Analytical Solution of the Dictionary Update}

For fixed $\m S_O$, the dictionary update seeks the solution of the following problem:
\begin{equation}
\argmin_{\m D_O^\top \m D_O=\m I_k}\ \|\widetilde{\m W}-\m D_O\m S_O\|_F^2.
\label{eq:apx_procrustes}
\end{equation}
Expanding \eqref{eq:apx_procrustes}, we obtain
\begin{align}
\|\widetilde{\m W}-\m D_O\m S_O\|_F^2
&=
\|\widetilde{\m W}\|_F^2
+\|\m D_O\m S_O\|_F^2
-2\langle \widetilde{\m W},\m D_O\m S_O\rangle \nonumber\\
&=
\|\widetilde{\m W}\|_F^2
+\|\m S_O\|_F^2
-2\langle \m D_O,\widetilde{\m W}\m S_O^\top\rangle,
\end{align}
where we used $\m D_O^\top\m D_O=\m I_k$. Let
\begin{equation}
\m M\triangleq \widetilde{\m W}\m S_O^\top\in\R^{m\times k}.
\end{equation}
Dropping terms independent of $\m D_O$, the dictionary update is equivalent to
\begin{equation}
\argmax_{\m D_O^\top \m D_O=\m I_k}\ \mathrm{Tr}(\m D_O^\top\m M),
\label{eq:apx_procrustes_max}
\end{equation}
which is the classical orthogonal Procrustes problem. Let $\m M=\m P\m\Lambda\m Q^\top$ be the thin SVD; then the optimizer is
\begin{equation}
\m D_O \leftarrow \m P\m Q^\top.
\label{eq:apx_procrustes_solution}
\end{equation}

\subsubsection{Convergence Analysis}
\label{app:compot_convergence}

This section analyzes the convergence behavior of the alternating minimization procedure used in COMPOT. We first show that the sparse-coding and dictionary-update steps are exact global minimizers of their corresponding block subproblems, which implies monotone convergence of the objective values. We then provide a KL-based iterate-convergence result under standard non-degeneracy conditions that rule out support switching in hard thresholding and non-unique Procrustes updates.

Let
\begin{equation}
    \widetilde{\m W} \equiv \m L^\top \m W \in \R^{m \times n},
    \qquad
    \mathrm{St}(m,k) \equiv
    \{\m D \in \R^{m \times k}: \m D^\top \m D = \m I_k\},
\end{equation}
and define the column-wise sparsity set
\begin{equation}
    \mathcal{S}_s
    \equiv
    \{\m S \in \R^{k \times n}: \|\m s_j\|_0 \le s
    \ \text{for all } j \in [n]\}.
\end{equation}
In all main experiments, we use SVD initialization,
\(\m D_O^{(0)}=\m U_{[:,1:k]}\), where
\(\widetilde{\m W}=\m U\m\Sigma\m V^\top\). Hence
\(\m D_O^{(0)}\in\mathrm{St}(m,k)\), and every subsequent Procrustes update preserves
the Stiefel constraint. The analysis below is stated for this initialization.

The COMPOT objective is
\begin{equation}
    \Phi(\m D,\m S)
    \equiv
    \frac12 \|\widetilde{\m W} - \m D\m S\|_F^2,
    \qquad
    \m D \in \mathrm{St}(m,k), \quad \m S \in \mathcal{S}_s .
    \label{eq:compot_obj}
\end{equation}
Equivalently, this can be written as the nonsmooth constrained objective
\begin{equation}
    \Psi(\m D,\m S)
    =
    \frac12\|\widetilde{\m W} - \m D\m S\|_F^2
    + \iota_{\mathrm{St}(m,k)}(\m D)
    + \iota_{\mathcal{S}_s}(\m S),
    \label{eq:compot_nonsmooth}
\end{equation}
where \(\iota_{\mathcal C}\) is the indicator function of a set \(\mathcal C\).
The feasible set is nonconvex because of both the Stiefel constraint and the
\(\ell_0\)-type sparsity constraint. Nevertheless, the two COMPOT blocks admit
closed-form global minimizers.

At iteration \(t\), COMPOT performs
\begin{equation}
    \m S^{(t+1)}
    \in
    \arg\min_{\m S \in \mathcal{S}_s}
    \Phi(\m D^{(t)},\m S),
    \qquad
    \m D^{(t+1)}
    \in
    \arg\min_{\m D \in \mathrm{St}(m,k)}
    \Phi(\m D,\m S^{(t+1)}).
    \label{eq:compot_updates}
\end{equation}
The first update is implemented by hard thresholding,
\begin{equation}
    \m S^{(t+1)}
    =
    \mathcal{H}_s\!\left((\m D^{(t)})^\top \widetilde{\m W}\right),
    \label{eq:compot_ht}
\end{equation}
where \(\mathcal{H}_s(\cdot)\) keeps the \(s\) largest-magnitude entries in each column.
The second update is the orthogonal Procrustes update: if
\begin{equation}
    \m M^{(t+1)}
    \equiv
    \widetilde{\m W}(\m S^{(t+1)})^\top
    =
    \m P^{(t+1)}\m \Lambda^{(t+1)}(\m Q^{(t+1)})^\top
\end{equation}
is a thin SVD, then
\begin{equation}
    \m D^{(t+1)} = \m P^{(t+1)}(\m Q^{(t+1)})^\top .
    \label{eq:compot_procrustes}
\end{equation}

\paragraph{Lemma A.1 (exact sparse-coding update).}
For fixed \(\m D \in \mathrm{St}(m,k)\), the update
\(\m S^+ = \mathcal{H}_s(\m D^\top \widetilde{\m W})\) is a global minimizer of
\(\Phi(\m D,\m S)\) over \(\m S \in \mathcal{S}_s\).

\paragraph{Proof.}
This follows directly from the derivation in
\eqref{eq:apx_ortho_expand}--\eqref{eq:apx_sparse_matrix_update}: under
\(\m D^\top\m D=\m I_k\), sparse coding is equivalent to projecting
\(\m D^\top\widetilde{\m W}\) onto the set of column-wise \(s\)-sparse matrices.
This projection decouples over columns and is obtained by keeping the \(s\)
largest-magnitude entries in each column, with arbitrary tie-breaking when needed.
Hence \(\m S^+=\mathcal{H}_s(\m D^\top\widetilde{\m W})\) is globally optimal. \(\square\)

\paragraph{Lemma A.2 (exact orthogonal Procrustes update).}
For fixed \(\m S\), let \(\m M=\widetilde{\m W}\m S^\top\) and let
\(\m M=\m P\m \Lambda\m Q^\top\) be a thin SVD. Then
\begin{equation}
    \m D^+ = \m P\m Q^\top
\end{equation}
is a global minimizer of \(\Phi(\m D,\m S)\) over
\(\m D\in \mathrm{St}(m,k)\).

\paragraph{Proof.}
This follows directly from the derivation in
\eqref{eq:apx_procrustes}--\eqref{eq:apx_procrustes_solution}: for fixed
\(\m S\), minimizing \(\Phi(\m D,\m S)\) is equivalent to maximizing
\(\langle \m D,\widetilde{\m W}\m S^\top\rangle\) subject to
\(\m D^\top\m D=\m I_k\). By the classical orthogonal Procrustes theorem and
von Neumann's trace inequality, a global maximizer is
\(\m D^+=\m P\m Q^\top\). \(\square\)

\paragraph{Proposition A.3 (monotone objective-value convergence).}
The sequence of objective values
\begin{equation}
    \phi_t \equiv \Phi(\m D^{(t)},\m S^{(t)})
\end{equation}
generated by COMPOT is monotonically nonincreasing and convergent.

\paragraph{Proof.}
By Lemma~A.1, \(\m S^{(t+1)}\) globally minimizes the objective over the sparse-code
block while \(\m D^{(t)}\) is fixed. Hence
\begin{equation}
    \Phi(\m D^{(t)},\m S^{(t+1)})
    \le
    \Phi(\m D^{(t)},\m S^{(t)}).
    \label{eq:descent_s}
\end{equation}
By Lemma~A.2, \(\m D^{(t+1)}\) globally minimizes the objective over the orthogonal
dictionary block while \(\m S^{(t+1)}\) is fixed. Hence
\begin{equation}
    \Phi(\m D^{(t+1)},\m S^{(t+1)})
    \le
    \Phi(\m D^{(t)},\m S^{(t+1)}).
    \label{eq:descent_d}
\end{equation}
Combining \eqref{eq:descent_s} and \eqref{eq:descent_d} gives
\begin{equation}
    \Phi(\m D^{(t+1)},\m S^{(t+1)})
    \le
    \Phi(\m D^{(t)},\m S^{(t)})
    \qquad
    \text{for all } t .
\end{equation}
Since \(\Phi(\m D,\m S)\ge 0\), the monotone sequence \(\{\phi_t\}\) is bounded below
and therefore converges. \(\square\)

We next state a KL-based convergence guarantee for the iterates. Because the sparse
coding map can be discontinuous when the hard-thresholding support changes, and
because the Procrustes update can be non-unique when
\(\widetilde{\m W}\m S^\top\) is rank-deficient, the result is stated under the following
standard non-degeneracy conditions.

\paragraph{Assumption A.4 (non-degenerate regime).}
There exists an iteration \(T_0\) such that for all \(t\ge T_0\):
\begin{enumerate}
    \item the support pattern selected by
    \(\mathcal{H}_s((\m D^{(t)})^\top\widetilde{\m W})\)
    is fixed, so the sparse-code iterates lie in a fixed coordinate subspace;
    \item the Procrustes matrix
    \(\m M^{(t+1)}=\widetilde{\m W}(\m S^{(t+1)})^\top\) has full column rank, and
    its smallest singular value is uniformly bounded below:
    \(\sigma_{\min}(\m M^{(t+1)})\ge \mu>0\).
\end{enumerate}

\paragraph{Lemma A.5 (sufficient descent under Assumption~A.4).}
Under Assumption~A.4, for all \(t\ge T_0\) there exists \(a>0\) such that
\begin{equation}
    \Psi(\m D^{(t)},\m S^{(t)})-\Psi(\m D^{(t+1)},\m S^{(t+1)})
    \ge
    a\left(
    \|\m S^{(t+1)}-\m S^{(t)}\|_F^2
    +
    \|\m D^{(t+1)}-\m D^{(t)}\|_F^2
    \right).
    \label{eq:kl_sufficient_descent}
\end{equation}

\paragraph{Proof.}
For the sparse-code step, once the support pattern is fixed, the feasible sparse-code
set locally coincides with a fixed coordinate subspace. Since
\(\m D^{(t)}\in\mathrm{St}(m,k)\), the sparse-coding objective can be written as
\[
\Phi(\m D^{(t)},\m S)
=
\frac12\|\m S-(\m D^{(t)})^\top\widetilde{\m W}\|_F^2
+
\mathrm{const}.
\]
Thus the update \(\m S^{(t+1)}\) is the Euclidean projection onto a fixed linear subspace,
and the Pythagorean identity gives
\begin{equation}
    \Phi(\m D^{(t)},\m S^{(t)})-\Phi(\m D^{(t)},\m S^{(t+1)})
    \ge
    \frac12\|\m S^{(t+1)}-\m S^{(t)}\|_F^2 .
    \label{eq:s_descent_strong}
\end{equation}
For the dictionary step, let
\[
\m M^{(t+1)}=\widetilde{\m W}(\m S^{(t+1)})^\top
=\m P^{(t+1)}\m\Lambda^{(t+1)}(\m Q^{(t+1)})^\top
\]
and \(\m D^{(t+1)}=\m P^{(t+1)}(\m Q^{(t+1)})^\top\).
For fixed \(\m S^{(t+1)}\), the objective differs from
\(-\langle \m D,\m M^{(t+1)}\rangle\) only by constants. Therefore
\[
\Phi(\m D^{(t)},\m S^{(t+1)})-\Phi(\m D^{(t+1)},\m S^{(t+1)})
=
\langle \m D^{(t+1)}-\m D^{(t)},\m M^{(t+1)}\rangle .
\]
Writing
\(\m R=(\m P^{(t+1)})^\top \m D^{(t)}\m Q^{(t+1)}\), we have
\[
\langle \m D^{(t+1)}-\m D^{(t)},\m M^{(t+1)}\rangle
=
\sum_{\ell=1}^k \lambda_\ell^{(t+1)}(1-R_{\ell\ell})
\ge
\frac{\mu}{2}\|\m D^{(t+1)}-\m D^{(t)}\|_F^2 ,
\]
where the last inequality uses \(\lambda_\ell^{(t+1)}\ge \mu\) and
\[
\|\m D^{(t+1)}-\m D^{(t)}\|_F^2
=
2\sum_{\ell=1}^k(1-R_{\ell\ell}).
\]
Combining this bound with \eqref{eq:s_descent_strong} proves
\eqref{eq:kl_sufficient_descent} with \(a=\frac12\min\{1,\mu\}\). \(\square\)

\paragraph{Lemma A.6 (relative-error bound under Assumption~A.4).}
Under Assumption~A.4, for all \(t\ge T_0\) there exists \(b>0\) such that
\begin{equation}
    \mathrm{dist}\!\left(
    \m 0,\partial \Psi(\m D^{(t+1)},\m S^{(t+1)})
    \right)
    \le
    b\left(
    \|\m D^{(t+1)}-\m D^{(t)}\|_F
    +
    \|\m S^{(t+1)}-\m S^{(t)}\|_F
    \right),
    \label{eq:kl_relative_error}
\end{equation}
where \(\partial\Psi\) denotes the limiting subdifferential.

\paragraph{Proof.}
The Procrustes update is an exact minimizer of
\(\Phi(\m D,\m S^{(t+1)})\) over \(\mathrm{St}(m,k)\), hence the dictionary block satisfies
the first-order condition
\[
\m 0\in \nabla_{\m D}\Phi(\m D^{(t+1)},\m S^{(t+1)})
+
N_{\mathrm{St}(m,k)}(\m D^{(t+1)}).
\]
For the sparse-code block, under the fixed-support condition in Assumption~A.4,
\(\mathcal{S}_s\) locally coincides with a fixed coordinate subspace. Since
\(\m S^{(t+1)}\) minimizes \(\Phi(\m D^{(t)},\m S)\) over this subspace, there exists
\(\m N_S^{(t+1)}\in N_{\mathcal{S}_s}(\m S^{(t+1)})\) such that
\[
\m 0
=
\nabla_{\m S}\Phi(\m D^{(t)},\m S^{(t+1)})
+
\m N_S^{(t+1)}.
\]
Therefore, an element of \(\partial\Psi(\m D^{(t+1)},\m S^{(t+1)})\) is obtained by
replacing \(\nabla_{\m S}\Phi(\m D^{(t)},\m S^{(t+1)})\) with
\(\nabla_{\m S}\Phi(\m D^{(t+1)},\m S^{(t+1)})\). Since
\(\m D^{(t)},\m D^{(t+1)}\in\mathrm{St}(m,k)\),
\[
\nabla_{\m S}\Phi(\m D,\m S)
=
\m S-\m D^\top\widetilde{\m W},
\]
and hence
\[
\left\|
\nabla_{\m S}\Phi(\m D^{(t+1)},\m S^{(t+1)})
-
\nabla_{\m S}\Phi(\m D^{(t)},\m S^{(t+1)})
\right\|_F
=
\|(\m D^{(t)}-\m D^{(t+1)})^\top\widetilde{\m W}\|_F
\le
\|\widetilde{\m W}\|_F\|\m D^{(t+1)}-\m D^{(t)}\|_F .
\]
This proves \eqref{eq:kl_relative_error} with
\(b=\|\widetilde{\m W}\|_F\). \(\square\)

\paragraph{Proposition A.7 (KL property).}
The constrained objective \(\Psi\) in \eqref{eq:compot_nonsmooth} satisfies the
Kurdyka--\L{}ojasiewicz property.

\paragraph{Proof.}
The squared reconstruction term is polynomial in \((\m D,\m S)\). The Stiefel set
\(\mathrm{St}(m,k)\) is defined by polynomial equalities
\(\m D^\top\m D=\m I_k\), and the column-wise sparsity set
\(\mathcal{S}_s\) is a finite union of coordinate subspaces. Hence both constraint
sets are semi-algebraic, and \(\Psi\) is a proper lower-semicontinuous semi-algebraic
function. Proper lower-semicontinuous semi-algebraic functions satisfy the
Kurdyka--\L{}ojasiewicz property. \(\square\)

\paragraph{Theorem A.8 (conditional KL convergence of iterates).}
Under Assumption~A.4, the sequence
\(\{(\m D^{(t)},\m S^{(t)})\}_{t\ge T_0}\) has finite length and converges to a critical point
\((\m D^\star,\m S^\star)\) of \(\Psi\). Consequently,
\[
\Phi(\m D^{(t)},\m S^{(t)})\to \Phi(\m D^\star,\m S^\star).
\]

\paragraph{Proof.}
The sequence is bounded: \(\m D^{(t)}\in\mathrm{St}(m,k)\), and the sparse-code iterates
are obtained by hard thresholding \((\m D^{(t)})^\top\widetilde{\m W}\), hence are bounded.
By Proposition~A.7, \(\Psi\) satisfies the KL property. Lemma~A.5 gives sufficient
descent, and Lemma~A.6 gives the relative-error condition. Applying the standard
KL convergence theorem for nonsmooth descent methods yields finite length,
\[
\sum_{t=T_0}^{\infty}
\left(
\|\m D^{(t+1)}-\m D^{(t)}\|_F
+
\|\m S^{(t+1)}-\m S^{(t)}\|_F
\right)
<\infty,
\]
and therefore convergence of \((\m D^{(t)},\m S^{(t)})\) to a limiting point
\((\m D^\star,\m S^\star)\). The relative-error condition then implies
\(\m 0\in\partial\Psi(\m D^\star,\m S^\star)\), so the limit is a critical point.
The convergence of objective values follows from continuity of \(\Phi\) on the active
feasible manifold and Proposition~A.3. \(\square\)

\paragraph{Discussion.}
Proposition~A.3 holds unconditionally for the exact COMPOT updates with the SVD
initialization used in the main experiments. The stronger iterate-convergence result
in Theorem~A.8 follows under the stated non-degeneracy assumptions. These assumptions
exclude degenerate cases such as persistent hard-thresholding support switches or
rank-deficient Procrustes matrices, where standard KL sufficient-descent and
relative-error arguments may fail.
\subsection{Dynamic Compression Ratio Allocation}
\label{apx:dynamic_alloc}

Transformer models exhibit heterogeneous redundancy patterns across different layers and projection types. Uniform compression ratios often lead to suboptimal performance as sensitive layers may be over-compressed while redundant layers remain under-compressed. To address this challenge, we propose a one-shot global allocation strategy that distributes a model-wide compression budget while respecting per-matrix constraints.

Our allocation procedure operates on the singular values of weight matrices in the original, non-whitened space. In the main experiments, we normalize each weight matrix $\m W_i$ by its Frobenius norm before computing its spectrum,
$\bar{\m W}_i=\m W_i/\|\m W_i\|_F$.
This normalization equalizes spectral scales across layers and projection types. Consequently, the pooled singular-value rule should be interpreted as balancing relative reconstruction error across matrices, rather than as optimizing absolute Frobenius reconstruction error.

The core heuristic is simple: smaller singular values of the normalized matrices are treated as less important and are truncated first. By pooling singular values across the model, layer-specific compression ratios emerge from a single global selection rule. The global parameter budget is enforced separately by a binary search over the total truncation budget, using the rank-storage proxy $r_i(m_i+n_i)$ for a matrix of shape $m_i\times n_i$.

To prevent pathological allocation scenarios, we incorporate the following constraints:
\begin{itemize}
    \item \textbf{Minimum compression guard}: We impose a lower bound $cr_{\min}$ on each factorized matrix by converting it into an upper bound on the retained rank. This prevents layers from receiving negligible compression.
    \item \textbf{Maximum compression guard}: We impose an upper bound $cr_{\max}$ by converting it into a lower bound on the retained rank. This prevents over-compression of sensitive layers.
    \item \textbf{Non-beneficial factorization handling}: If the feasible rank interval does not admit a parameter-saving factorization under the proxy $r_i(m_i+n_i)<m_in_i$, the matrix is left dense and assigned $cr_i=0$.
\end{itemize}
In all main experiments, we use unwhitened Frobenius-normalized spectra with $cr_{\min}=0.05$ and $cr_{\max}=0.8$. With $cr_{\min}=0.05$, every factorized matrix is constrained to achieve at least $5\%$ parameter reduction under the rank-storage proxy, so the \textsc{dense} fallback is inactive in our main runs; it is retained only as a safety mechanism for other configurations.

Our allocation algorithm (Algorithm~\ref{alg:alloc_dense}) implements this deterministic, parameter-budgeted heuristic. First, it converts the compression guards into per-matrix rank and truncation bounds. Then, for a candidate global truncation budget, it allocates mandatory truncations induced by $cr_{\min}$ and greedily selects the smallest remaining singular values of Frobenius-normalized matrices while respecting the cap induced by $cr_{\max}$. Finally, a binary search chooses the smallest truncation budget whose implied parameter count satisfies the target model-wide compression ratio. We therefore do not claim optimality for the absolute Frobenius-error knapsack problem; the procedure is a practical heuristic for balancing relative spectral error under explicit storage constraints.

The computational complexity of our allocation strategy is dominated by the SVD computations for each weight matrix, which is $O(\sum_i \min(m_i,n_i)^2\max(m_i,n_i))$. However, since these computations are performed only once during compression preparation and can be parallelized across layers, the practical overhead remains modest. For a Llama3.2-1B model, the entire allocation procedure completes in under 1 minute on a single A100 GPU without any optimizations.

Our one-shot approach differs from iterative methods that require multiple rounds of compression and validation. By avoiding iterative search over layer-wise compression ratios, we eliminate the need for additional calibration data passes or performance estimation heuristics, making our method both deterministic and computationally efficient.

% ==========================================
% Algorithm 2: Allocation (dense + abstract)
% ==========================================
\begin{algorithm}[]
\caption{\textsc{Allocate-Global}: pooled-SV truncation with per-matrix CR guards}
\label{alg:alloc_dense}
\begin{algorithmic}[1]
\REQUIRE Matrices $\{\m W_i\in\mathbb{R}^{m_i\times n_i}\}_{i=1}^N$, target global compression ratio $cr$, guard bounds $(cr_{\min},cr_{\max})$.
\ENSURE Per-matrix compression ratios $\{cr_i\}$ and implied retained ranks $\{r_i\}$.

\STATE \textbf{Normalize and compute spectra:}
$\bar{\m W}_i\leftarrow \m W_i/\|\m W_i\|_F$; compute singular values
$\sigma_{i,1}\ge\cdots\ge\sigma_{i,L_i}$, where $L_i=\min(m_i,n_i)$.

\STATE \textbf{Convert CR guards to rank bounds:}
using the proxy storage model $r_i(m_i+n_i)$, define integer bounds
\[
r_i^{\min}\approx \frac{(1-cr_{\max})m_in_i}{m_i+n_i},
\qquad
r_i^{\max}\approx \frac{(1-cr_{\min})m_in_i}{m_i+n_i},
\]
with rounding to feasible integer ranks and clipping to $[0,L_i]$.

\STATE \textbf{Convert rank bounds to truncation bounds:}
$t_i=L_i-r_i$, giving
\[
t_i^{\min}=L_i-r_i^{\max},
\qquad
t_i^{\max}=L_i-r_i^{\min}.
\]

\STATE Initialize all matrices as active and set the \textsc{dense} mask to false.
\STATE Let $P_0=\sum_i m_in_i$ and $P_{\mathrm{tgt}}=(1-cr)P_0$.

\REPEAT
    \STATE Mark any active matrix as \textsc{dense} if no feasible rank satisfies
    $r_i(m_i+n_i)<m_in_i$.
    \STATE Let $\mathcal{A}$ be the remaining active matrices.
    \STATE Set binary-search bounds
    \[
    K_{\min}=\sum_{i\in\mathcal{A}} t_i^{\min},
    \qquad
    K_{\max}=\sum_{i\in\mathcal{A}} t_i^{\max}.
    \]

    \WHILE{$K_{\min}<K_{\max}$}
        \STATE Set candidate truncation budget
        $\widehat K=\lfloor(K_{\min}+K_{\max})/2\rfloor$.
        \STATE Allocate mandatory truncations $t_i\leftarrow t_i^{\min}$ for all $i\in\mathcal{A}$.
        \STATE From the remaining singular values of active matrices, greedily select the smallest values until the candidate budget $\widehat K$ is reached, while enforcing $t_i\le t_i^{\max}$.
        \STATE Set $r_i(\widehat K)=L_i-t_i(\widehat K)$ and compute
        \[
        P(\widehat K)=
        \sum_{i\in\textsc{dense}} m_in_i
        +
        \sum_{i\in\mathcal{A}} r_i(\widehat K)(m_i+n_i).
        \]
        \IF{any active matrix satisfies $r_i(\widehat K)(m_i+n_i)\ge m_in_i$}
            \STATE Reclassify it as \textsc{dense} and restart the outer loop.
        \ELSIF{$P(\widehat K)\le P_{\mathrm{tgt}}$}
            \STATE $K_{\max}\leftarrow \widehat K$.
        \ELSE
            \STATE $K_{\min}\leftarrow \widehat K+1$.
        \ENDIF
    \ENDWHILE

    \STATE Set $K^\star=K_{\max}$ and recompute $\{t_i\}$ using the same constrained greedy selection.
\UNTIL{no additional matrix is reclassified as \textsc{dense}}

\STATE \textbf{Return per-matrix ratios:}
for non-dense matrices, set
\[
cr_i\leftarrow 1-\frac{r_i(m_i+n_i)}{m_in_i};
\]
for dense matrices, set $cr_i\leftarrow 0$.
\end{algorithmic}
\end{algorithm}
\subsection{Comparison with Low-rank and Sparse Dictionary Learning Methods on Small Models}
\label{apx:small_models}
In this section we provide comparison of COMPOT method across different models: Qwen3-0.6B and Llama3.2-1B across different compression ratios to verify its scalability. The corresponding results are presented in Tables~\ref{tab:qwen3_0.6b} and~\ref{tab:llama_32_1b}. For small models COMPOT in both static and dynamic CR cases outperforms SVD-based and dictionary learning based methods by a wide margin.

\begin{table}[]
\caption{Performance comparison under static (COMPOT\txtdag) and dynamic (COMPOT) vs state-of-the-art SVD-based SVD-LLM and dictionary-learning CoSpaDi on Llama3.2-1B at different compression levels on different benchmarks. Best results are shown in \textbf{bold}.}
\label{tab:llama_32_1b}
\resizebox{\textwidth}{!}{%
\renewcommand{\arraystretch}{1.05}
\begin{tabular}{ccccccccccccc}
\hline
                            &                         & \multicolumn{9}{c}{\textbf{Accuracy$\uparrow$}}                                                                           & \multicolumn{2}{c}{\textbf{Perplexity$\downarrow$}}       \\ \cline{3-13}
\multirow{-2}{*}{\textbf{Method}}    & \multirow{-2}{*}{\textbf{CR}}    & \textbf{PIQA} & \textbf{Hella Swag} & \textbf{LAMBADA} & \textbf{ARC-e} & \textbf{ARC-c} & \textbf{SciQ} & \textbf{Race} & \textbf{MMLU} & \textbf{Avg.} & \textbf{WikiText} & \textbf{LAMBADA} \\ \hline \hline

\textbf{Llama3.2-1B} & --                   & 74.5 & 63.7 & 63.0 & 60.5 & 36.2 & 88.3 & 37.8 & 37.0 & 57.6 & 1.2E+01 & 5.7E+00 \\ \hline

SVD-LLM             &                       & 62.1 & 36.4 & 24.4 & 36.0 & 25.1 & 64.9 & 29.0 & 23.0 & 37.6 & 1.7E+02 & 1.7E+02 \\
CoSpaDi             &                       & 66.1 & 42.9 & 38.4 & 39.9 & 26.0 & 71.6 & 31.7 & 24.8 & 42.7 & 6.4E+01 & 3.5E+01 \\
COMPOT\txtdag       &                       & 69.7 & 52.1 & 42.1 & 49.9 & 30.2 & 81.8 & 33.9 & 28.5 & 48.5 & 2.5E+01 & 2.1E+01 \\
COMPOT              & \multirow{-4}{*}{0.2} & 70.6 & 54.6 & 47.6 & 52.5 & 31.1 & 81.6 & 35.7 & 26.7 & \textbf{50.1} & \textbf{2.1E+01} & \textbf{1.4E+01} \\ \hline

SVD-LLM             &                       & 55.7 & 30.1 & 9.1  & 30.5 & 21.5 & 45.9 & 25.8 & 23.2 & 30.2 & 5.9E+02 & 2.5E+03 \\
CoSpaDi             &                       & 56.9 & 32.4 & 18.2 & 31.9 & 22.1 & 56.7 & 28.0 & 23.1 & 33.7 & 2.9E+02 & 6.6E+02 \\
COMPOT\txtdag       &                       & 64.2 & 41.0 & 27.5 & 39.8 & 26.0 & 73.2 & 31.4 & 23.5 & 40.8 & 5.5E+01 & 8.4E+01 \\
COMPOT              & \multirow{-4}{*}{0.3} & 67.0 & 46.8 & 37.9 & 47.1 & 27.1 & 77.3 & 33.1 & 23.8 & \textbf{45.0} & \textbf{3.6E+01} & \textbf{3.2E+01} \\ \hline

SVD-LLM             &                       & 51.8 & 27.3 & 1.3  & 26.9 & 22.9 & 32.3 & 24.4 & 23.0 & 26.2 & 1.6E+03 & 3.3E+04 \\
CoSpaDi             &                       & 53.5 & 28.2 & 3.8  & 27.8 & 23.0 & 36.9 & 24.0 & 23.1 & 27.5 & 8.0E+02 & 9.2E+03 \\
COMPOT\txtdag       &                       & 57.0 & 31.1 & 7.9  & 31.7 & 22.1 & 53.0 & 27.0 & 23.1 & 31.6 & 2.8E+02 & 2.2E+03 \\
COMPOT              & \multirow{-4}{*}{0.4} & 62.9 & 37.5 & 21.3 & 38.0 & 25.3 & 62.4 & 30.6 & 23.0 & \textbf{37.6} & \textbf{1.3E+02} & \textbf{1.9E+02} \\ \hline

\end{tabular}
}
\end{table}

\begin{table}[]
\caption{Performance comparison under static (COMPOT\txtdag) and dynamic (COMPOT) vs state-of-the-art SVD-based SVD-LLM and dictionary-learning CoSpaDi on Qwen3-0.6B at different compression levels on different benchmarks. Best results are shown in \textbf{bold}.}
\label{tab:qwen3_0.6b}
\resizebox{\textwidth}{!}{%
\renewcommand{\arraystretch}{1.05}
\begin{tabular}{ccccccccccccc}
\hline
                            &                         & \multicolumn{9}{c}{\textbf{Accuracy$\uparrow$}}                                                                           & \multicolumn{2}{c}{\textbf{Perplexity$\downarrow$}}       \\ \cline{3-13}
\multirow{-2}{*}{\textbf{Method}}    & \multirow{-2}{*}{\textbf{CR}}    & \textbf{PIQA} & \textbf{Hella Swag} & \textbf{LAMBADA} & \textbf{ARC-e} & \textbf{ARC-c} & \textbf{SciQ} & \textbf{Race} & \textbf{MMLU} & \textbf{Avg.} & \textbf{WikiText} & \textbf{LAMBADA} \\ \hline \hline
\textbf{Qwen3-0.6B} & -- & 67.5 & 47.4 & 40.0 & 55.6 & 34.3 & 83.5 & 33.7 & 40.2 & 50.3 & 2.6E+01 & 2.5E+01 \\ \hline

SVD-LLM         &                       & 57.0 & 33.0 & 12.6 & 33.6 & 24.8 & 54.9 & 29.5 & 26.1 & 33.9 & 3.1E+03 & 7.3E+03 \\
CoSpaDi         &                       & 59.3 & 37.5 & 18.7 & 37.2 & 27.1 & 68.6 & 32.4 & 26.6 & 38.4 & 1.9E+03 & 9.4E+02 \\
COMPOT\txtdag   &                       & 62.2 & 40.6 & 15.9 & 41.3 & 29.6 & 70.8 & 33.0 & 26.3 & 40.0 & 1.4E+03 & 4.2E+02 \\
COMPOT          & \multirow{-4}{*}{0.2} & 61.2 & 42.6 & 20.6 & 42.9 & 28.8 & 74.0 & 32.3 & 24.9 & \textbf{40.9} & \textbf{7.0E+01} & \textbf{1.9E+02} \\ \hline

SVD-LLM         &                       & 52.2 & 29.0 & 8.7  & 28.2 & 23.8 & 42.1 & 29.1 & 24.4 & 29.7 & 1.9E+04 & 1.3E+05 \\
CoSpaDi         &                       & 56.4 & 32.3 & 14.3 & 31.0 & 24.0 & 57.4 & 31.0 & 24.6 & 33.9 & 3.8E+03 & 1.1E+04 \\
COMPOT\txtdag   &                       & 58.2 & 34.8 & 13.6 & 34.1 & 27.1 & 59.9 & 31.7 & 26.3 & 35.7 & 3.1E+03 & 3.6E+03 \\
COMPOT          & \multirow{-4}{*}{0.3} & 61.3 & 38.3 & 20.8 & 40.5 & 27.1 & 65.2 & 32.7 & 27.2 & \textbf{39.1} & \textbf{2.5E+02} & \textbf{3.8E+02} \\ \hline

SVD-LLM         &                       & 53.3 & 26.9 & 3.8  & 26.5 & 23.9 & 32.5 & 25.4 & 23.6 & 27.0 & 7.2E+04 & 5.4E+06 \\
CoSpaDi         &                       & 51.6 & 28.2 & 7.9  & 27.3 & 24.1 & 43.8 & 27.0 & 24.0 & 29.2 & 1.2E+04 & 2.4E+05 \\
COMPOT\txtdag   &                       & 54.5 & 29.7 & 8.4  & 28.8 & 23.6 & 44.3 & 28.3 & 23.5 & 30.1 & 6.7E+03 & 1.2E+05 \\
COMPOT          & \multirow{-4}{*}{0.4} & 56.9 & 32.7 & 17.5 & 31.7 & 25.1 & 57.4 & 31.2 & 24.7 & \textbf{34.7} & \textbf{3.8E+03} & \textbf{5.0E+03} \\ \hline

\end{tabular}
}
\end{table}
\subsection{Evaluation on Recent Benchmarks}
\label{apx:more_benchs}

To validate our method, we report results on a set of recent and more challenging benchmarks that better reflect modern LLM evaluation. We include Open LLM Leaderboard v2 as an aggregate, community-maintained reference \cite{open-llm-leaderboard-v2}, which includes IFEval \cite{zhou2023instructionfollowingevaluationlargelanguage}, MATH \cite{hendrycks2021measuringmathematicalproblemsolving}, GPQA \cite{rein2023gpqagraduatelevelgoogleproofqa}, BBH \cite{suzgun2022challengingbigbenchtaskschainofthought} and MUSR \cite{sprague2024musrtestinglimitschainofthought}, ensuring that observed improvements generalize beyond a single evaluation suite. See Table~\ref{tab:new_benchmarks} for detailed results.

\begin{table}[]
\caption{Performance comparison under static (COMPOT\txtdag) and dynamic (COMPOT) vs SVD-LLM on Open LLM Leaderboard v2 \cite{open-llm-leaderboard-v2}}
\label{tab:new_benchmarks}
\centering
\resizebox{0.7\textwidth}{!}{%
\renewcommand{\arraystretch}{1.05}
\begin{tabular}{cccccccc}
\hline
                            &                         \multicolumn{7}{c}{\textbf{Accuracy$\uparrow$}}\\ \cline{3-8}
\multirow{-2}{*}{\textbf{Method}}    & \multirow{-2}{*}{\textbf{CR}}    & \textbf{BBH}& \textbf{GPQA}& \textbf{IFEVAL}& \textbf{MATH HARD}& \textbf{MMLU Pro}& \textbf{MUSR}\\ \hline \hline
\textbf{Qwen3-0.6B} & -- & 36.85& 28.27& 27.58& 12.69& 24.26& 31.35\\ \hline

SVD-LLM         &                       & 28.95& 24.75& 25.3& 0& 11.29& 35.58\\
COMPOT\txtdag   &                       & 29.2& 27.94& 27.22& 1.06& 12.81& 35.19\\
COMPOT          & \multirow{-3}{*}{0.2} & 30.19& 25.34& 25.9& 0.91& 14.81& 33.2\\ \hline

SVD-LLM         &                       & 29.73& 25.5& 24.34& 0& 11.22& 33.73\\
COMPOT\txtdag   &                       & 30.83& 23.49& 25.06& 0.23& 10.85& 32.8\\
COMPOT          & \multirow{-3}{*}{0.3}& 29.99& 26.59& 27.7& 0.68& 11.43& 35.71\\ \hline

SVD-LLM         &                       & 29.01& 24.33& 22.3& 0& 10.98& 36.24\\
COMPOT\txtdag   &                       & 29.21& 25.84& 24.58& 0.98& 10.85& 36.51\\
COMPOT          & \multirow{-3}{*}{0.4}& 29.06& 23.74& 26.62& 0.38& 10.66& 33.73\\ \hline\hline
 
 \textbf{Qwen3-8B}& -- & 60.84& 36.33& 39.21& 52.49& 47.62& 43.12\\ \hline

SVD-LLM         &                       &  41.03&  28.27&  25.66&  1.06&  26.3&  39.81\\
COMPOT\txtdag   &                       & 49.33& 30.54& 30.46& 6.72& 35.92& 45.11\\
COMPOT          & \multirow{-3}{*}{0.2} & 50.22& 31.96& 33.45& 12.76& 38.55& 41.8\\ \hline
 SVD-LLM         & & 34.3& 25.59& 22.66& 1.06& 18.73& 41.4\\
 COMPOT\txtdag   & & 41.45& 28.86& 26.74& 1.51& 29.46& 40.61\\
 COMPOT          & \multirow{-3}{*}{0.3}& 44.87& 28.94& 31.89& 2.49& 32.85& 40.34\\ \hline
 SVD-LLM         & & 30.24& 22.99& 22.42& 0.83& 11.54& 37.57\\
 COMPOT\txtdag   & & 34.13& 24.75& 23.14& 1.06& 21.23& 41.27\\
 COMPOT          & \multirow{-3}{*}{0.4}& 37.51& 28.1& 26.14& 0.98& 26.44& 38.89\\ \hline

\end{tabular}
}
\end{table}
\subsection{Wall-clock Time for Optimization}
\label{apx:wallclock}

In Table~\ref{tab:wall_clock_llama_1b} we report the wall-clock time required to compress a \emph{single} layer of Llama 3.2--1B under a static compression ratio of $0.2$ with $k/s = 2$. All measurements were obtained on a single NVIDIA A100 GPU; note that different layers are independent, so the procedure is trivially parallelizable across processes or devices.

For CoSpaDi estimation, we follow the original paper and implement K-SVD-based dictionary learning, i.e., we replace the orthogonal Procrustes (OPP) update in COMPOT with a K-SVD update. In line with CoSpaDi, we use power iterations instead of a full SVD for dictionary updates, with $60$ K-SVD iterations and $8$ power iterations per update. For COMPOT we use $20$ alternating iterations, which is the configuration used for all main results unless otherwise stated.

We observe that SVD-LLM is substantially faster than both CoSpaDi and COMPOT, reflecting the computational simplicity of its single truncated SVD factorization per matrix. Nevertheless, COMPOT achieves significantly lower wall-clock time than CoSpaDi: even \emph{per iteration}, COMPOT's closed-form Orthogonal Procrustes Problem (OPP) update (thin SVD on an $m \times k$ matrix) combined with hard-thresholding sparse coding is approximately $2\times$ faster than CoSpaDi's K-SVD dictionary updates with power iterations and OMP sparse coding. The computational advantage widens further when accounting for CoSpaDi's typically higher iteration requirements.

A key insight behind this efficiency is that when the dictionary is constrained to be orthonormal (as in COMPOT), hard-thresholding yields identical sparse coding solutions to Orthogonal Matching Pursuit (OMP) in theory, while being substantially faster in practice. This equivalence holds except in rare cases involving numerical instabilities at identical coefficient magnitudes.

This computational profile highlights a fundamental advantage of orthogonal dictionary learning for large-scale transformer compression: it preserves the modeling flexibility of sparse, union-of-subspaces representations while maintaining optimization overhead comparable to SVD-style methods. The result is a favorable trade-off between compression performance and computational cost.

\begin{table}[h]
\centering
\caption{Wall-clock time in seconds required for optimization with SVD-LLM, CoSpaDi and COMPOT for Llama3.2-1B with static 0.2 compression ratio.}
\label{tab:wall_clock_llama_1b}
\resizebox{\textwidth}{!}{%
\renewcommand{\arraystretch}{1.05}
\begin{tabular}{cccccc}
\hline
Layer & Dimensions & SVD-LLM & CoSpaDi & COMPOT & Speedup over CoSpaDi \\
\hline
model.layers.0.mlp.up\_proj & (2048, 8192)      & 1.96  & 508.82 & 17.84 & 28.52x \\
model.layers.0.mlp.down\_proj & (8192, 2048)    & 13.99 & 255.51 & 19.23 & 13.29x \\
model.layers.0.mlp.gate\_proj & (2048, 8192)    & 2.19  & 522.17 & 17.78 & 29.37x \\
model.layers.0.self\_attn.q\_proj & (2048, 2048)& 1.98  & 154.01 &  5.37 & 28.68x \\
model.layers.0.self\_attn.k\_proj & (2048, 512) & 1.03  &  44.70 &  1.99 & 22.41x \\
model.layers.0.self\_attn.v\_proj & (2048, 512) & 0.50  &  44.56 &  1.77 & 25.12x \\
model.layers.0.self\_attn.o\_proj & (2048, 2048)& 1.21  & 153.92 &  5.49 & 28.06x \\
\hdashline
\textbf{AVERAGE} &  & \textbf{3.27} & \textbf{240.53} & \textbf{9.92} & \textbf{24.23x} \\
\hline
\end{tabular}
}
\end{table}
\subsection{Computational Efficiency and Acceleration}
\label{apx:acceleration}

\paragraph{Inference complexity.}
For a projection $\m W\in\mathbb{R}^{m\times n}$ and $N$ input tokens, a rank-$r$
SVD factorization requires
\begin{equation}
    \mathcal{C}_{\mathrm{SVD}}
    =
    Nr(m+n)
\end{equation}
multiplications. COMPOT evaluates $\m X\m A\m S_O$, where
$\m A\in\mathbb{R}^{m\times k}$ contains $k$ dictionary atoms and each column of
$\m S_O\in\mathbb{R}^{k\times n}$ contains at most $s$ nonzero coefficients, giving
\begin{equation}
    \mathcal{C}_{\mathrm{COMPOT}}
    =
    N(mk+sn).
\end{equation}
Here, $k/s$ denotes the ratio between the total number of dictionary atoms and the
number activated for reconstructing one weight column. When SVD and COMPOT are
configured for the same value-storage budget,
\begin{equation}
    r(m+n)=mk+sn,
\end{equation}
and therefore their nominal multiplication counts are identical. Under the complete
storage accounting of Eq.~\eqref{eq:compression_ratio}, the binary support mask occupies
part of the COMPOT storage budget without adding arithmetic, resulting in slightly
smaller factors. Thus, COMPOT and SVD have comparable theoretical arithmetic
complexity at matched storage. Actual latency additionally depends on sparse metadata,
memory access, batching, projection shapes, and kernel efficiency.
\paragraph{Practical inference efficiency.}
To evaluate whether this theoretical efficiency translates into practical gains, we
use MACKO~\citep{macko2026macko}, a compressed sparse representation and GPU SpMV kernel
designed for low-to-moderate unstructured sparsity and previously used for the
deployment evaluation of CoSpaDi~\citep{shopkhoev2025cospadi}. We reproduce the same
setup using Llama3-8B on an NVIDIA RTX 3090, compressing only the MLP up and gate
projections at $\mathrm{CR}=0.6$ and $k/s=2$ with the same MACKO backend. We evaluate
prompt lengths of $1$, $16$, $32$, $64$, and $128$ with $1$, $4$, $8$, and $16$
concurrent users.
\begin{table}[ht]
\centering
\caption{End-to-end inference throughput of COMPOT relative to the original dense model
and CoSpaDi on Llama3-8B using the MACKO backend. Values above $1$ indicate higher
throughput for COMPOT.}
\label{tab:appendix_macko_inference}
\setlength{\tabcolsep}{8pt}
\renewcommand{\arraystretch}{1.08}
\begin{tabular}{cccc}
\hline
\textbf{Prompt length} &
\textbf{Concurrent users} &
\textbf{COMPOT / Dense} &
\textbf{COMPOT / CoSpaDi} \\
\hline \hline
1   & 1  & 1.340$\times$ & 1.079$\times$ \\
1   & 4  & 1.373$\times$ & 1.060$\times$ \\
1   & 8  & 1.350$\times$ & 1.049$\times$ \\
1   & 16 & 1.136$\times$ & 1.008$\times$ \\
\hline
16  & 1  & 1.214$\times$ & 1.103$\times$ \\
16  & 4  & 1.201$\times$ & 1.102$\times$ \\
16  & 8  & 1.203$\times$ & 1.112$\times$ \\
16  & 16 & 1.163$\times$ & 1.066$\times$ \\
\hline
32  & 1  & 1.198$\times$ & 1.101$\times$ \\
32  & 4  & 1.223$\times$ & 1.106$\times$ \\
32  & 8  & 1.191$\times$ & 1.076$\times$ \\
32  & 16 & 1.182$\times$ & 1.092$\times$ \\
\hline
64  & 1  & 1.194$\times$ & 1.000$\times$ \\
64  & 4  & 1.202$\times$ & 0.984$\times$ \\
64  & 8  & 1.181$\times$ & 0.953$\times$ \\
64  & 16 & 1.188$\times$ & 1.051$\times$ \\
\hline
128 & 1  & 1.172$\times$ & 1.028$\times$ \\
128 & 4  & 1.199$\times$ & 1.019$\times$ \\
128 & 8  & 1.167$\times$ & 1.013$\times$ \\
128 & 16 & 1.142$\times$ & 0.960$\times$ \\
\hline
\textbf{Average} & -- &
\textbf{1.211$\times$} &
\textbf{1.048$\times$} \\
\hline
\end{tabular}
\end{table}
COMPOT outperforms the dense model in all $20$ configurations, with throughput gains
ranging from $13.6\%$ to $37.3\%$. It matches or outperforms CoSpaDi in $17/20$
configurations and improves throughput by $4.8\%$ on average over the complete sweep.
In the three configurations where CoSpaDi is faster, COMPOT remains within $4.7\%$.
Overall, COMPOT improves throughput by $21.1\%$ on average over dense inference.
These results show that COMPOT's storage and arithmetic reduction translates into
consistent end-to-end acceleration over the original dense model when paired with a
kernel designed for its sparsity regime. Since COMPOT and CoSpaDi use the same
dense-dictionary--sparse-coefficient inference operator, we do not attribute their
runtime difference directly to orthogonality. Rather, the experiment shows that
COMPOT's improved compression quality and substantially faster offline optimization
are achieved without sacrificing practical deployment efficiency.

\paragraph{Hardware-oriented 2:4 sparsity.}
We additionally evaluate COMPOT with hardware-oriented 2:4 coefficient sparsity.
With dynamic allocation, it achieves average accuracies of $42.92$, $38.98$, and
$30.58$ at CR $=0.2$, $0.3$, and $0.4$, respectively, compared with $37.62$,
$30.22$, and $26.21$ for SVD-LLM. Thus, COMPOT retains its advantage under a
hardware-oriented semi-structured sparsity constraint. Joint optimization of the
dictionary size, sparsity level, layer-wise allocation, and sparse backend is left
for future work.

\paragraph{Early stopping for compression.}

While the main experiments use a fixed budget of $T=20$ alternating minimization
steps, we also study an optional early-stopping mechanism to reduce compression time.
The stopping rule is based on the relative Frobenius reconstruction error
\[
\rho_t =
\frac{\|\widetilde{\m W}-\m D_O^{(t)}\m S_O^{(t)}\|_F}{\|\widetilde{\m W}\|_F},
\]
where $\widetilde{\m W}$ is the whitened weight matrix optimized by COMPOT. Let $\tau$
denote the early-stopping tolerance and $p$ the patience. At iteration $t$, we compute
the relative improvement between consecutive iterations as
\[
\Delta_j =
\frac{\rho_{j-1}-\rho_j}{\rho_{j-1}},
\qquad j=t-p+1,\ldots,t .
\]
Early stopping is triggered if none of the most recent $p$ intervals improves by at
least $\tau$, i.e.,
\[
\max_{j\in\{t-p+1,\ldots,t\}} \Delta_j < \tau .
\]
If $\rho_{j-1}=0$, we set $\Delta_j=0$ to avoid division by zero. In the early-stopping
ablation, we use patience $p=3$ and set the maximum number of iterations to
$T_{\max}=150$.

The results in Table~\ref{tab:appendix_early_stopping} illustrate the influence of the
relative tolerance $\tau$ on the final performance of the framework under this optional
early-stopping rule. In our experimental setup, the dictionary is initialized with left
singular vectors and the execution time corresponds to setup with 4 GPUs and 2 processes each in parallel. We observe that lower tolerance thresholds generally improve final
accuracy and perplexity, as they allow additional alternating minimization steps before
termination.

\begin{table}[ht]
\centering
\caption{Performance of COMPOT under varying relative tolerance ($\tau$) thresholds for early stopping. Results are evaluated using Llama 3.2-1B at 20\% compression with dictionary-to-sparsity ratio $k/s=2.0$. The maximum number of iterations is set to 150.}
\label{tab:appendix_early_stopping}
\resizebox{\textwidth}{!}{%
\setlength{\tabcolsep}{2pt}
\renewcommand{\arraystretch}{1.05}
\small
\begin{tabular}{c c c c c c c c c c c c c c c c c c c c}
\hline
\multirow{2}{*}{\textbf{Tolerance}} & \multirow{2}{*}{\textbf{Time (s)}} & \multicolumn{7}{c}{\textbf{Avg. Iters. per Layer}} & \multicolumn{8}{c}{\textbf{Accuracy$\uparrow$}} &  & \multicolumn{2}{c}{\textbf{Perplexity$\downarrow$}} \\ \cline{3-20}
& & \textbf{Up} & \textbf{Down} & \textbf{Gate} & \textbf{Q} & \textbf{K} & \textbf{V} & \textbf{O} & \textbf{PIQA} & \textbf{HellaSwag} & \textbf{LAMBADA} & \textbf{ARC-e} & \textbf{ARC-c} & \textbf{SciQ} & \textbf{Race} & \textbf{MMLU} & \textbf{Avg.} & \textbf{WikiText} & \textbf{LAMBADA} \\ \hline \hline

Non-compressed& --    & --    & --    & --    & --    & --    & --    & --    & 74.5 & 63.7 & 63.0 & 60.5 & 36.2 & 88.3 & 37.8 & 37.0 & 57.6 & 11.57 & 5.73  \\ \hline
$10^{-1.0}$ & 203.4 &  4.0 &  4.0 &  4.0 &  4.0 &  5.0 &  4.5 &  4.0 & 69.4 & 52.9 & 45.3 & 51.1 & 30.1 & 81.4 & 34.2 & 25.9 & 48.8 & 22.45 & 16.50 \\
$10^{-1.5}$ & 207.1 &  7.6 &  5.0 &  6.3 &  5.0 &  7.3 &  6.3 &  5.9 & 69.9 & 53.5 & 45.8 & 51.4 & 30.9 & 82.1 & 34.8 & 26.1 & 49.3 & 21.94 & 15.76 \\
$10^{-2.0}$ & 246.8 & 15.6 &  7.6 & 12.2 &  7.5 & 13.8 & 10.8 & 10.2 & 70.8 & 53.8 & 47.3 & 50.8 & 30.3 & 82.2 & 35.4 & 26.3 & 49.6 & 21.20 & 14.52 \\
$10^{-2.5}$ & 342.4 & 32.5 & 13.4 & 25.4 & 14.1 & 28.3 & 19.9 & 19.9 & 70.9 & 54.6 & 48.1 & 51.7 & 30.2 & 81.9 & 34.8 & 26.1 & 49.8 & 20.56 & 13.88 \\
$10^{-3.0}$ & 504.5 & 64.3 & 24.9 & 50.6 & 28.5 & 56.0 & 36.9 & 39.1 & 71.7 & 54.7 & 47.8 & 53.1 & 30.6 & 83.2 & 34.7 & 26.2 & 50.3 & 20.18 & 13.80 \\
$10^{-3.5}$ & 810.5 &119.6 & 46.2 & 95.3 & 55.6 &107.6 & 67.9 & 73.4 & 71.5 & 54.9 & 47.6 & 53.4 & 31.7 & 82.6 & 34.3 & 26.2 & 50.3 & 19.93 & 13.79 \\
$10^{-4.0}$ &1056.9 &148.8 & 83.7 &146.7 &106.4 &150.0 &121.5 &131.6 & 71.7 & 55.2 & 47.6 & 52.6 & 31.7 & 82.1 & 34.1 & 26.5 & 50.2 & 19.86 & 14.06 \\ \hline
\end{tabular}
}
\end{table}
\subsection{Allocation Strategy Results}
\label{apx:allocation}
\begin{figure}[!ht]
  % \vskip 0.2in
  \begin{center}
    \centerline{\includegraphics[width=0.85\textwidth]{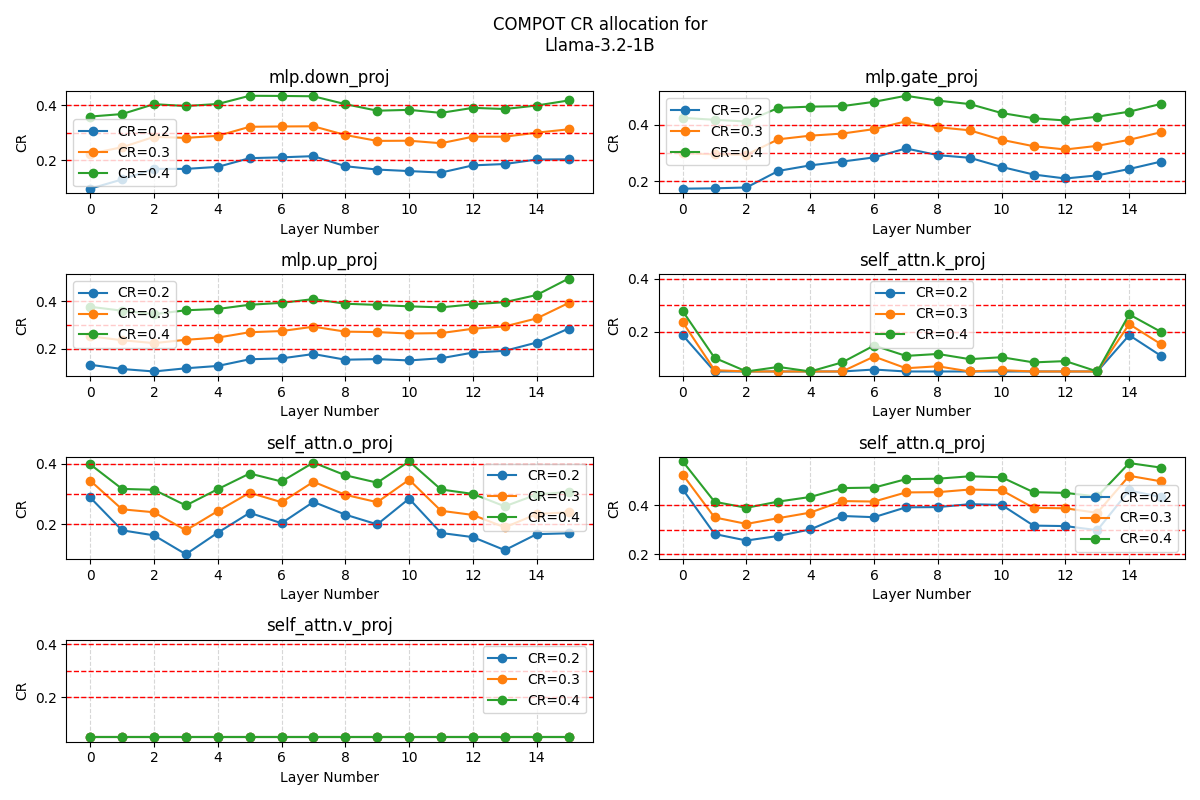}}
    \caption{Results of COMPOT allocation strategy for Llama3.2-1B}
    \label{fig:alloc_llama32_1b}
  \end{center}
\end{figure}

\begin{figure}[!ht]
  % \vskip 0.2in
  \begin{center}
    \centerline{\includegraphics[width=0.85\textwidth]{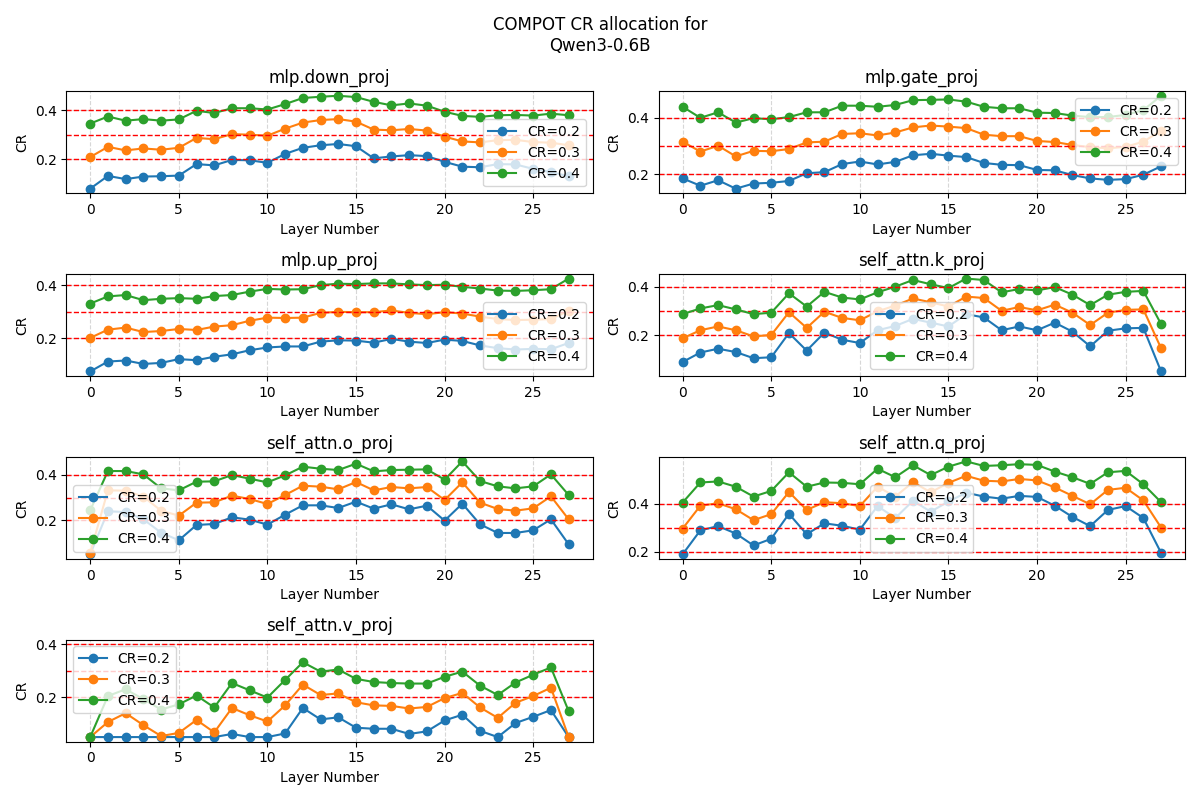}}
    \caption{Results of COMPOT allocation strategy for Qwen3-0.6B}
    \label{fig:alloc_qwen3_06b}
  \end{center}
\end{figure}

\begin{figure}[!ht]
  % \vskip 0.2in
  \begin{center}
    \centerline{\includegraphics[width=0.85\textwidth]{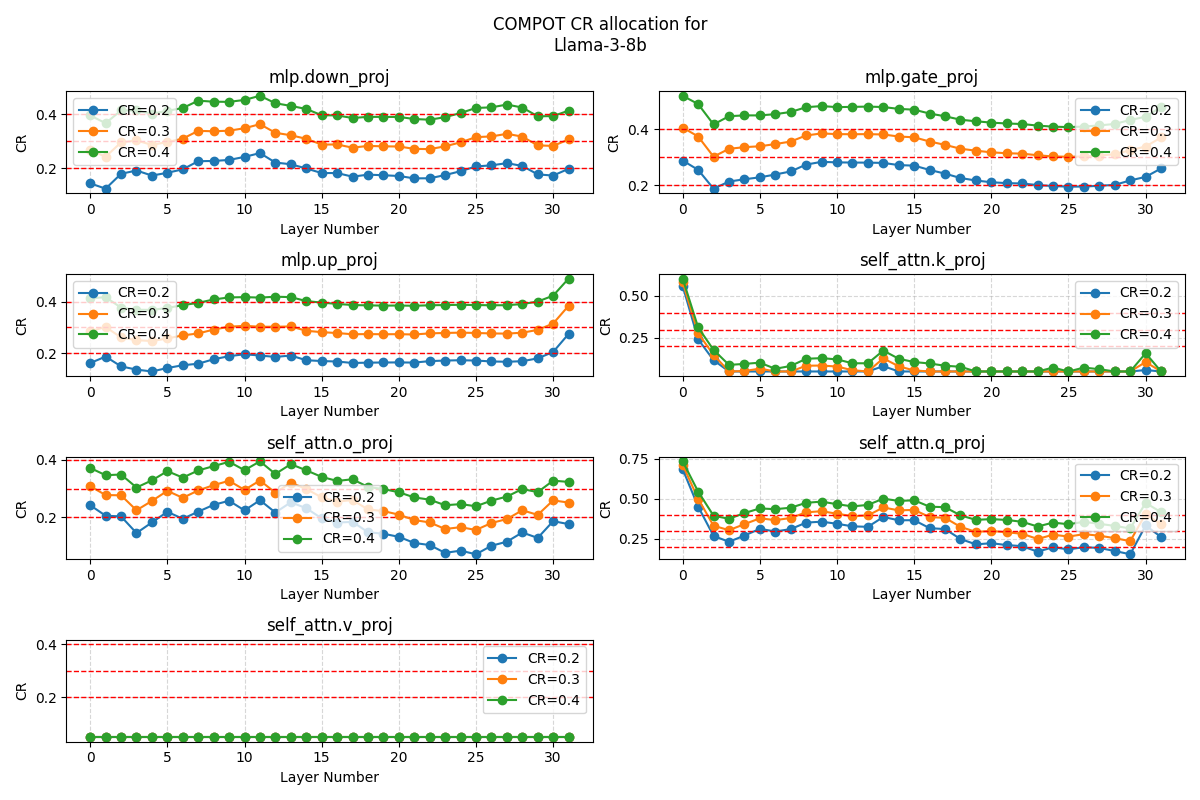}}
    \caption{Results of COMPOT allocation strategy for Llama3-8B}
    \label{fig:alloc_llama3_8b}
  \end{center}
\end{figure}

\begin{figure}[!ht]
  % \vskip 0.2in
  \begin{center}
    \centerline{\includegraphics[width=0.85\textwidth]{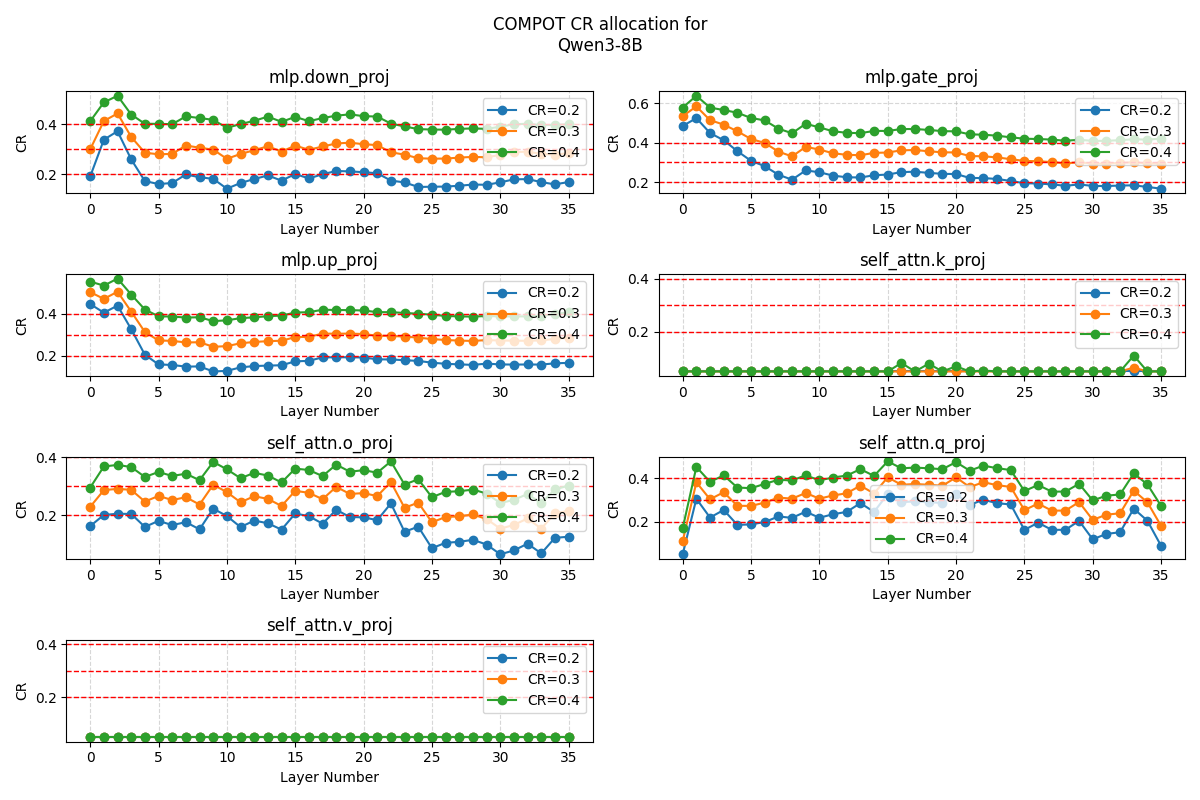}}
    \caption{Results of COMPOT allocation strategy for Qwen3-8B}
    \label{fig:alloc_qwen3_8b}
  \end{center}
\end{figure}

\begin{figure}[!ht]
  % \vskip 0.2in
  \begin{center}
    \centerline{\includegraphics[width=0.85\textwidth]{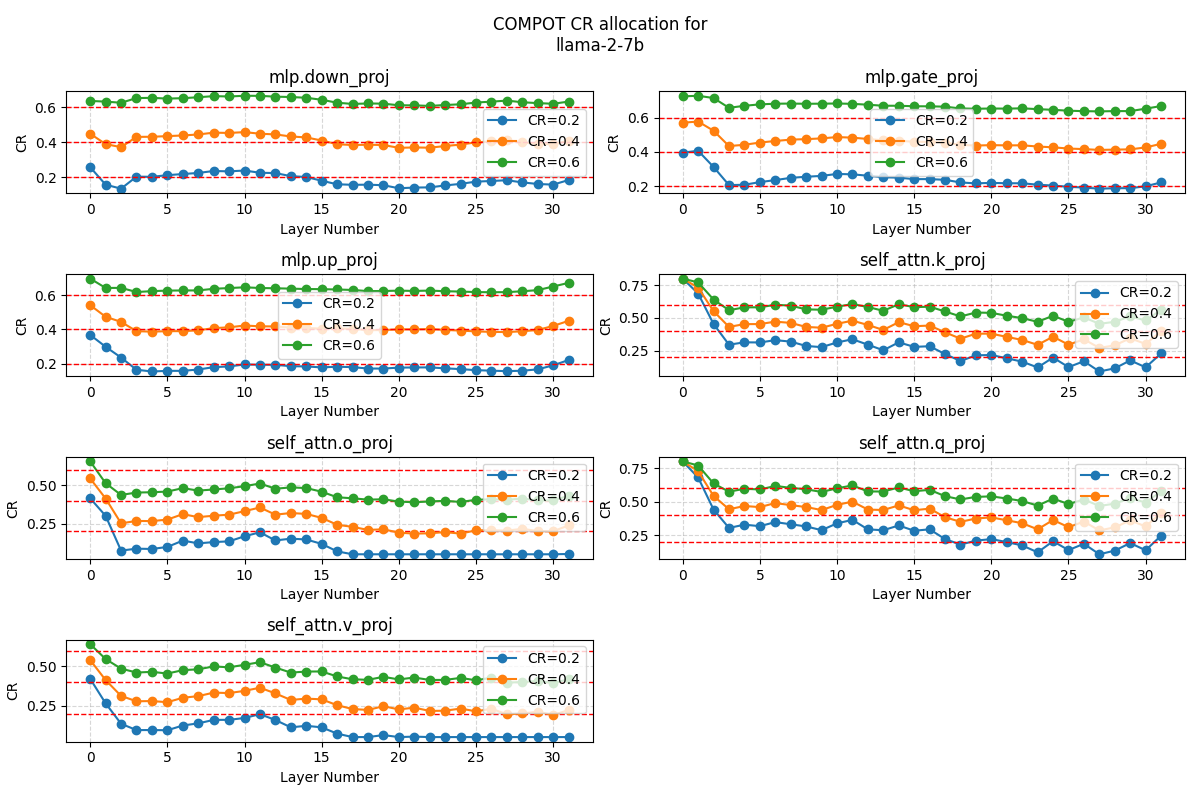}}
    \caption{Results of COMPOT allocation strategy for Llama2-7B}
    \label{fig:alloc_llama2_7b}
  \end{center}
\end{figure}

\begin{figure}[!ht]
  % \vskip 0.2in
  \begin{center}
    \centerline{\includegraphics[width=0.85\textwidth]{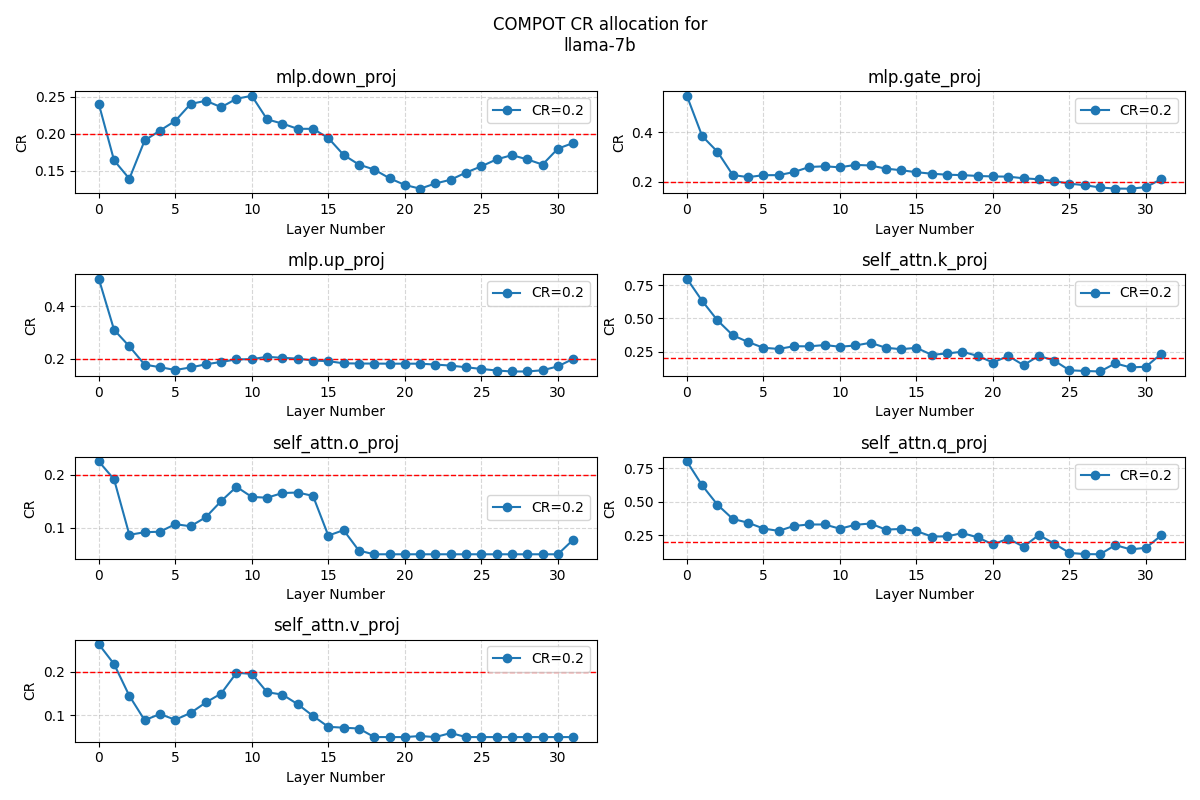}}
    \caption{Results of COMPOT allocation strategy for Llama-7B}
    \label{fig:alloc_llama_7b}
  \end{center}
\end{figure}

\begin{figure}[!ht]
  % \vskip 0.2in
  \begin{center}
    \centerline{\includegraphics[width=0.85\textwidth]{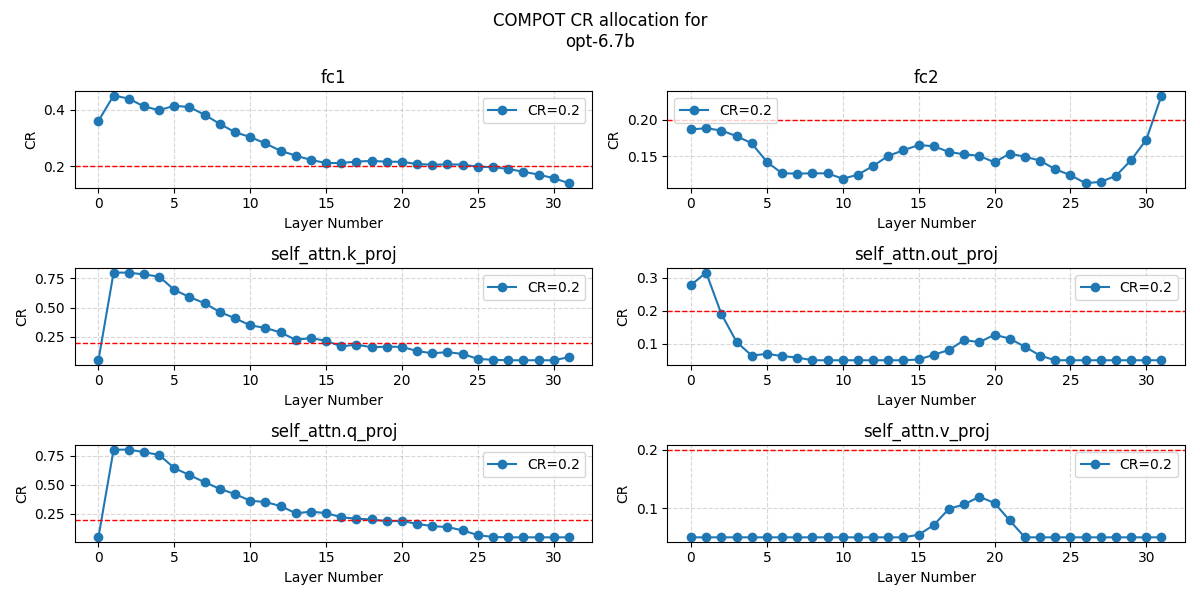}}
    \caption{Results of COMPOT allocation strategy for OPT-6.7B}
    \label{fig:alloc_opt_67b}
  \end{center}
\end{figure}

\begin{figure}[!ht]
  % \vskip 0.2in
  \begin{center}
    \centerline{\includegraphics[width=0.85\textwidth]{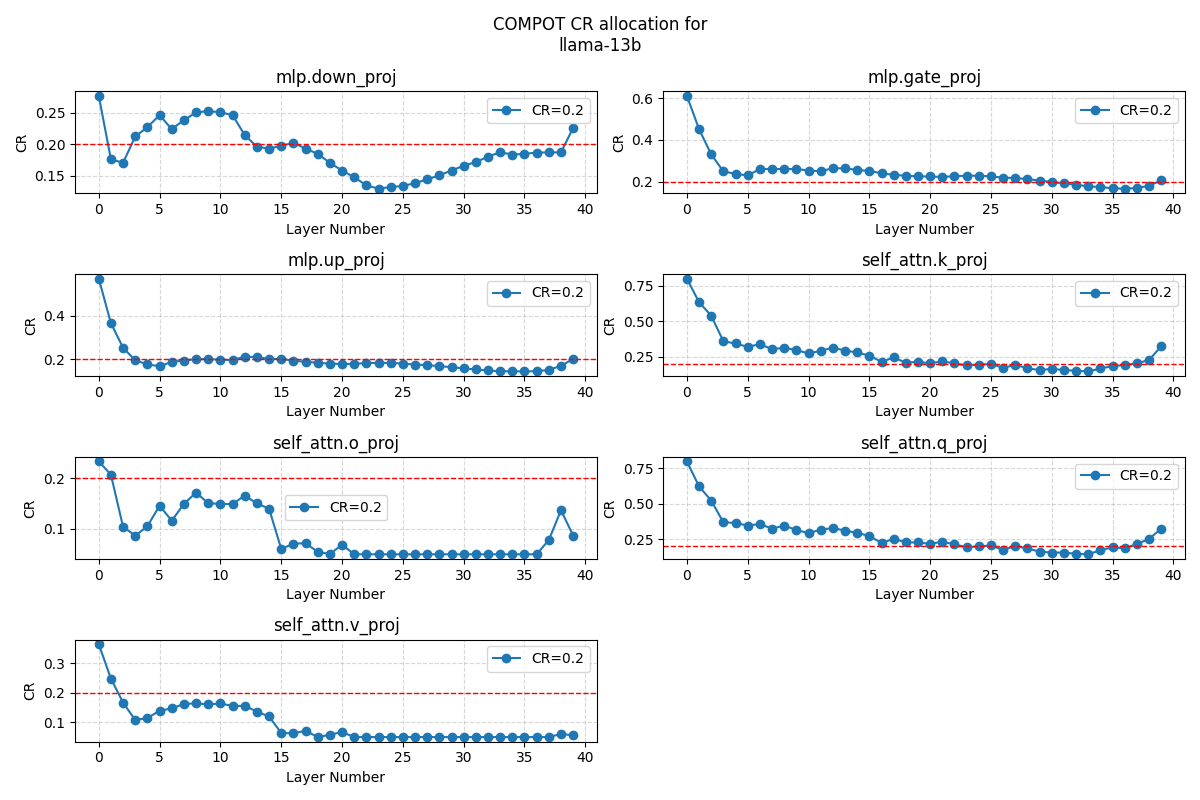}}
    \caption{Results of COMPOT allocation strategy for Llama-13B}
    \label{fig:alloc_llama_13b}
  \end{center}
\end{figure}

\begin{figure}[!ht]
  % \vskip 0.2in
  \begin{center}
    \centerline{\includegraphics[width=0.85\textwidth]{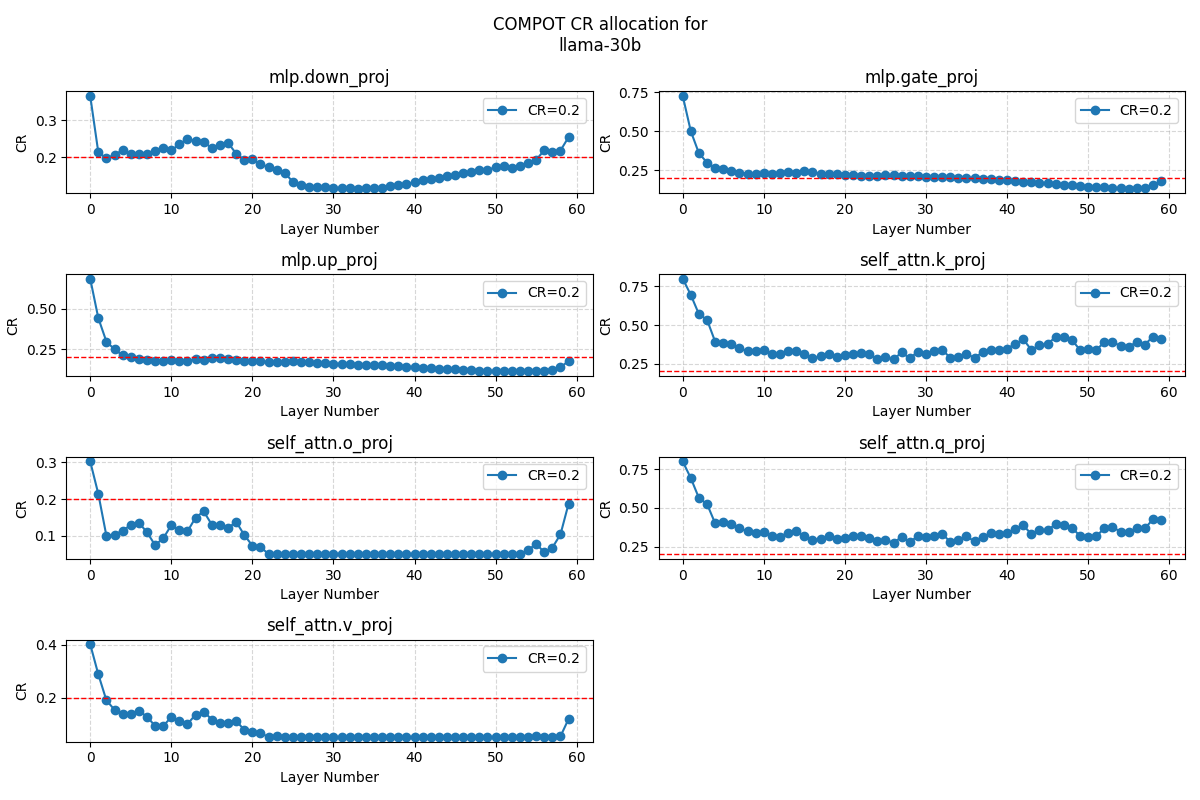}}
    \caption{Results of COMPOT allocation strategy for Llama-30B}
    \label{fig:alloc_llama_30b}
  \end{center}
\end{figure}

In this part we visualize the results of our allocation strategy across different models used throughout our extensive study (Figures~\ref{fig:alloc_llama32_1b} --~\ref{fig:alloc_llama_30b})



\subsection{Optimal Dictionary-to-sparsity ratio}
\label{apx:ks_ratio}
We evaluate the performance of the proposed COMPOT framework using the Llama 3.2-1B architecture as the primary backbone. The experimental setup involves optimizing the model under varying dictionary-to-sparsity ($k/s$) ratios, while maintaining a fixed computational budget of $N=100$ iterations. To ensure a robust assessment of the learning process, the dictionary is initialized randomly. Our empirical results demonstrate (see Table~\ref{tab:ablation_dictionary_sparsity_ratio}) that a $k/s$ ratio of 2 yields the optimal balance between representation capacity and sparse regularization, achieving the highest accuracy across the evaluated benchmarks.

\begin{table*}[!ht]
\caption{Performance of COMPOT across different dictionary-to-sparsity ($k/s$) ratios. The model compression rate is fixed at 20\%. All experiments are conducted using the Llama 3.2-1B architecture. Values in \textbf{bold} indicate the highest performance achieved for each respective metric.}
\label{tab:ablation_dictionary_sparsity_ratio}
\resizebox{\textwidth}{!}{%
\renewcommand{\arraystretch}{1.35}
\begin{tabular}{cccccccccc:c:cc}
\hline
   & \textbf{$k/s$} & \multicolumn{7}{c}{\textbf{Accuracy$\uparrow$}} &  & & \multicolumn{2}{c}{\textbf{Perplexity$\downarrow$}} \\
\multirow{-2}{*}{\textbf{Method}} & \textbf{ratio} & \textbf{PIQA} & \textbf{HellaSwag} & \textbf{LAMBADA} & \textbf{ARC-e} & \textbf{ARC-c} & \textbf{SciQ} & \textbf{Race}& \textbf{MMLU} & \multirow{-2}{*}{\textbf{Avg.}} & \textbf{WikiText} & \textbf{LAMBADA}    \\ \hline \hline

Llama 3.2-1B    &  --        & 74.5  &  63.7  & 62.95  & 60.5 & 36.2 & 88.3 & 37.8 & 37.0 & 57.6 & 11.57 &  5.73  \\ \hline
COMPOT\txtdag & 1.2          & 64.2	 &  40.3  & 30.22  & 39.6 & 25.8 & 72.5 & 31.1 & 23.1 & 40.9 & 73.73 & 83.31  \\
COMPOT\txtdag & 1.4          & 66.7	 &  45.7  & 36.64  & 43.7 & 27.9 & 77.9 & 31.4 & 24.9 & 44.3 & 35.38 & 41.48  \\
COMPOT\txtdag & 1.6          & 68.8	 &  49.1  & 40.69  & 47.3 & 29.2 & 79.9 & 33.4 & 25.8 & 46.8 & 25.29 & 29.75  \\
COMPOT\txtdag & 1.8          & 69.2	 &  51.4  & 43.06  & 48.5 & 29.2 & 81.4 & 34.5 & 25.9 & 47.9 & 19.95 & 26.18  \\ \hdashline
COMPOT\txtdag & 2.0          & 69.6	 &  52.3  & 41.53  & 48.9 & 30.1 & 80.9 & 34.1 & 27.4 & \textbf{48.1} & 20.52 & \textbf{24.55}  \\ \hdashline
COMPOT\txtdag & 2.4          & 69.1	 &  51.6  & 42.62  & 48.3 & 30.2 & 78.9 & 34.4 & 25.1 & 47.5 & \textbf{18.65} & 25.42  \\
COMPOT\txtdag & 2.8          & 68.6	 &  50.8  & 39.53  & 47.8 & 30.1 & 79.6 & 33.8 & 25.6 & 47.0 & 23.77 & 26.76  \\
COMPOT\txtdag & 3.2          & 67.5	 &  49.6  & 35.71  & 47.1 & 27.7 & 77.1 & 33.9 & 24.8 & 45.4 & 30.55 & 28.78  \\
COMPOT\txtdag & 3.6          & 66.2	 &  47.3  & 37.47  & 45.6 & 26.8 & 78.6 & 33.1 & 25.1 & 45.0 & 31.01 & 31.86  \\
COMPOT\txtdag & 4.0          & 65.9	 &  45.5  & 29.90  & 40.4 & 27.4 & 71.5 & 32.0 & 23.9 & 42.1 & 50.58 & 37.71  \\ \hline
\end{tabular}
}
\end{table*}
\subsection{Additional results for vision and audio models}
\label{apx:omni}
This section provides complete evaluation results for vision and audio models across various compression ratios (CR). For each benchmark, we report the corresponding CR and include the uncompressed model alongside SVD-LLM~\cite{svdllm} as reference points. Unless otherwise noted, we adhere to the same evaluation protocol used in the main paper (employing identical prompts, decoding strategies, preprocessing, and metric computation) to ensure direct comparability across all methods.

\paragraph{Vision.} For the Qwen 3 VL 8B-Instruct model, we compressed only the language module using 256 samples from the MathVerse training data for calibration. Comprehensive results on four vision-language benchmarks, namely, MMMU, OCRBench, RealWorldQA, and MMStar, are presented in Table~\ref{tab:qwen3vl_full}. Across all compression ratios, our COMPOT method consistently preserves substantially more performance than the SVD-LLM baseline.

\paragraph{Audio.} For all Whisper models in this experiment, we compressed the decoder using 1024 samples from the LibriSpeech clean validation dataset. Table~\ref{tab:full_whisper_results} reports automatic speech recognition (ASR) results, measured in word error rate (WER, lower is better), for Whisper Base-EN, Medium-EN V3, and Large V3 on the LibriSpeech test-clean and test-other sets. We evaluate multiple compression ratios where applicable. Overall, while SVD-LLM exhibits rapid performance degradation with stronger compression (particularly for smaller models), COMPOT remains close to the uncompressed baseline across a wide range of CRs. In several settings, COMPOT matches or even slightly improves upon the original model's performance.

\begin{table}[ht]
    \centering
    \caption{Full vision-language results for Qwen3-VL 8B under compression. We report benchmark scores on MMMU, OCRBench, RealWorldQA, and MMStar, along with the uncompressed model.}
    \label{tab:qwen3vl_full}
    \resizebox{0.9 \textwidth}{!}{%
    \renewcommand{\arraystretch}{1.05}
    \begin{tabular}{lccccc:c}
        \hline
        Method & CR & MMMU val $\uparrow$ & OCRBench $\uparrow$ & RealWorldQA $\uparrow$ & MMStar $\uparrow$ & Average $\uparrow$ \\
        \hline \hline
        Qwen 3 VL 8B & -- & 0.5256 & 0.8259 & 0.6967 & 0.6315 & 0.67 \\ \hline
        SVD-LLM & 0.20 & 0.2922 & 0.3510 & 0.5294 & 0.3161 & 0.37 \\
        COMPOT\txtdag  & 0.20 & 0.4189 & 0.6410 & 0.5922 & 0.5080 & 0.54 \\
        COMPOT  & 0.20 & 0.4467 & 0.6690 & 0.6222 & 0.5417 & 0.57 \\ \hline
        SVD-LLM & 0.30 & 0.2333 & 0.2120 & 0.2562 & 0.0398 & 0.19 \\
        COMPOT\txtdag  & 0.30 & 0.3378 & 0.5260 & 0.5137 & 0.3574 & 0.43 \\
        COMPOT  & 0.30 & 0.3578 & 0.5760 & 0.5098 & 0.4050 & 0.46 \\ \hline
        SVD-LLM & 0.40 & 0.2467 & 0.0390 & 0.0000 & 0.0030 & 0.07 \\
        COMPOT\txtdag  & 0.40 & 0.2367 & 0.3300 & 0.3359 & 0.0576 & 0.24 \\
        COMPOT  & 0.40 & 0.2511 & 0.2630 & 0.4588 & 0.1638 & 0.28 \\
        \hline
    \end{tabular}
    }
\end{table}

\begin{table}[]
\centering
\caption{ASR Performance (WER $\downarrow$) on LibriSpeech benchmarks for Whisper family.}
\label{tab:full_whisper_results}
\resizebox{0.5 \textwidth}{!}{%
\renewcommand{\arraystretch}{1.05}
\begin{tabular}{@{}lccc@{}}
\hline
\textbf{Method} & \textbf{CR} & $\textbf{WER}_{\textbf{test-clean}}$ $\downarrow$ & $\textbf{WER}_{\textbf{test-other}}$ $\downarrow$ \\ \hline \hline
\textbf{Whisper Base-EN} & -    & 4.40 & 10.40 \\ \hline
SVD-LLM                    & 0.05& 32.37 & 32.18\\
\textbf{COMPOT\txtdag}     & 0.05& \textbf{5.41} &	\textbf{12.34}\\ \hdashline
SVD-LLM                    & 0.10& 52.80 & 51.02\\
\textbf{COMPOT\txtdag}     & 0.10& \textbf{5.88} & \textbf{12.75}\\ \hdashline
SVD-LLM                    & 0.15& 85.64&83.31\\
\textbf{COMPOT\txtdag}     & 0.15& \textbf{8.57} & \textbf{15.30}\\ \hline \hline
\textbf{Whisper Medium-EN} & -    & 2.93 & 5.93\\ \hline
SVD-LLM                    & 0.05& 3.84 & 7.26\\
\textbf{COMPOT\txtdag}     & 0.05& \textbf{2.92} & \textbf{6.06}\\ \hdashline
SVD-LLM                    & 0.10& 5.11 & 8.39\\
\textbf{COMPOT\txtdag}     & 0.10& \textbf{3.07} & \textbf{6.18}\\ \hdashline
SVD-LLM                    & 0.15& 6.95 & 10.14\\
\textbf{COMPOT\txtdag}     & 0.15& \textbf{3.12} & \textbf{6.34}\\ \hdashline
SVD-LLM                    & 0.20& 12.16 & 15.23\\
\textbf{COMPOT\txtdag}     & 0.20& \textbf{3.26} & \textbf{6.53}\\ \hdashline
SVD-LLM                    & 0.30& 32.20 & 33.09\\
\textbf{COMPOT\txtdag}     & 0.30& \textbf{4.08} & \textbf{7.78}\\ \hline \hline

\textbf{Whisper Large V3}  & -    & 2.74& 4.53\\ \hline
SVD-LLM                    & 0.2& 4.12& 6.8\\
\textbf{COMPOT\txtdag}     & 0.2& \textbf{2.46}& \textbf{4.51}\\ \hdashline
SVD-LLM                    & 0.3& 12.78& 15.54\\
\textbf{COMPOT\txtdag}     & 0.3& \textbf{2.74}& \textbf{5.21}\\ \hline
\end{tabular}
}
\end{table}
\subsection{Reproducing SVD-LLM V2}
\label{apx:svd_llm_v2_details}

\paragraph{Codebase and evaluation protocol.}
We based our reproduction on the official SVD-LLM repository\footnote{\url{https://github.com/AIoT-MLSys-Lab/SVD-LLM}}, which provides scripts for compression and evaluation. To ensure comparability, we adopted the same evaluation protocol used by the repository and aligned our calibration setup accordingly: 256 WikiText samples with context length 2048, and perplexity evaluated using the provided scripts on WikiText and C4.

\paragraph{Implementing SVD-LLM V2 components.}
SVD-LLM V2 proposes two modifications over SVD-LLM: (i) assigning
non-uniform compression ratios using theoretical truncation loss and
(ii) replacing Cholesky-based truncation with a loss-optimized two-SVD
procedure~\citep{svdllmv2}. Since the public SVD-LLM repository does
not provide a dedicated, ready-to-run SVD-LLM V2 script, we implemented
these two components on top of the released SVD-LLM codebase. The
implementation below follows the V2 paper as closely as possible while
making explicit several conventions that are implicit or ambiguous in
the paper pseudocode.

\paragraph{Compression-ratio convention.}
Throughout this paper, we report the compression ratio \(c\) as the
fraction of parameters removed. Thus \(c=0.2\) means that approximately
\(80\%\) of the original parameters are retained. By contrast, the rank
formula for an SVD factorization is naturally expressed using the
retained-parameter ratio
\[
\rho = 1-c .
\]
For a matrix \(\m W\in\mathbb{R}^{m\times n}\), the rank used by the
compressed two-layer low-rank factorization is
\[
r=\left\lfloor \frac{\rho mn}{m+n}\right\rfloor .
\]
Therefore, our implementation uses \(c\) only for reporting and
allocation of removed parameters, while all SVD rank computations use
\(\rho\). This distinction is important: treating a compression ratio as
a retained ratio reverses the intended allocation and changes the
parameter budget.
\vspace{1cm}
\lstset{style=Python}
\begin{lstlisting}[language=Python, caption={Explicit compression-ratio and retained-ratio utilities}]
def retained_ratio_from_compression(compression_ratio: float) -> float:
    """
    Paper/tables report c = fraction of parameters removed.
    SVD rank computation uses rho = fraction of parameters retained.
    """
    if not (0.0 <= compression_ratio < 1.0):
        raise ValueError("compression_ratio must be in [0, 1).")
    return 1.0 - compression_ratio


def low_rank_svd_rank(W: torch.Tensor, retained_ratio: float) -> int:
    """
    Rank r for replacing W in R^{m x n} by two matrices
    U in R^{m x r} and V in R^{r x n}. The retained parameter
    count is approximately rho * m * n.
    """
    m, n = W.shape
    r = int(m * n * retained_ratio / (m + n))
    return max(1, min(r, min(m, n)))
\end{lstlisting}

\paragraph{V2 theoretical truncation loss.}
The original SVD-LLM implementation computes a whitening factor using
Cholesky decomposition. In contrast, SVD-LLM V2 explicitly replaces the
Cholesky-based truncation with a two-SVD construction. Given the input
covariance/statistics matrix \(\m S = \m X^\top \m X\), we compute
\[
\m S = \m U_s \operatorname{diag}(\sigma_s) \m V_s^\top,
\qquad
\m D = \m W \m U_s \operatorname{diag}(\sqrt{\sigma_s}).
\]
The theoretical loss at rank \(r\) is then the Frobenius norm of the
discarded singular spectrum of \(\m D\). We use this quantity in the
allocation step.

\begin{lstlisting}[language=Python, caption={SVD-LLM V2 theoretical truncation loss without Cholesky}]
def theoretical_loss_v2(
    W: torch.Tensor,
    S: torch.Tensor,
    retained_ratio: float,
    dev: str = "cuda",
    eps: float = 1e-12,
) -> float:
    """
    Compute the SVD-LLM V2 theoretical truncation loss.

    W: weight matrix with shape [out_features, in_features].
    S: raw input covariance/statistics matrix X^T X with shape
       [in_features, in_features].
    retained_ratio: rho, not the reported compression ratio c.

    The loss is computed from the singular values of
        D = W U_s sqrt(S_s),
    following the loss-optimized truncation view in SVD-LLM V2.
    """
    W = W.to(device=dev, dtype=torch.float64)
    S = S.to(device=dev, dtype=torch.float64)

    # Symmetrize for numerical stability, since S is accumulated
    # from finite calibration data.
    S = 0.5 * (S + S.T)

    U_s, s_s, _ = torch.linalg.svd(S, full_matrices=False)
    s_s = torch.clamp(s_s, min=eps)

    D = W @ U_s @ torch.diag(torch.sqrt(s_s))
    singular_values = torch.linalg.svdvals(D)

    r = low_rank_svd_rank(W, retained_ratio)
    discarded = singular_values[r:]

    if discarded.numel() == 0:
        return 0.0
    return torch.linalg.norm(discarded, ord=2).item()
\end{lstlisting}

\paragraph{Best-effort dynamic allocation.}
For heterogeneous allocation, we group matrices by projection type, as
described in SVD-LLM V2. For each group \(G\), we first compute the
theoretical truncation loss \(L_i\) at the target retained ratio
\(\rho=1-c\). The paper then applies an inverse-log transformation to
obtain allocation weights. Since the pseudocode does not specify how to
handle numerically invalid cases such as \(\log(L_i)\le 0\), we use a
guarded implementation and normalize the resulting values within each
projection-type group. The allocated per-matrix compression ratio is
\[
c_i = |G|\,c\,\frac{\omega_i}{\sum_{j\in G}\omega_j},
\]
and the retained ratio used by the SVD rank formula is
\[
\rho_i = 1-c_i .
\]
The dictionary returned by the function below therefore stores retained
ratios, because these are the values consumed by the low-rank SVD
modules.

\begin{lstlisting}[language=Python, caption={Best-effort dynamic compression-ratio allocation with explicit ratio conversion}]
def cr_allocation_v2(
    model_name: str,
    model: nn.Module,
    profiling_mat,
    target_compression_ratio: float,
    dev: str = "cuda",
    eps: float = 1e-12,
    min_keep_ratio: float = 0.01,
    max_keep_ratio: float = 0.99,
):
    """
    Allocate per-matrix retained ratios following the SVD-LLM V2
    heterogeneous allocation idea.

    target_compression_ratio is c, the fraction of parameters removed.
    The returned dictionary contains rho_i = 1 - c_i, i.e. retained
    ratios used in rank computation.
    """
    print("Starting SVD-LLM V2 dynamic allocation...")
    model.eval()

    target_retained_ratio = retained_ratio_from_compression(
        target_compression_ratio
    )

    if "opt" in model_name:
        layers = model.model.decoder.layers
        layer_prefix = "model.decoder.layers"
        projection_types = [
            "fc1",
            "fc2",
            "self_attn.q_proj",
            "self_attn.k_proj",
            "self_attn.v_proj",
            "self_attn.out_proj",
        ]
    else:
        layers = model.model.layers
        layer_prefix = "model.layers"
        projection_types = [
            "mlp.up_proj",
            "mlp.down_proj",
            "mlp.gate_proj",
            "self_attn.q_proj",
            "self_attn.k_proj",
            "self_attn.v_proj",
            "self_attn.o_proj",
        ]

    retained_ratio_dict = {}

    for projection_type in projection_types:
        losses = []
        names = []

        for layer_idx in tqdm(range(len(layers)), total=len(layers)):
            full_name = f"{layer_prefix}.{layer_idx}.{projection_type}"

            # HuggingFace Linear weights are [out_features, in_features].
            W = model.get_submodule(full_name).weight.data.float()

            # Use raw covariance/statistics S = X^T X, not the Cholesky factor.
            S = profiling_mat[layer_idx]["raw_scaling_diag_matrix"][
                projection_type
            ]

            loss = theoretical_loss_v2(
                W=W,
                S=S,
                retained_ratio=target_retained_ratio,
                dev=dev,
                eps=eps,
            )
            losses.append(loss)
            names.append(full_name)

        # The paper specifies an inverse-log transformation. We guard it
        # to avoid sign changes or division by zero when losses are close
        # to one or smaller than one.
        weights = []
        for loss in losses:
            safe_loss = max(float(loss), 1.0 + eps)
            weights.append(1.0 / max(math.log(safe_loss), eps))

        weight_sum = sum(weights)

        for full_name, weight in zip(names, weights):
            # Allocate removed-parameter fractions c_i.
            c_i = (
                len(weights)
                * target_compression_ratio
                * weight
                / weight_sum
            )

            # Convert to retained ratio rho_i for rank computation.
            rho_i = 1.0 - c_i
            rho_i = max(min_keep_ratio, min(max_keep_ratio, rho_i))
            retained_ratio_dict[full_name] = rho_i

    return retained_ratio_dict
\end{lstlisting}

\paragraph{Loss-optimized truncation.}
After obtaining the per-matrix retained ratios, we compress each weight
matrix using the two-SVD truncation from SVD-LLM V2. The function below
returns the two factors stored in the compressed model. It also clamps
small singular values of the calibration covariance before applying the
inverse square root, which avoids numerical failures for rank-deficient
or nearly rank-deficient calibration statistics.

\begin{lstlisting}[language=Python, caption={Loss-optimized SVD-LLM V2 truncation used for each linear layer}]
def loss_optimized_svd_factors_v2(
    W: torch.Tensor,
    S: torch.Tensor,
    retained_ratio: float,
    dev: str = "cuda",
    eps: float = 1e-12,
):
    """
    Compute low-rank factors for SVD-LLM V2 loss-optimized truncation.

    W: dense weight with shape [out_features, in_features].
    S: raw covariance/statistics matrix X^T X.
    retained_ratio: rho_i used for this matrix.

    Returns:
        svd_u: [out_features, r]
        svd_v: [r, in_features]
    so that W is approximated by svd_u @ svd_v.
    """
    dtype = W.dtype
    W = W.to(device=dev, dtype=torch.float64)
    S = S.to(device=dev, dtype=torch.float64)

    S = 0.5 * (S + S.T)

    U_s, s_s, _ = torch.linalg.svd(S, full_matrices=False)

    # Stabilize sqrt and inverse sqrt.
    scale = torch.clamp(s_s.max(), min=1.0)
    s_floor = eps * scale
    s_s = torch.clamp(s_s, min=s_floor)

    sqrt_s = torch.sqrt(s_s)
    inv_sqrt_s = torch.rsqrt(s_s)

    D = W @ U_s @ torch.diag(sqrt_s)
    U_d, s_d, Vh_d = torch.linalg.svd(D, full_matrices=False)

    r = low_rank_svd_rank(W, retained_ratio)

    svd_u = U_d[:, :r] @ torch.diag(s_d[:r])
    svd_v = Vh_d[:r, :] @ torch.diag(inv_sqrt_s) @ U_s.T

    return svd_u.to(dtype).cpu(), svd_v.to(dtype).cpu()
\end{lstlisting}

\paragraph{Integration into the compression pass.}
The function above replaces the Cholesky-based whitening truncation used
in the original SVD-LLM code path. In our implementation, the profiling
pass still collects the same raw activation statistics, but the actual
V2 truncation consumes \(\m S=\m X^\top \m X\) directly.

\begin{lstlisting}[language=Python, caption={Replacing each linear layer using V2 retained ratios}]
@torch.no_grad()
def compress_linear_layers_v2(
    model_name: str,
    model: nn.Module,
    profiling_mat,
    retained_ratio_dict,
    save_path: str,
    dev: str = "cuda",
):
    """
    Skeleton of the V2 replacement loop. The surrounding module
    replacement code follows the released SVD-LLM implementation,
    but each linear layer is decomposed using the V2 two-SVD routine.
    """
    model.eval()

    if "opt" in model_name:
        layers = model.model.decoder.layers
        layer_prefix = "model.decoder.layers"
    else:
        layers = model.model.layers
        layer_prefix = "model.layers"

    for layer_idx in tqdm(range(len(layers))):
        layer = layers[layer_idx]
        subset = find_layers(layer)

        for name, linear in subset.items():
            full_name = f"{layer_prefix}.{layer_idx}.{name}"

            W = linear.weight.data.float()
            S = profiling_mat[layer_idx]["raw_scaling_diag_matrix"][name]

            retained_ratio = retained_ratio_dict[full_name]

            svd_u, svd_v = loss_optimized_svd_factors_v2(
                W=W,
                S=S,
                retained_ratio=retained_ratio,
                dev=dev,
            )

            # The assignment to q_u_proj/q_v_proj, k_u_proj/k_v_proj,
            # etc. follows the same module-specific logic as the
            # released SVD-LLM codebase. For example:
            #
            #   svd_attn.q_u_proj.weight.data = svd_u
            #   svd_attn.q_v_proj.weight.data = svd_v
            #
            # and analogously for K, V, O, Gate, Up, and Down
            # projections.
\end{lstlisting}

\paragraph{Ambiguities and fine-tuning.}
The SVD-LLM repository contains an optional sequential local update
stage inherited from the original SVD-LLM pipeline. We did not use this
stage in our SVD-LLM V2 reproduction. The V2 paper describes SVD-LLM V2
as a post-training compression method and attributes the improvement to
dynamic compression-ratio allocation and loss-optimized truncation, but
it does not clearly state whether any additional fine-tuning or
post-compression local update was applied in the reported V2 numbers.
Because the paper also reports SVD-LLM results as a baseline without
fine-tuning, and because the V2 method description emphasizes
post-training truncation rather than optimization with gradient updates,
we conclude that no additional fine-tuning was intended for the main
SVD-LLM V2 results. Accordingly, our reproduction uses calibration data
only for collecting activation statistics and does not apply
post-compression fine-tuning or the sequential local update step.

\paragraph{Implementation caveats.}
The published SVD-LLM V2 pseudocode leaves several numerical details
underspecified: the exact convention for compression versus retained
ratios, how to handle inverse-log allocation when \(\log(L_i)\) is
ill-conditioned or changes sign, and how to stabilize the inverse square
root of the calibration covariance spectrum. We therefore treat our
implementation as a best-effort reproduction rather than an official
implementation. To avoid relying solely on these reproduction details,
the main comparison reports paper-reported SVD-LLM V2 numbers separately
from our reproduced values.

\paragraph{Model coverage and reproduction gaps.}
We attempted to reproduce the reported SVD-LLM V2 results for Llama-7B, OPT-6.7B, and Llama3-8B. For Llama-7B, our reproduction follows the released SVD-LLM code convention and gives results in the same broad range as the reported values, providing a useful sanity check for the implementation. For OPT-6.7B and Llama3-8B, we encountered model- and environment-specific difficulties in the public repository; community reports also describe failures or substantial degradations for related settings (e.g., GitHub Issues~\#49 and \#34)\footnote{\url{https://github.com/AIoT-MLSys-Lab/SVD-LLM/issues/49}}\footnote{\url{https://github.com/AIoT-MLSys-Lab/SVD-LLM/issues/34}}. For Llama3-8B, the run completed but produced substantially worse results than those reported in the SVD-LLM V2 paper. We therefore report our reproduction separately from the paper-reported SVD-LLM V2 numbers and use the latter as the primary reference when discussing V2 performance.

% \paragraph{Possible sources of discrepancy.}
% The SVD-LLM repository includes an optional post-compression parameter update step (fine-tuning) and related discussions in the issue tracker (e.g., Issue~\#20)\footnote{\url{https://github.com/AIoT-MLSys-Lab/SVD-LLM/issues/20}}. If additional fine-tuning or other unpublished post-processing was used in the SVD-LLM V2 experiments, it could partially explain gaps between paper-claimed numbers and strictly post-training reproductions. We therefore (i) report our reproduction strictly without additional fine-tuning, and (ii) separate ``paper-claimed'' vs.\ ``reproduced'' results in all SVD-LLM V2 comparisons.

\paragraph{Possible sources of discrepancy.}
Several factors may explain the gap between reproduced and paper-reported SVD-LLM V2 results: unresolved ratio-notation ambiguity in the V2 pseudocode, numerical sensitivity of whitening/statistics inversion, model-version differences, and optional post-compression updates in the SVD-LLM repository (e.g., Issue~\#20)\footnote{\url{https://github.com/AIoT-MLSys-Lab/SVD-LLM/issues/20}}. We do not use additional fine-tuning, sequential local update, or other post-compression optimization in our SVD-LLM V2 reproduction. Accordingly, all tables distinguish paper-reported SVD-LLM V2 numbers from our reproduced values. Importantly, even when compared against the paper-reported SVD-LLM V2 values rather than our reproduction, COMPOT achieves lower WikiText/C4 perplexity on two of the three models in Table~\ref{tab:main_dynamic_svdllmv2}.

\paragraph{Results on larger scale.}
Table~\ref{tab:llama13b_30b_results} reports results for Llama-13B and Llama-30B using the SVD-LLM evaluation protocol to enable a like-for-like comparison. COMPOT (dynamic allocation) matches or improves upon prior SVD-based and activation-aware baselines, and is comparable to the SVD-LLM V2 numbers reported in the literature. However, we treat SVD-LLM V2 as a \emph{reference} point: a dedicated, ready-to-run V2 implementation is not publicly available, and our attempts to reproduce V2-style results at smaller scales showed sizable gaps relative to the paper-reported values. We therefore report only the paper-claimed numbers in the appendix, and interpret the V2 comparison in Table~\ref{tab:llama13b_30b_results} accordingly.

\begin{table}[t]
\centering
\caption{Perplexity (PPL, $\downarrow$) and average accuracy (Avg.\ Acc., $\uparrow$) for Llama-13B and Llama-30B.}
\label{tab:llama13b_30b_results}
\renewcommand{\arraystretch}{1.05}
\begin{tabular}{c|cc|cc}
\hline
 & \multicolumn{2}{c|}{\textbf{Llama-13B}} & \multicolumn{2}{c}{\textbf{Llama-30B}} \\
\multirow{-2}{*}{\textbf{Method}} & \textbf{PPL} & \textbf{Avg.\ Acc.} & \textbf{PPL} & \textbf{Avg.\ Acc.} \\
\hline
Original & 5.09 & 0.59 & 4.10 & 0.61 \\ \hline\hline
FWSVD & 15.98 & 0.43 & 20.54 & 0.42 \\
ASVD & 6.74 & 0.54 & 22.71 & 0.44 \\
SVD-LLM & 6.61 & 0.55 & 5.63 & 0.57 \\ \hdashline
SVD-LLM v2 & 5.46 & 0.56 & 4.71 & 0.60 \\
COMPOT & 6.22 & 0.57 & 5.00 & 0.60 \\
\hline
\end{tabular}
\end{table}
\subsection{Detailed Description on Comparison with Dobi-SVD}
\label{apx:dobi_details}

A direct comparison between COMPOT and Dobi-SVD requires careful methodological alignment, as the two approaches differ fundamentally in their compression paradigms. We identify two critical distinctions that limit comparability:

\textbf{(1) Training-free vs. training-based compression.} COMPOT operates strictly in the post-training regime without any gradient-based optimization. In contrast, Dobi-SVD employs backpropagation to optimize layer-wise ranks through a differentiable truncation objective~\citep{dobisvd}. This training step—absent in COMPOT—introduces a confounding factor that advantages Dobi-SVD through implicit fine-tuning of the compressed model, violating the core premise of training-free compression.

\textbf{(2) Remapping-induced overparameterization.} Dobi-SVD introduces a ``remapping'' operation that reparameterizes factorized layers to match the original dense matrix dimensions before quantization. Crucially, this remapping often \emph{increases} the parameter count of the factorized representation. When matrix factorization (yielding factorization compression ratio $CR_{fact}$) is followed by quantization to $b$ bits (original weights stored in 16-bit FP16), the effective model-wide compression ratio is:
\begin{equation}
    CR_{target} = 1 - (1 - CR_{fact}) \cdot \frac{b}{16}.
    \label{eq:combined_cr}
\end{equation}
For Dobi-SVD's 8-bit quantization ($b=8$), Equation~\eqref{eq:combined_cr} simplifies to $CR_{target} = \frac{1}{2}(1 + CR_{fact})$. Critically, when remapping yields $CR_{fact} < 0$ (i.e., the factorized representation consumes \emph{more} parameters than the original dense matrix), the net compression arises \emph{entirely} from quantization rather than structural factorization. In such cases, Dobi-SVD effectively trades parameter inflation for quantization headroom—a strategy orthogonal to COMPOT's goal of reducing stored parameters through structural compression.

To ensure a methodologically sound comparison, we evaluate Dobi-SVD \emph{without} remapping (denoted Dobi-SVD\txtdag), which preserves its factorization-based compression objective. Table~\ref{tab:dobisvd_full} reports results on Llama2-7B under identical evaluation protocols (WikiText/C4 perplexity via SVD-LLM scripts; zero-shot accuracy via \texttt{lm-eval} 0.4.8). COMPOT consistently outperforms Dobi-SVD\txtdag across compression ratios while remaining strictly training-free.

When remapping is enabled, Dobi-SVD's gains derive predominantly from quantization rather than factorization efficiency. For instance, at target $CR_{target}=0.2$, Dobi-SVD applies \emph{negative} factorization compression ($CR_{fact}=-0.6$) followed by 8-bit quantization to achieve net compression—effectively overparameterizing the model by 60\% to exploit quantization benefits. This paradigm shift renders comparisons against pure factorization methods like COMPOT fundamentally misaligned. We include these results in Table~\ref{tab:dobisvd_full} for completeness, but emphasize that they reflect quantization efficacy rather than factorization quality.

\begin{table}[htbp]
\centering
\caption{Comparison of COMPOT with Dobi-SVD on Llama2-7B without (denoted with \txtdag) and with remapping. We provide also CR from matrix factorization (Fact. CR) and quantization (Quant. CR) to achieve specified target CR. In most cases Dobi-SVD with remapping leads to overparametrization with matrix factorization rather than compression}
\label{tab:dobisvd_full}
\resizebox{\textwidth}{!}{
\renewcommand{\arraystretch}{1.05}
\begin{tabular}{l *{13}{c}}
\hline
\textbf{Method} & \textbf{Target CR} & \textbf{Fact. CR} & \textbf{Quant. CR} & \textbf{WikiText} & \textbf{C4} & \textbf{Openb.} & \textbf{Arc\_e} & \textbf{WinoG.} & \textbf{HellaS.} & \textbf{Arc\_c} & \textbf{PIQA} & \textbf{MathQA} & \textbf{Average} \\ \hline

Llama2 7B   & --- & --- & --- & 5.49 & 7.27 & 44.0 & 74.6 & 69.0 & 76.3 & 45.5 & 78.6 & 28.4 & 59.5 \\ \hline

Dobi-SVD\txtdag   & \multirow{3}{*}{0.2}  & 0.2 & --- & 9.39 & 19.70 & 32.8 & 47.6 & 58.5 & 51.5 & 29.3 & 65.2 & 22.7 & 43.9 \\
Dobi-SVD    &                       & -0.6& 0.5 & 5.92 & 8.68  & 42.2 & 72.3 & 69.1 & 72.9 & 44.0 & 78.1 & 27.5 & 58.0 \\
COMPOT      &                       & 0.2 & --- & 6.22 & 9.34  & 30.4 & 72.6 & 67.6 & 51.8 & 39.6 & 76.4 & 26.2 & 52.1 \\ \hline

Dobi-SVD\txtdag   & \multirow{3}{*}{0.4}  & 0.4 & --- & 17.50 & 55.58 & 29.2 & 34.4 & 54.4 & 35.4 & 24.5 & 57.3 & 21.3 & 36.6 \\
Dobi-SVD    &                       & -0.2& 0.5 & 7.89 & 14.23  & 37.2 & 59.4 & 62.7 & 60.6 & 35.0 & 72.1 & 25.5 & 50.4 \\
COMPOT      &                       & 0.4 & --- & 8.91 & 20.40  & 25.2 & 63.1 & 63.5 & 39.6 & 28.5 & 67.0 & 23.6 & 44.4 \\ \hline

Dobi-SVD\txtdag   & \multirow{3}{*}{0.6}  & 0.6 & --- & 53.37 & 200.11 & 26.0 & 28.3 & 50.5 & 28.4 & 24.7 & 51.9 & 22.0 & 33.1 \\
Dobi-SVD    &                       & 0.2 & 0.5 & 9.66 & 20.28   & 35.6 & 44.7 & 57.9 & 49.6 & 27.9 & 64.3 & 23.6 & 43.4 \\
COMPOT      &                       & 0.6 & --- & 32.10 & 179.02 & 13.4 & 33.8 & 53.4 & 28.8 & 18.8 & 54.9 & 22.1 & 32.2 \\ \hline
\end{tabular}
}
\end{table}
\subsection{Calibration Robustness and Reproducibility}
\label{apx:seed_ablation}

\paragraph{Number of calibration samples.}
We evaluate sensitivity to the calibration-set size on Llama3.2-1B at CR $0.2$, using FineWeb while keeping the sequence length, dynamic allocation, SVD initialization, and $20$ COMPOT iterations fixed. Results are shown in Table~\ref{tab:calib_size}.

\begin{table}[h]
\centering
\caption{Effect of the number of calibration samples on COMPOT for Llama3.2-1B at CR $0.2$.}
\label{tab:calib_size}
\resizebox{0.5\textwidth}{!}{%
\renewcommand{\arraystretch}{1.05}
\begin{tabular}{cccc}
\hline
\textbf{Samples} &
\textbf{Avg. Acc.$\uparrow$} &
\textbf{WikiText PPL$\downarrow$} &
\textbf{LAMBADA PPL$\downarrow$} \\ \hline \hline
64  & 49.86 & 21.52 & 14.52 \\
128 & 48.83 & 20.76 & 14.01 \\
256 & 50.06 & 20.71 & 13.97 \\
512 & \textbf{50.65} & \textbf{20.49} & \textbf{12.37} \\
\hline
\end{tabular}
}
\end{table}

Performance remains stable across relatively small calibration sets: from 64 to 256 samples,
average accuracy varies within $1.23$ percentage points, while perplexity changes moderately.
Increasing the calibration set from 256 to 512 samples yields only a further $0.59$-point
accuracy improvement, supporting 256 samples as a practical quality--calibration-cost trade-off.

\paragraph{Sampling reproducibility.}
To assess reproducibility, we evaluate our compression method across 5 random seeds for data sampling on Llama3.2-1B with 20\% compression. Results (Table~\ref{tab:seed_ablation}) show low variance for most metrics (e.g., Avg. Acc.: 50.2 ± 0.3), indicating robustness to sampling variation. Lambada perplexity exhibits higher sensitivity (13.51 ± 0.63), likely due to its dependence on long-range context; however, all runs remain within a competitive range relative to baselines.

\begin{table}[h]
\centering
\caption{Results across different data sampling seeds. We report mean $\pm$ standard deviation over 5 independent runs.}
\label{tab:seed_ablation}
\resizebox{0.5\textwidth}{!}{%
\renewcommand{\arraystretch}{1.05}
\begin{tabular}{cccc}
\hline
 & & \multicolumn{2}{c}{\textbf{Perplexity$\downarrow$}} \\ \cline{3-4}
\multirow{-2}{*}{\textbf{Seed}}    & \multirow{-2}{*}{\textbf{Avg. Acc.$\uparrow$}} & \textbf{WikiText} & \textbf{LAMBADA} \\ \hline
2023 & 50.1 & 20.71 & 13.97 \\
43   & 50.2 & 20.51 & 13.08 \\
44   & 50.7 & 20.56 & 12.67 \\
45   & 50.0 & 20.56 & 14.20 \\
46   & 49.9 & 20.77 & 13.65 \\
\hdashline
\textbf{Mean $\pm$ Std} & \textbf{50.2 $\pm$ 0.3} & \textbf{20.62 $\pm$ 0.11} & \textbf{13.51 $\pm$ 0.63} \\
\textbf{Min--Max} & \textbf{49.9--50.7} & \textbf{20.51--20.77} & \textbf{12.67--14.20} \\
\hline
\end{tabular}
}
\end{table}
\subsection{Hardware setup}
\label{apx:hardware_setup}
For all experiments if not specified otherwise ran on a server with 256-core AMD EPYC 7742 CPU and 4$\times$ NVIDIA A100-SXM4-40GB GPUs. In some cases we utilized only a single GPU (e.g. for time measurements)

%%%%%%%%%%%%%%%%%%%%%%%%%%%%%%%%%%%%%%%%%%%%%%%%%%%%%%%%%%%%

\newpage
\section*{NeurIPS Paper Checklist}

\begin{enumerate}

\item {\bf Claims}
    \item[] Question: Do the main claims made in the abstract and introduction accurately reflect the paper's contributions and scope?
    \item[] Answer: \answerYes{}
    \item[] Justification: The abstract and introduction state the paper's main contributions and scope: a training-free Transformer compression method based on orthogonal dictionary factorization, closed-form Procrustes and sparse-coding updates, one-shot dynamic compression allocation, and compatibility with post-training quantization. These claims are supported by method description and experimental results in the main text, and additional evaluations and ablations in the Appendix.

\item {\bf Limitations}
    \item[] Question: Does the paper discuss the limitations of the work performed by the authors?
    \item[] Answer: \answerYes{}
    \item[] Justification: The paper discusses limitations in Section~\ref{sec:lim_conc}, including dependence on representative calibration data, sensitivity of the whitening step to ill-conditioned Gram/covariance matrices, and the current use of a fixed sparsity pattern. The section also outlines possible extensions such as stabilized whitening and learning the sparsity structure.

\item {\bf Theory assumptions and proofs}
    \item[] Question: For each theoretical result, does the paper provide the full set of assumptions and a complete (and correct) proof?
    \item[] Answer: \answerYes{}
    \item[] Justification: The theoretical results are stated with their assumptions in Appendix~\ref{app:compot_convergence}: the dictionary is constrained to the Stiefel set, sparse codes are column-wise \(s\)-sparse, and both subproblems are solved exactly. The appendix provides complete proofs for the closed-form sparse-coding update, the orthogonal Procrustes dictionary update, and monotone convergence of the COMPOT objective with corresponding references to the prior art.

    \item {\bf Experimental result reproducibility}
    \item[] Question: Does the paper fully disclose all the information needed to reproduce the main experimental results of the paper to the extent that it affects the main claims and/or conclusions of the paper (regardless of whether the code and data are provided or not)?
    \item[] Answer: \answerYes{}
    \item[] Justification: The paper provides the main information needed to reproduce the experimental results in Section~\ref{sec:experiments} and the appendix, including model families, compressed projection types, compression ratios, calibration datasets and sequence lengths, evaluation benchmarks, metrics, baseline protocols, implementation details, allocation procedures, and hyperparameter choices.

\item {\bf Open access to data and code}
    \item[] Question: Does the paper provide open access to the data and code, with sufficient instructions to faithfully reproduce the main experimental results, as described in supplemental material?
    \item[] Answer: \answerYes{}
    \item[] Justification: The code is submitted with the paper, together with instructions for reproducing the main experimental results. The experiments use publicly available models, calibration datasets, and evaluation benchmarks, with data sources and experimental protocols described in Section~\ref{sec:experiments} and the appendix.

\item {\bf Experimental setting/details}
    \item[] Question: Does the paper specify all the training and test details (e.g., data splits, hyperparameters, how they were chosen, type of optimizer) necessary to understand the results?
    \item[] Answer: \answerYes{}
    \item[] Justification: The paper specifies the experimental setup in Section~\ref{sec:experiments} and the appendix, including calibration datasets, sequence lengths, compressed projection types, compression ratios, evaluation benchmarks, metrics, and key hyperparameters such as dictionary size, sparsity ratio, initialization, and number of alternating updates. Since COMPOT is training-free, no optimizer or training schedule is used.

\item {\bf Experiment statistical significance}
    \item[] Question: Does the paper report error bars suitably and correctly defined or other appropriate information about the statistical significance of the experiments?
    \item[] Answer: \answerYes{}
    \item[] Justification: The paper reports statistical variability in the appendix using repeated COMPOT runs with different random seeds. We provide the corresponding mean and standard deviation for the relevant metrics, enabling assessment of run-to-run variability and supporting the statistical reliability of the reported improvements.

\item {\bf Experiments compute resources}
    \item[] Question: For each experiment, does the paper provide sufficient information on the computer resources (type of compute workers, memory, time of execution) needed to reproduce the experiments?
    \item[] Answer: \answerYes{}
    \item[] Justification: The paper reports the compute setup and runtime information needed to reproduce the experiments in the appendix, including GPU type, memory-relevant model scales, wall-clock compression time, and implementation details for the main compression runs. The reported timing experiments describe the hardware used and the per-layer optimization time, while the experimental setup specifies the models, calibration sizes, and evaluation protocols.
    
\item {\bf Code of ethics}
    \item[] Question: Does the research conducted in the paper conform, in every respect, with the NeurIPS Code of Ethics \url{https://neurips.cc/public/EthicsGuidelines}?
    \item[] Answer: \answerYes{}
    \item[] Justification: The research conforms to the NeurIPS Code of Ethics. The paper focuses on post-training compression of existing Transformer models, uses publicly available models and datasets, and does not involve human-subject experiments, private data, or data collection from vulnerable populations.

\item {\bf Broader impacts}
    \item[] Question: Does the paper discuss both potential positive societal impacts and negative societal impacts of the work performed?
    \item[] Answer: \answerYes{}
    \item[] Justification: The paper discusses broader impacts in Section~\ref{sec:lim_conc}. It notes potential positive impacts from reducing memory, compute, deployment cost, and energy use, which can improve accessibility. It also discusses possible negative impacts from lowering the barrier to deploying powerful models in settings where risks of the underlying models remain, including bias, privacy leakage, misinformation, or misuse, and recommends using compressed models with appropriate safeguards and monitoring.
    
\item {\bf Safeguards}
    \item[] Question: Does the paper describe safeguards that have been put in place for responsible release of data or models that have a high risk for misuse (e.g., pre-trained language models, image generators, or scraped datasets)?
    \item[] Answer: \answerNA{}
    \item[] Justification: The paper does not release new pretrained models, generated content, scraped datasets, or other high-risk assets. The released code implements a compression method for existing models, and any compressed models inherit the usage restrictions, safeguards, and licenses of the corresponding base models and datasets.

\item {\bf Licenses for existing assets}
    \item[] Question: Are the creators or original owners of assets (e.g., code, data, models), used in the paper, properly credited and are the license and terms of use explicitly mentioned and properly respected?
    \item[] Answer: \answerYes{}
    \item[] Justification: The paper credits the original creators of the publicly available models, datasets, benchmarks, and baseline codebases used in the experiments through citations in the main text and appendix. All assets are used according to their corresponding licenses and terms of use. Any compressed model produced by \textsc{COMPOT} is a derivative of the corresponding base model and therefore inherits the original model's license, usage restrictions, and attribution requirements.

\item {\bf New assets}
    \item[] Question: Are new assets introduced in the paper well documented and is the documentation provided alongside the assets?
    \item[] Answer: \answerYes{}
    \item[] Justification: The paper introduces and releases the \textsc{COMPOT} code as a new asset. The submitted code package includes documentation for running the method, reproducing the main experiments, key hyperparameters, expected inputs and outputs, and licensing information. The method does not introduce new datasets or pretrained base models; compressed models are derivatives of existing models and inherit their corresponding licenses and usage restrictions.

\item {\bf Crowdsourcing and research with human subjects}
    \item[] Question: For crowdsourcing experiments and research with human subjects, does the paper include the full text of instructions given to participants and screenshots, if applicable, as well as details about compensation (if any)? 
    \item[] Answer: \answerNA{}
    \item[] Justification: The paper does not involve crowdsourcing, human-subject experiments, user studies, or new human data collection.

\item {\bf Institutional review board (IRB) approvals or equivalent for research with human subjects}
    \item[] Question: Does the paper describe potential risks incurred by study participants, whether such risks were disclosed to the subjects, and whether Institutional Review Board (IRB) approvals (or an equivalent approval/review based on the requirements of your country or institution) were obtained?
    \item[] Answer: \answerNA{}
    \item[] Justification: The paper does not involve crowdsourcing, human-subject experiments, user studies, or new human data collection.

\item {\bf Declaration of LLM usage}
    \item[] Question: Does the paper describe the usage of LLMs if it is an important, original, or non-standard component of the core methods in this research? Note that if the LLM is used only for writing, editing, or formatting purposes and does \emph{not} impact the core methodology, scientific rigor, or originality of the research, declaration is not required.
    %this research? 
    \item[] Answer: \answerNA{}
    \item[] Justification: The core method development in this research does not use LLMs as an important, original, or non-standard component. Any LLM use, if applicable, was limited to writing, editing, or formatting assistance and did not affect the methodology, experiments, scientific rigor, or originality of the work.

\end{enumerate}

\end{document}